\documentclass{article} 
\usepackage{iclr2027_conference,times}

\usepackage{amsmath,amsfonts,bm}

\def\eqref#1{equation~\ref{#1}}

\def\1{\bm{1}}

\DeclareMathAlphabet{\mathsfit}{\encodingdefault}{\sfdefault}{m}{sl}
\SetMathAlphabet{\mathsfit}{bold}{\encodingdefault}{\sfdefault}{bx}{n}

\usepackage{fix-cm}
\usepackage[utf8]{inputenc}
\usepackage[T1]{fontenc}
\usepackage{graphicx,booktabs,array,tabularx,longtable,adjustbox}
\usepackage{amsmath,amssymb,mathtools}
\usepackage{microtype}
\usepackage[table]{xcolor}
\usepackage{pifont}
\usepackage{enumitem}
\usepackage{caption,subcaption}
\usepackage{float}
\usepackage{multirow}
\usepackage{placeins}
\usepackage[most]{tcolorbox}
\usepackage{listings}
\usepackage{hyperref,url}
\usepackage{fvextra}
\usepackage{chngcntr}
\hypersetup{hidelinks}
\newcolumntype{Y}{>{\raggedright\arraybackslash}X}
\newcolumntype{L}[1]{>{\raggedright\arraybackslash}p{#1}}
\newcommand{\cmark}{\textcolor{green!50!black}{\ding{51}}}
\newcommand{\xmark}{\textcolor{red!70!black}{\ding{55}}}
\newcommand{\metric}[2]{\mbox{#1\,$\pm$\,#2}}
\newcommand{\Note}[1]{\par\vspace{2pt}{\raggedright\scriptsize #1\par}}

\hypersetup{
    colorlinks=true,
    citecolor=teal
}

\title{Learning to Select Source-Traceable Evidence for Language-Model Prediction from Irregular Clinical Time Series}

\author{Anonymous authors}

\iclrfinalcopy 

\author{\textbf{Kwanhyung Lee}$^{1,2}$,
\textbf{Juhwan Choi}$^{2}$,
\textbf{Jongheon Kim}$^{1}$,
\textbf{Joohyung Lee}$^{2}$, 
\textbf{Hyeongwon Jang}$^{1}$, \\
\textbf{Jeonguk Lee}$^{2}$, 
\textbf{Jisoo Jung}$^{2}$, 
\textbf{Eunho Yang}$^{1,2}$ \\
$^{1}$Kim Jaechul Graduate School of AI, KAIST \quad
$^{2}$AITRICS \\
\texttt{kwanlee9209@kaist.ac.kr}
}

\begin{document}
\maketitle

\begin{abstract}
Numerical time-series models effectively process irregular electronic health record (EHR) trajectories, but do not expose which temporal patterns support each prediction as readable evidence. Existing text-based interfaces either serialize observations, preserving source traceability but offering limited clinical interpretation, or generate patient-level summaries that improve readability but can obscure links to source measurements. We introduce \textbf{STEP-CTS} (\textbf{S}ource-\textbf{T}raceable \textbf{E}vidence for \textbf{P}rediction from \textbf{C}linical \textbf{T}ime-\textbf{S}eries), which learns to select source-traceable text evidence for a language-model predictor. Multi-scale window statistics of each trajectory are verbalized as sets of deterministic threshold predicates, each set linked to its source window (e.g., ``[5–7 h]: last temperature at least 38\,$^\circ$C''). An offline LLM, run once per unique predicate set without access to patient records, the prediction task, or outcome labels, attaches clinical concepts such as ``\emph{fever}'' to their supporting predicates and abstains when none applies. Each predicate set, together with its concepts, forms an evidence unit. A learned Evidence Selector selects a fixed-size subset of the evidence units, which a pretrained clinical language-model encoder reads to make the prediction, with no patient-level text generation. Across three ICU benchmarks, STEP-CTS outperforms evaluated text-based baselines, improving AUPRC over the strongest by 5.1, 3.7, and 11.7 percentage points on P2012, MIMIC-III, and P2019, respectively, and is competitive with dedicated numerical time-series models. Ablations show that clinical concepts and learned evidence selection each contribute to predictive performance. In a blinded clinician study, the selected evidence is rated as traceable as deterministic serialization, near the ceiling of the scale, and above generated summaries on clinical interpretation. 

\end{abstract}


\section{Introduction}
Large language models (LLMs) encode broad prior knowledge from large-scale corpora, which is especially valuable in healthcare, where clinical measurements such as vital signs and laboratory values require domain knowledge to interpret. Language also provides a natural medium for presenting evidence for human inspection, which matters for healthcare prediction, where models should support both predictive accuracy and interpretable evidence. Recent EHR modeling efforts have increasingly adopted LLMs through approaches such as instruction-tuned EHR interfaces \citep{wu2024llemr}, knowledge-augmented prediction and reasoning \citep{jiang2024graphcare,jiang2025kare}, and LLM-derived representations for irregular ICU time series \citep{ji2026record2vec}.


Irregular EHR time-series prediction, however, remains dominated by numerical models that directly ingest values, timestamps, and missingness patterns \citep{che2018grud,shukla2021mtan,horn2020seft,tipirneni2022strats,zhang2022raindrop,luo2024kedgn,labach2023duett}. Existing LLM-based EHR interfaces often underperform such predictors \citep{gao2024raw,chen2026clinicalbench}. A key challenge is that raw EHR time series are not naturally linguistic, but consist of irregularly timed labs, vitals, and interventions, missingness patterns, and static covariates. Applying LLMs to EHR time series therefore requires an interface that converts these observations into language without losing the predictive signal that numerical models exploit, while keeping the resulting text inspectable.

Existing EHR-to-language interfaces fall into two broad groups. Rule-based approaches serialize observations with fixed templates, preserving source values and avoiding patient-level generation \citep{gao2024raw,fadlon2025decode}. They are deterministic, but often heuristic and exhaustive: they verbalize observations because they are present, not because they are informative. This yields long, redundant inputs that are difficult to inspect. Model-based approaches instead use LLMs to summarize, contextualize, or embed patient trajectories \citep{ji2026record2vec,lee2025timecp}. They can improve readability, but may require costly patient-level generation, raise deployment and privacy concerns, and require faithfulness checks because generated text can omit or distort source measurements \citep{jonnagaddala2025privacy,asgari2025framework,vishwanath2024faithfulness}. These concerns are especially consequential in healthcare, where omitted or distorted measurements can lead to misleading clinical interpretations, and where the evidence for a given record is expected to be reproducible. These limitations point to a missing interface: selective and readable like a summary, yet deterministic and source-traceable like rule-based serialization.

In response to these challenges, we propose \textbf{STEP-CTS} (\textbf{S}ource-\textbf{T}raceable \textbf{E}vidence for \textbf{P}rediction from \textbf{C}linical \textbf{T}ime-\textbf{S}eries), a language interface for irregular EHR time-series prediction built from three modules that verbalize, select, and read evidence. A Temporal Evidence Verbalizer (\textsc{TEV}) converts multi-scale look-back window summaries of each trajectory into candidate evidence units: it applies a fixed inventory of threshold rules learned once from outcome labels on the training split (instantiated with gradient-boosted trees; Section~\ref{sec:tev}), and renders every activated rule as deterministic threshold predicates, e.g., ``[32--48h]: minimum MAP is at most 65'' or ``[16--20h]: last heart rate is higher than 110.'' It then attaches outcome-agnostic clinical concepts (e.g., \emph{hypotension}, \emph{tachycardia}) to the predicates that support them, abstaining when none applies. Each resulting evidence unit carries explicit links to its source time window and rule. An Evidence Selector (\textsc{ES}) selects a fixed-size set of $K$ units per patient, and a clinical language-model reader (LM reader) predicts from those units; \textsc{ES} and the LM reader are trained jointly.

This design combines strengths of prior interfaces. Like rule-based serialization, the predicate text is deterministic and source-traceable, while each attached clinical concept is linked to its supporting predicate clauses and, through them, to the source time windows. Like model-based summarization, the evidence is selective and readable, yet nothing is LLM-generated per patient: concepts are produced once per rule, offline, and inference involves no autoregressive decoding. Because the LM reader accesses the time series only through the selected evidence units, the evidence a clinician inspects is a deterministic regrouping of the units the LM reader read, rather than a post-hoc attribution, in the spirit of select-then-predict rationale models \citep{lei2016rationalizing,jain2020fresh}. Unlike methods that use LLMs to generate features or refine rules \citep{nam2024octree,ye2025delta}, STEP-CTS keeps the rule inventory fixed and uses the language model only to read verbalized evidence.

\paragraph{Contributions.}
Our contributions are threefold.
\begin{itemize}[leftmargin=1.25em,itemsep=0.1em,topsep=0.2em]
    \item \textbf{A source-traceable evidence interface for irregular clinical time series.}
    STEP-CTS is an EHR-to-language interface that is at once deterministic, selective, and traceable to source time windows: multi-scale window summaries are verbalized as threshold predicates with outcome-agnostic clinical concepts, without patient-level LLM generation or autoregressive decoding at inference.

    \item \textbf{Learned evidence selection under a fixed budget.}
    An Evidence Selector (\textsc{ES}), trained jointly with the language-model reader, selects $K$ evidence units per patient. The selected units are the reader's entire time-series input, so the evidence presented to a clinician is a deterministic regrouping of what the model read, not a separate post-hoc explanation.

    \item \textbf{Predictive performance and clinician-rated evidence.}
    Across three ICU benchmarks, STEP-CTS achieves the strongest results among text-based EHR time-series interfaces and remains competitive with dedicated numerical time-series models. Blinded clinicians rate its selected evidence as traceable as deterministic serialization, near the ceiling of the scale, and as interpretable  as generated summaries.
\end{itemize}

\section{Related work}

\paragraph{Numerical models for irregular clinical time series.}
Models for irregular EHR time series handle missingness and uneven sampling through recurrent decay, set functions, time-aware attention, or variable graphs \citep{che2018grud,shukla2021mtan,horn2020seft,tipirneni2022strats,zhang2022raindrop,luo2024kedgn,labach2023duett,wang2026garlic}. They predict directly from values, timestamps, and missingness patterns and provide strong baselines, but expose at most importance scores over observations, signals, or relations \citep{horn2020seft,tipirneni2022strats,wang2026garlic}, not readable evidence with clinical interpretation.

\paragraph{Text-based interfaces for irregular clinical time series.}
Serialization approaches verbalize observations with fixed templates \citep{gao2024raw,zhu2024prompting,fadlon2025decode}; the text reflects the source observations but is often exhaustive, offers limited clinical interpretation, and depends on heuristic choices in temporal aggregation, ordering, and verbalization. Generation approaches instead use an LLM to summarize or contextualize the trajectory \citep{lee2025timecp,ji2026record2vec}, yielding compact text at the cost of patient-level generation and harder verification against source measurements. Neither provides evidence that is deterministically constructed, traceable to source time windows, and organized by clinical concept, which is the gap STEP-CTS targets.


\paragraph{Threshold rules as language-model input.}
TP-BERTa \citep{yan2024tpberta} bins numerical features by target-aware splits and treats bins as tokens for language models. Tree ensembles form multi-feature threshold conjunctions, which LLM-tree hybrids use to generate features or refine rules \citep{nam2024octree,ye2025delta}, while ReSS \citep{yi2026ress} generates rationales on static tables. STEP-CTS instead verbalizes multi-scale thresholds as readable predicates for the language model.


\paragraph{Select-then-predict rationales.}
Extractive rationale models train a selector and a predictor that reads only the selected subset, making selection faithful by construction and testable by deletion \citep{lei2016rationalizing,jain2020fresh,deyoung2020eraser,rudin2019stop}; post-hoc attribution instead explains another predictor and, for time series, yields saliency masks rather than statements \citep{lundberg2020treeshap,crabbe2021dynamask,queen2023timex}. Unlike token rationales or event retrieval \citep{kim2024remed}, STEP-CTS selects clinician-readable threshold evidence.
\begin{figure}[t]
\centering
\includegraphics[width=1.00\linewidth]{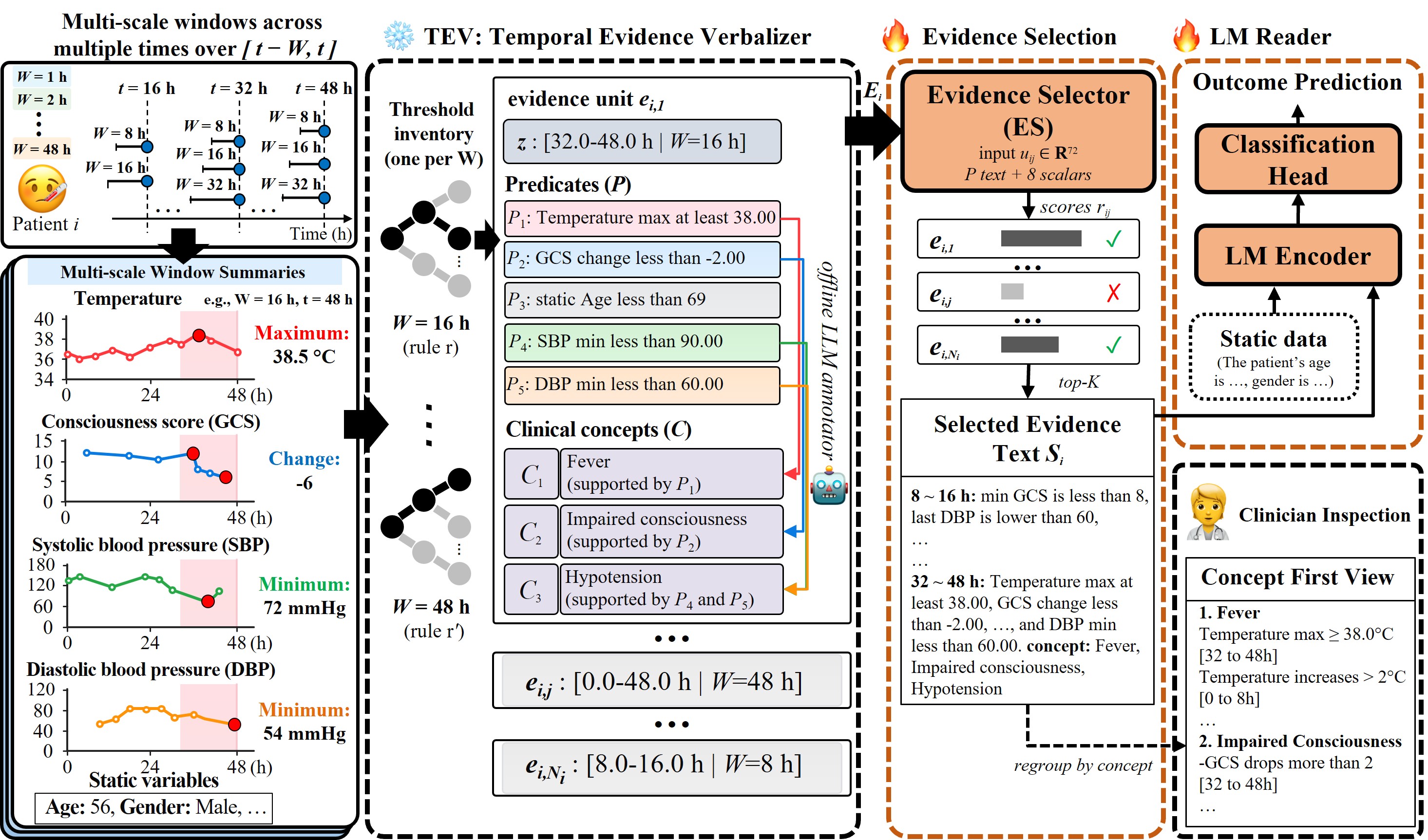}
\caption{Overview. TEV (frozen) summarizes each trajectory over multi-scale windows and verbalizes activated rules as source-linked predicates with clinical concepts; ES keeps $K$ units; the LM reader predicts from the static prefix and selected evidence. ES and reader are trained jointly.}
\label{fig:method}
\end{figure}

\section{Method}
\label{sec:method}

\paragraph{Overview.}
STEP-CTS builds a readable, source-traceable text interface between irregular EHR trajectories and a clinical language-model reader (LM reader). As shown in Figure~\ref{fig:method}, each trajectory is summarized over multi-scale look-back windows, and the Temporal Evidence Verbalizer (\textsc{TEV}) summarizes each trajectory over multi-scale windows, activates rules from a fixed threshold inventory, and verbalizes each activated rule as threshold predicates with clinical concepts. (\textsc{ES}) keeps the top $K$ resulting evidence units, and the LM reader predicts from a static-covariate prefix followed by the selected evidence. \textsc{TEV} is fit once on the training split and frozen, so a given record always yields the same evidence text; \textsc{ES} and the LM reader are trained jointly.

\paragraph{Problem setup.}
For patient $i$, let $\mathcal{O}_i=\{(v_j,t_j,x_j)\}_{j=1}^{n_i}$ denote the $n_i$ irregular EHR observations, where $v_j$ is the measured variable, $t_j$ the observation time, and $x_j$ the observed value. Let $s_i$ denote static covariates and $y_i\in\{0,1\}$ the binary outcome. We render $s_i$ as a deterministic text prefix $c_i$, such as ``Age: 69, ...'', which is prepended to the selected evidence read by the LM reader.

From $\mathcal{O}_i$, \textsc{TEV} constructs a candidate set of evidence units $\mathcal{E}_i=\{e_{i1},\ldots,e_{iN_i}\}$, from which \textsc{ES} selects $S_i\subseteq\mathcal{E}_i$. The final predictor is
\begin{equation}
    \hat p_i
    =
    \sigma\!\left(
    w^\top f_\theta\!\left(
    c_i \;\Vert\;
    \operatorname{Render}(S_i)
    \right) + b_0
    \right),
\end{equation}

where $\hat p_i$ is the predicted probability of $y_i=1$, $\sigma$ is the sigmoid function, $\Vert$ denotes text concatenation, $\operatorname{Render}$ serializes the selected units in chronological order, $f_\theta$ is the LM reader's encoder representation, and $w,b_0$ form the linear classification head.

\paragraph{Multi-scale window summaries.}
Clinical risk signals appear at different temporal scales, from abrupt short-window changes to sustained abnormalities over longer histories, so we summarize each trajectory over multiple look-back windows. For a prediction time $t^\ast$, set by each benchmark's standard protocol (Appendix~\ref{app:data}), and a window size $W$, we tile the history with non-overlapping windows $[t-W,t]$ anchored at $t^\ast$ and extending backward, whose endpoints form a grid $\mathcal{G}_i(W)$; at $t^\ast=24$h, for example, the $W=1$h window ending at $t^\ast$ cover $[23,24]$h. For each variable and window, we compute nine fixed statistics: last value, mean, standard deviation, minimum, maximum, count, net change, time since last observation, and a missingness indicator.

\paragraph{Temporal Evidence Verbalizer.}
\label{sec:tev}
\textsc{TEV} forms evidence units from window summaries in three steps.

\textit{Threshold inventory.} For each window size $W$, we learn a set of threshold rules from the summary rows $(\phi_W(i,t),\,y_i)$ of all training patients $i$ and all grid endpoints $t\in\mathcal{G}_i(W)$. We instantiate the inventory with gradient-boosted trees (XGBoost): each root-to-leaf path is a conjunction of thresholds, so the trees act as an outcome-supervised discretizer of the window statistics, analogous to the target-aware binning of \citet{yan2024tpberta}, and are not used as the predictor. The inventory is frozen after training; later patients can only activate rules, not modify them. Each summary $\phi_W(i,t)$ activates one rule $r$ per tree, defining a candidate unit indexed by the source tuple $(t,W,r)$. Because many rules fire per $(i,t,W)$, we retain the $M$ with the largest score $s_r=\log(1+n_r)\,\lvert p_r-p_0\rvert$, where $n_r$ and $p_r$ are the number and positive rate of training rows satisfying $r$ and $p_0$ is the training-set base rate. Consecutive windows of the same size that activate the same rule are merged into one unit spanning their union, in which case $t$ denotes the spanned interval.

\textit{Predicate rendering.} Each unit stores its source tuple $(t,W,r)$, so the original rule is recoverable. Each threshold in the rule is its learned split value, rendered by a fixed template into a threshold predicate, a deterministic condition on a window-summary feature (e.g., a split on $x<40$ yields ``$x$ less than 40''), and redundant bounds within a rule are merged.

\textit{Concept annotation.} An LLM annotator, run offline once per rule (i.e., per unique predicate set; 5,173/5,747/5,452 calls on P2012/MIMIC-III/P2019 against 7,670/14,632/24,817 training patients), receives the clauses $P(e)$ and feature definitions but no patient record, task description, outcome label, or outcome association, so the resulting concepts are outcome-agnostic and no record is ever exposed to a generative model. Because not every statistically predictive rule is clinically interpretable, the annotator may abstain rather than forcing a concept onto every unit. It returns zero or more concepts $C(e)=\{(c_k,J_k)\}_k$, where $J_k\subseteq\{1,\ldots,J\}$ indexes the clauses cited as support for $c_k$. A concept with a cutoff in Table~\ref{tab:cutoffs} is kept only if its cited clauses reach it, e.g., temperature at least $38^\circ$C for fever (Appendix~\ref{app:concept_prompt}). Beyond this check, support links are not model inputs; they record the annotator's justification and are used only for clinician-facing presentation. We write $P$ for predicate-only text and $P+C$ for predicate text followed by $C(e)$ when nonempty (Figure~\ref{fig:evidence_example}); our full model uses the $P+C$ format.

\paragraph{Evidence Selector.}
\textsc{TEV} yields many candidate evidence units per patient; \textsc{ES} ranks them and keeps the top $K$. For each unit $e_{ij}$, a cached embedding $v_{ij}$ of its predicate text is passed through a trainable projection $\Pi_\psi$ and concatenated with metadata features $m_{ij}$ (Table~\ref{tab:selector_metadata}) to form $u_{ij}=[\Pi_\psi(v_{ij});m_{ij}]$. A token projection maps $u_{ij}$ to the width of a lightweight Transformer encoder $T_\phi$, which contextualizes the candidates as a set without positional encodings, and a linear head produces scores and selection probabilities:
\begin{equation}
h_{ij}=T_\phi(u_{i1},\ldots,u_{iN_i})_j,\qquad
r_{ij}=w_r^\top h_{ij}+b_r,\qquad
q_{ij}=\sigma(r_{ij}).
\end{equation}

Concepts enter the selector only through a binary availability feature in $m_{ij}$. \textsc{ES} retains the top-$K$ units by $r_{ij}$, a fixed number per patient, so evidence length is bounded and explanations are comparable across patients; sensitivity to $K$ is reported in Appendix~\ref{app:retrieval}. 

\paragraph{Self-critical selector learning.}
Selection is discrete, and the quality of a selected set is observed only through the prediction loss after the LM reader reads the assembled input. We therefore train \textsc{ES} with a self-critical policy-gradient objective \citep{rennie2017selfcritical}. For each patient, a sampled set $S_i^s$ keeps the $K$ highest-scoring units among those sampled from $\mathrm{Bernoulli}(q_{ij})$, and a greedy baseline $S_i^g$ takes the top-$K$ units by score. At inference, only the greedy selection is used.

The LM reader reads two assemblies that share $c_i$ and differ only in the selected evidence, with binary cross-entropy losses $L_i^s$ and $L_i^g$. The LM reader and its head are trained on both, and the selector receives a normalized self-critical advantage:
\begin{equation}
    \mathcal{L}_{\mathrm{task}}
    =
    \frac{1}{2|\mathcal{B}|}
    \sum_{i\in\mathcal{B}}
    \left(L_i^s+L_i^g\right),
    \qquad
    \hat A_i
    =
    \operatorname{stopgrad}\!\left(
    \frac{(L_i^g-L_i^s)-\mu_{\mathcal{B}}}
    {s_{\mathcal{B}}+\epsilon}
    \right),
\end{equation}

where $\mu_{\mathcal{B}}$ and $s_{\mathcal{B}}$ are the mini-batch mean and standard deviation of $L_i^g-L_i^s$, so a sampled set receives a positive advantage when it lowers the loss relative to the greedy set by more than the batch average. The selector loss is
\begin{equation}
    \mathcal{L}_{\mathrm{sel}}^{\mathrm{SC}} =
    -\frac{1}{|\mathcal{B}|}
    \sum_{i\in\mathcal{B}}
    \hat A_i\,
    \frac{1}{N_i}
    \sum_{j=1}^{N_i}
    \left[
    z^s_{ij}\log q_{ij}
    +
    (1-z^s_{ij})\log(1-q_{ij})
    \right],
\end{equation}

where $z^s_{ij}\in\{0,1\}$ indicates whether $e_{ij}$ is in $S_i^s$ after the top-$K$ cap, a tractable surrogate for the composite policy, and the inner sum is normalized by $N_i$ so that patients with few and many candidates contribute comparably. The full objective is
\begin{equation}
    \mathcal{L}
    =
    \mathcal{L}_{\mathrm{task}}
    +
    \lambda_{\mathrm{sel}}\mathcal{L}_{\mathrm{sel}}^{\mathrm{SC}}
    -
    \lambda_{\mathrm{ent}}\mathcal{H}(\pi_\phi),
\end{equation}

where $\mathcal{H}(\pi_\phi)$ is an entropy regularizer on the Bernoulli selection policy that discourages premature collapse of $q_{ij}$ to 0 or 1, and no sparsity penalty is needed because the top-$K$ cap enforces the evidence budget. The \textsc{TEV} outputs (inventory, predicate text, concepts) and the embedding cache are fixed throughout, while the selector, LM reader, and classification head are trained jointly: $\mathcal{L}_{\mathrm{task}}$ updates the reader and head, and the selector receives gradients only through $\mathcal{L}_{\mathrm{sel}}^{\mathrm{SC}}$ and the entropy term (hyperparameters in Appendix~\ref{app:impl}).

\paragraph{Evidence presentation for inspection.}
The LM reader reads selected units chronologically, from source window to predicates to concept. For clinician inspection, we regroup the same evidence concept-first: clinical concept $\rightarrow$ supporting predicates $\rightarrow$ source windows (e.g., \textsc{Fever} $\rightarrow$ temperature $\geq 38^\circ$C $\rightarrow$ [5--7 h], [12--16 h]; Figure~\ref{fig:method}, right), ordered by average selection score $r_{ij}$. Units sharing a concept are merged, and a predicate recurring across windows is listed once with all its windows, so the view is more compact than the chronological input and a clinician can start from the interpretation and trace it back to the measurement (Appendix~\ref{app:explanation}). The view involves no generation and leaves the LM reader's input and prediction unchanged. It is a projection of $S_i$ rather than a substitute for it: predicate-only units are listed below the concept groups in the full rendering (Appendix~\ref{app:explanation}).

\section{Experiments}
\label{sec:experiments}

\paragraph{Datasets and tasks.}
We evaluate on three public ICU benchmarks: PhysioNet 2012 (P2012) in-hospital mortality \citep{silva2012physionet2012}, MIMIC-III in-hospital mortality \citep{johnson2016mimic}, and PhysioNet 2019 (P2019) sepsis onset within the next six hours \citep{reyna2020physionet2019}. All methods share identical train, validation, and test cohorts (preprocessing and cohort details in Appendix~\ref{app:data}).

\paragraph{Shared components.}
To isolate differences in text representation from differences in downstream predictors, all adapted text baselines use the same BioClinical ModernBERT \citep{sounack2025bioclinicalmodernbert} reader as STEP-CTS. Generation-based components in Record2Vec, TimeCP, and STEP-CTS use the same Qwen3.5-27B \citep{qwen35}. Faithful variants otherwise retain their method-specific reproduced pipelines; exact instantiations, including embedding models, are listed in Table~\ref{tab:implementation}.

\paragraph{Baselines.}
We compare against two groups. The first is text-based EHR time-series interfaces: WRDP \citep{gao2024raw}, Decode Like a Clinician \citep{fadlon2025decode} (its 6-h per-event verbalization, the best-performing variant reported by the authors), TimeCP \citep{lee2025timecp}, and Record2Vec \citep{ji2026record2vec}. For each, we report a \textbf{faithful} variant that follows the original generator, embedder, and predictor, and an \textbf{adapted} variant that keeps the same patient text but replaces the predictor with the shared BioClinical ModernBERT. TimeCP and Record2Vec require patient-level generation; when it is needed at test time, it is counted in the latency protocol. The second group is numerical irregular time-series models: LOCF LSTM \citep{hochreiter1997long}, LOCF Transformer \citep{vaswani2017attention}, GRU-D \citep{che2018grud}, mTAND \citep{shukla2021mtan}, SeFT \citep{horn2020seft}, STraTS \citep{tipirneni2022strats}, KEDGN \citep{luo2024kedgn}, DuETT \citep{labach2023duett}, and GARLIC \citep{wang2026garlic}. We also report \textsc{XGB-WindowStats}, a single XGBoost model on the same multi-scale window statistics as STEP-CTS, with a 30-tree budget matched to the $K=30$ evidence budget so that both consume 30 activated rules per patient (Table~\ref{tab:numerical}; the full budget sweep, a validation-tuned model, and pooled per-window controls are in Appendix~\ref{app:xgb}). Numerical-baseline protocols are in Appendix~\ref{app:numeric_full}. Here, TimeCP denotes the contextualize-and-predict component introduced in the TimeCAP paper, rather than the full TimeCAP framework.

\paragraph{STEP-CTS configuration.}
Window-specific threshold inventories are instantiated with XGBoost for $W\in\{1,2,4,8,16,32,48\}$\,h (Appendix~\ref{app:impl}) with $M=5$ candidate rules per $(i,t,W)$. Concepts are annotated once rule by the shared Qwen3.5-27B, and predicate-text embeddings from Qwen3-Embedding-8B are computed once and projected to 64 dimensions; the rule inventory, predicate text and concepts are built on the training split and fixed thereafter. Unless stated otherwise, we use $K=30$, BioClinical ModernBERT as the LM reader (Details in Appendix~\ref{app:impl}, \ref{app:ablation_full}, and \ref{app:retrieval}).


\paragraph{Evaluation protocol.}
We report AUROC and AUPRC. Results are mean $\pm$ standard deviation over three seeds. For each seed, the checkpoint with the best validation AUROC is evaluated once on the test set. Validation labels are used only for model selection (learning rate, epoch, XGBoost early stopping, and the tree cap of the tuned control) and for the display threshold in the clinician study; test labels are never used. Latency is end-to-end wall-clock time per test patient (Appendix~\ref{app:latency}).

\paragraph{Explanation evaluation.}
Two clinicians rate four presentations for 50 P2012 test patients on three five-point axes: source traceability, clinical interpretation quality, and clinical usefulness (Table~\ref{tab:explanation}). Presentations are reviewed alongside the raw record in a dedicated interface (Figure~\ref{fig:clinician_view}), with method identities and outcomes hidden so that ratings reflect explanation quality, not prediction correctness. Three are text interfaces (STEP-CTS, Record2Vec, Decode); the fourth, SHAP-to-Text, renders the top-30 TreeSHAP attributions \citep{lundberg2020treeshap} of the 30-rule \textsc{XGB-WindowStats} as template text grouped by risk direction, following clinician-facing SHAP displays \citep{lundberg2018prescience}. Two LLM judges, GPT-5.5 \citep{openai2026gpt55} and Claude Sonnet 5 \citep{anthropic2026sonnet5}, apply the same rubric to 200 P2012 test patients, including the 50 clinician-reviewed cases, with two passes each. Full protocol and per-rater results are in Appendix~\ref{app:clinician}; examples in Appendix~\ref{app:case_study}.
\section{Results and Discussion}
\label{sec:results}
\begin{table}[h]
\centering
\caption{
Text-based EHR time-series interface comparison.
\textbf{Faithful} variants follow the original pipeline; \textbf{Adapted} variants keep the same patient text but use the shared BioClinical ModernBERT predictor (instantiations in Table~\ref{tab:implementation}).
Metrics are test mean $\pm$ standard deviation over three seeds; \textbf{bold} marks the highest mean per column within each panel.
}
\label{tab:text_baselines}

\newcommand{\stackmetric}[2]{%
  \shortstack{#1\\[-1pt]{\tiny $\pm$ #2}}%
}
\newcommand{\beststackmetric}[2]{%
  \shortstack{\textbf{#1}\\[-1pt]{\scriptsize $\pm$ #2}}%
}

\fontsize{7.2}{8.6}\selectfont
\setlength{\tabcolsep}{1.8pt}
\renewcommand{\arraystretch}{0.85}

\begin{tabularx}{\linewidth}{
@{}
l
ccc
*{6}{>{\centering\arraybackslash}X}
@{}
}
\toprule
Method
& \shortstack{No patient\\generation}
& \shortstack{Clinical\\interpretation}
& \shortstack{Source-\\traceable}
& \multicolumn{2}{c}{P2012}
& \multicolumn{2}{c}{MIMIC-III}
& \multicolumn{2}{c}{P2019}
\\

\cmidrule(lr){5-6}
\cmidrule(lr){7-8}
\cmidrule(lr){9-10}

&
&
&
&
AUROC & AUPRC
& AUROC & AUPRC
& AUROC & AUPRC
\\

\midrule

WRDP-Faithful
& \cmark & \xmark & \cmark
& \stackmetric{0.7673}{0.0051}
& \stackmetric{0.3452}{0.0060}
& \stackmetric{0.7594}{0.0084}
& \stackmetric{0.2992}{0.0144}
& \stackmetric{0.8509}{0.0012}
& \stackmetric{0.3038}{0.0096}
\\

WRDP-Adapted
& \cmark & \xmark & \cmark
& \stackmetric{0.8236}{0.0008}
& \stackmetric{0.4636}{0.0181}
& \stackmetric{0.7843}{0.0017}
& \stackmetric{0.3580}{0.0089}
& \stackmetric{0.8853}{0.0042}
& \stackmetric{0.3792}{0.0025}
\\

Record2Vec-Faithful
& \xmark & \cmark & \xmark
& \stackmetric{0.7858}{0.0004}
& \stackmetric{0.4127}{0.0012}
& \stackmetric{0.8017}{0.0017}
& \stackmetric{0.4323}{0.0025}
& \stackmetric{0.7889}{0.0014}
& \stackmetric{0.1485}{0.0031}
\\

Record2Vec-Adapted
& \xmark & \cmark & \xmark
& \stackmetric{0.8023}{0.0055}
& \stackmetric{0.4147}{0.0186}
& \stackmetric{0.8148}{0.0016}
& \stackmetric{0.4308}{0.0149}
& \stackmetric{0.8058}{0.0029}
& \stackmetric{0.1716}{0.0021}
\\

Decode-Faithful
& \cmark & \xmark & \cmark
& \stackmetric{0.7821}{0.0217}
& \stackmetric{0.4376}{0.0310}
& \stackmetric{0.8394}{0.0126}
& \stackmetric{0.4292}{0.0041}
& \stackmetric{0.7913}{0.0236}
& \stackmetric{0.3266}{0.0072}
\\

Decode-Adapted
& \cmark & \xmark & \cmark
& \stackmetric{0.8444}{0.0031}                                                                                  
& \stackmetric{0.4768}{0.0001}
& \stackmetric{0.8474}{0.0026}
& \stackmetric{0.4657}{0.0060}
& \stackmetric{0.8943}{0.0041}
& \stackmetric{0.3836}{0.0011}
\\

TimeCP-Faithful
& \xmark & \cmark & \xmark
& \stackmetric{0.6634}{0.0001}
& \stackmetric{0.2051}{0.0009}
& \stackmetric{0.6700}{0.0003}
& \stackmetric{0.1740}{0.0015}
& \stackmetric{0.5526}{0.0073}
& \stackmetric{0.0479}{0.0008}
\\

TimeCP-Adapted
& \xmark & \cmark & \xmark
& \stackmetric{0.8090}{0.0006}
& \stackmetric{0.4339}{0.0063}
& \stackmetric{0.8081}{0.0013}
& \stackmetric{0.4000}{0.0059}
& \stackmetric{0.7964}{0.0031}
& \stackmetric{0.1689}{0.0092}
\\

\midrule

\rowcolor{gray!20}
\textbf{STEP-CTS (Ours)}
& \cmark & \cmark$^{\ast}$ & \cmark
& \beststackmetric{0.8561}{0.0017}
& \beststackmetric{0.5274}{0.0049}
& \beststackmetric{0.8563}{0.0024}
& \beststackmetric{0.5025}{0.0056}
& \beststackmetric{0.9115}{0.0016}
& \beststackmetric{0.5004}{0.0157}
\\

\bottomrule
\end{tabularx}

\Note{
\textbf{No patient-level generation} excludes generative LLM calls on individual patient records at both training and inference. \textbf{Clinical interpretation} denotes findings beyond numeric values, trends, or thresholds. \textbf{Source traceability} denotes a deterministic pathway to source variables and time windows. Properties refer to the variant evaluated in Section~\ref{sec:experiments} and apply to both its faithful and adapted forms. $^{\ast}$Attached to 39.2/44.4/5.4\% of selected units on P2012/MIMIC-III/P2019; predicates are always present; concept coverage is analyzed in Appendix~\ref{app:concept_entailment}.
}
\end{table}

\paragraph{Prediction against text-based interfaces.}
STEP-CTS attains the highest AUROC and AUPRC among all text-based interfaces on all three benchmarks (Table~\ref{tab:text_baselines}), improving AUPRC over the second-best method by 5.1, 3.7, and 11.7 percentage points (pp) on P2012, MIMIC-III, and P2019, and AUROC by 1.2, 0.9, and 1.7~pp.

The two baseline families trade off different properties. Summarization interfaces (Record2Vec, TimeCP) require patient-level generation, moving inference from 171~ms to 12.8--19.6~s per patient (Appendix~\ref{app:latency}) and leaving no deterministic link from statement to record; clinicians rate Record2Vec far lower on traceability and found statements that could not be verified against the record (Appendix~\ref{app:case_study}). Serialization interfaces (WRDP, Decode) keep the link but convey only what was observed. STEP-CTS keeps the link while attaching clinical concepts.

\begin{table}[!htbp]
\centering
\caption{Component ablations. Ours uses $P+C$, ES, and BioClinical ModernBERT with $K=30$.}
\label{tab:ablations}

\fontsize{7.5}{9}\selectfont
\setlength{\tabcolsep}{1.0pt}
\renewcommand{\arraystretch}{0.85}

\begin{tabular*}{\linewidth}{@{\extracolsep{\fill}}l*{6}{c}@{}}
\toprule
Variant
& \multicolumn{2}{c}{P2012}
& \multicolumn{2}{c}{MIMIC-III}
& \multicolumn{2}{c}{P2019}
\\
\cmidrule(lr){2-3}
\cmidrule(lr){4-5}
\cmidrule(lr){6-7}
& AUROC & AUPRC
& AUROC & AUPRC
& AUROC & AUPRC
\\
\midrule
\rowcolor{gray!15}
\textbf{STEP-CTS (Ours)}
& \textbf{\metric{0.8561}{0.0017}}
& \textbf{\metric{0.5274}{0.0049}}
& \textbf{\metric{0.8563}{0.0024}}
& \textbf{\metric{0.5025}{0.0056}}
& \metric{0.9115}{0.0016}
& \textbf{\metric{0.5004}{0.0157}}
\\

\midrule
\multicolumn{7}{@{}l}{\textit{A.  Temporal Evidence Verbalizer (TEV)}}\\
w/o clinical concepts ($P$)
& \metric{0.8510}{0.0001}
& \metric{0.4997}{0.0043}
& \metric{0.8503}{0.0015}
& \metric{0.4650}{0.0158}
& \textbf{\metric{0.9147}{0.0048}}
& \metric{0.4783}{0.0040}
\\

\midrule
\multicolumn{7}{@{}l}{\textit{B. Evidence Selector (ES)}}\\



Rule-score
& \metric{0.8362}{0.0028} & \metric{0.4902}{0.0073}
& \metric{0.8293}{0.0051} & \metric{0.4596}{0.0018}
& \metric{0.8839}{0.0015} & \metric{0.3505}{0.0016}\\
Recency
& \metric{0.8471}{0.0036} & \metric{0.4836}{0.0087}
& \metric{0.8335}{0.0033} & \metric{0.4693}{0.0092}
& \metric{0.9040}{0.0054} & \metric{0.4876}{0.0063}\\
Random
& \metric{0.8449}{0.0003} & \metric{0.4700}{0.0098}
& \metric{0.8409}{0.0031} & \metric{0.4310}{0.0114}
& \metric{0.8946}{0.0027} & \metric{0.4214}{0.0111}\\

\midrule
\multicolumn{7}{@{}l}{\textit{C. Language Model Reader}}\\

ES-no-LM + statics
& \metric{0.8447}{0.0158}
& \metric{0.4820}{0.0342}
& \metric{0.8360}{0.0047}
& \metric{0.4318}{0.0265}
& \metric{0.8758}{0.0002}
& \metric{0.2875}{0.0017}
\\

Reused-ES-no-LM
& \metric{0.8501}{0.0037}
& \metric{0.5143}{0.0065}
& \metric{0.8535}{0.0007}
& \metric{0.4687}{0.0131}
& \metric{0.9044}{0.0042}
& \metric{0.4311}{0.0044}
\\

Random-init ModernBERT
& \metric{0.8498}{0.0051}
& \metric{0.5000}{0.0050}
& \metric{0.8397}{0.0039}
& \metric{0.4234}{0.0035}
& \metric{0.9006}{0.0010}
& \metric{0.4155}{0.0292}
\\

\bottomrule
\end{tabular*}

\Note{
\textbf{Bold}: highest mean per column.
$P$: predicates; $C$: concepts. Panel B selectors rank the same $M{=}5$ candidate pool as Ours under the same $K{=}30$ budget; further variants are in Table~\ref{tab:full_ablations}.
No-LM rows replace the reader with a rule-ID MLP: ES-no-LM trains the selector and MLP jointly (+ static covariates); Reused-ES-no-LM freezes the selector trained for Ours and trains only the MLP.
}
\end{table}

\paragraph{Component contributions.}
Table~\ref{tab:ablations} ablates the three components of STEP-CTS.

\textit{Temporal Evidence Verbalizer (TEV).}
Removing clinical concepts reduces AUPRC by 2.8, 3.8, and 2.2~pp on P2012, MIMIC-III, and P2019, with a mixed effect on AUROC. A frozen-reader control confirms that the trained reader uses this text at test time: with selector, predicates, and reader fixed, retaining a larger fraction of the selected concept sentences raises AUPRC monotonically on all three datasets (0.5090 to 0.5274, 0.4949 to 0.5025, and 0.4763 to 0.5004; Table~\ref{tab:concept_retention_values}; Figure~\ref{fig:concept_retention}). The gain comes from supported concepts, not coverage. Full labeling (no abstention) raises rule coverage from 38.6/43.0/26.0\% to 90.0/86.9/86.3\% without improving AUPRC, task-aware annotation gives no consistent gain, and appending outcome-association text ($+A$) lowers AUPRC relative to $P+C$ on all datasets (Table~\ref{tab:full_ablations}). We therefore keep the outcome-agnostic, abstaining default; an independent judge rates 4.5\% of retained concepts unsupported (Appendix~\ref{app:concept_entailment}). After selection, 39.2/44.4/5.4\% of units carry concepts, making the concept view sparse on P2019, where 46.3\% of test patients have none. Every unit nevertheless remains a source-linked predicate set.


\textit{Evidence Selector (ES).}
Replacing ES with rule-score, recency, or random ranking over the same candidate pool and budget lowers mean AUPRC on all three datasets (Table~\ref{tab:ablations}). Training ES without an LM reader (ES-no-LM) performs substantially worse, whereas reusing the jointly trained ES with a rule-ID MLP in place of the LM reader retains AUPRC of 0.5143, 0.4687, and 0.4311, well above ES-no-LM, so LM-coupled training yields a selector whose choices remain predictive even for a non-text reader (the remaining gap to the full model is analyzed under the reader below). ES favors longer-window and more recent evidence (Table~\ref{tab:selection_enrichment}).

\textit{Language-model reader.}
Random initialization of ModernBERT reduces AUPRC by 2.7, 7.9, and 8.5 points (Table~\ref{tab:ablations}), so pretrained language representations matter on every dataset. Generic ModernBERT matches BioClinical ModernBERT on P2012 (0.5250 versus 0.5274) and MIMIC-III (0.4992 versus 0.5025) but drops on P2019 (0.4759 versus 0.5004; Table~\ref{tab:full_ablations}), so clinical pretraining helps most on the sepsis task. The Reused-ES control isolates what the reader adds beyond knowing which rules were selected: with the same selection, replacing the LM reader by a rule-ID MLP costs 1.3, 3.4, and 6.9 AUPRC points on P2012, MIMIC-III, and P2019.

\begin{table}[!htbp]
\centering
\caption{Prediction against numerical models. Test mean $\pm$ standard deviation.}
\label{tab:numerical}

\fontsize{7.5}{9}\selectfont
\setlength{\tabcolsep}{1.0pt}
\renewcommand{\arraystretch}{1.00}

\begin{tabular*}{\linewidth}{@{\extracolsep{\fill}}l*{6}{c}@{}}
\toprule
Method
& \multicolumn{2}{c}{P2012}
& \multicolumn{2}{c}{MIMIC-III}
& \multicolumn{2}{c}{P2019}
\\
\cmidrule(lr){2-3}
\cmidrule(lr){4-5}
\cmidrule(lr){6-7}
& AUROC & AUPRC
& AUROC & AUPRC
& AUROC & AUPRC
\\
\midrule

\textbf{STEP-CTS (Ours)}
& \metric{0.8561}{0.002}
& \metric{0.5274}{0.005}
& \textbf{\underline{\metric{0.8563}{0.002}}}
& \textbf{\underline{\metric{0.5025}{0.006}}}
& \metric{0.9115}{0.002}
& \textbf{\underline{\metric{0.5004}{0.016}}}
\\

LOCF LSTM
& \metric{0.8192}{0.004}
& \metric{0.4548}{0.015}
& \metric{0.8240}{0.002}
& \metric{0.4811}{0.011}
& \metric{0.8514}{0.004}
& \metric{0.2392}{0.006}
\\

LOCF Transformer
& \metric{0.8197}{0.004}
& \metric{0.4327}{0.004}
& \metric{0.8230}{0.000}
& \metric{0.4465}{0.008}
& \metric{0.8741}{0.004}
& \metric{0.3296}{0.010}
\\

GRU-D
& \metric{0.8595}{0.003}
& \metric{0.5383}{0.008}
& \metric{0.8541}{0.002}
& \metric{0.5008}{0.010}
& \metric{0.9134}{0.003}
& \metric{0.4637}{0.016}
\\

STraTS
& \underline{\metric{0.8702}{0.004}}
& \textbf{\underline{\metric{0.5543}{0.009}}}
& \metric{0.8459}{0.008}
& \metric{0.4705}{0.022}
& \underline{\metric{0.9174}{0.000}}
& \metric{0.4579}{0.003}
\\

mTAND Transformer
& \metric{0.8397}{0.003}
& \metric{0.5100}{0.004}
& \metric{0.8399}{0.002}
& \metric{0.4603}{0.007}
& \metric{0.8381}{0.007}
& \metric{0.2629}{0.010}
\\

SeFT
& \metric{0.8278}{0.007}
& \metric{0.4646}{0.009}
& \metric{0.8146}{0.008}
& \metric{0.4034}{0.012}
& \metric{0.9041}{0.002}
& \metric{0.3969}{0.010}
\\

KEDGN
& \metric{0.8661}{0.004}
& \metric{0.5374}{0.007}
& \metric{0.8425}{0.009}
& \metric{0.4734}{0.012}
& \metric{0.9084}{0.005}
& \metric{0.3906}{0.004}
\\

DuETT
& \metric{0.8672}{0.002}
& \underline{\metric{0.5420}{0.006}}
& \metric{0.8460}{0.004}
& \metric{0.4576}{0.007}
& \metric{0.9033}{0.003}
& \metric{0.4368}{0.030}
\\

GARLIC
& \metric{0.8596}{0.006}
& \metric{0.5289}{0.013}
& \metric{0.8530}{0.002}
& \underline{\metric{0.5009}{0.012}}
& \metric{0.9088}{0.003}
& \metric{0.4464}{0.011}
\\

\midrule
\multicolumn{7}{@{}l}{\textit{Gradient boosting on the same window statistics}}\\

XGB-WindowStats, 30 rules
& \metric{0.8482}{0.0022}
& \metric{0.5002}{0.0046}
& \metric{0.8391}{0.0005}
& \metric{0.4290}{0.0106}
& \metric{0.9051}{0.0011}
& \metric{0.4208}{0.0006}
\\

XGB-WindowStats, tuned
& \textbf{\underline{\metric{0.8716}{0.0005}}}
& \metric{0.5371}{0.0018}
& \underline{\metric{0.8552}{0.0006}}
& \metric{0.4897}{0.0125}
& \textbf{\underline{\metric{0.9279}{0.0002}}}
& \underline{\metric{0.4802}{0.0034}}
\\

\bottomrule
\end{tabular*}
\Note{Bold and underline: best, underline: second-best across all rows. \textsc{XGB-WindowStats} rows are same-feature controls on the statistics that \textsc{TEV} verbalizes, not dedicated time-series models: the $n=30$ row reads as many rules per patient as the $K=30$ evidence budget; the cap-tuned row selects only the tree cap by validation AUROC (other hyperparameters fixed; Appendix~\ref{app:xgb}) and reads 100--200 rules per patient.}
\end{table}

\paragraph{Prediction against dedicated numerical models.}
Among all rows of Table~\ref{tab:numerical}, STEP-CTS has the highest mean AUPRC on MIMIC-III and P2019 and the highest mean AUROC on MIMIC-III, with the MIMIC-III margins over GRU-D and GARLIC within one standard deviation; on P2012 it trails STraTS by 2.7 AUPRC points and GRU-D, KEDGN, and DuETT by 1.0--1.5. The \textsc{XGB-WindowStats} rows are a same-feature control: a cap of $n$ trees means $n$ activated rules per patient, so the budget is matched on what is read. At every budget from 5 to 30 rules, STEP-CTS matches or exceeds this control on all three datasets, by 2.7, 7.4, and 8.0 points at 30 (Figure~\ref{fig:paths_auprc}). Without a budget, the validation-tuned control reads 100--200 rules per patient and surpasses STEP-CTS on P2012 and on P2019 AUROC (Appendix~\ref{app:xgb}). Thus, STEP-CTS is the more accurate reader of a bounded 30-rule set, while an unbounded ensemble can exceed it at three to seven times the evidence.

\paragraph{Explanation quality.}
The clinician study places STEP-CTS at the intersection of the two existing interface types (Table~\ref{tab:explanation}). On source traceability it matches Decode (4.92 vs.\ 4.89), and both clearly exceed Record2Vec: both clinicians reported that its narrative often left the time window behind a statement unstated, and one noted that it sometimes asserted interventions or diagnoses that could not be confirmed in the record (Appendices~\ref{app:qual_feedback1} and~\ref{app:qual_feedback2}). On clinical interpretation and usefulness, both clinicians rate STEP-CTS above Decode (Table~\ref{tab:explanation_r1}(a,b)); Decode leaves clinical meaning to the reader (Appendix~\ref{app:qual_feedback1}). SHAP-to-Text, a post-hoc attribution over \textsc{XGB-WindowStats}, is rated below STEP-CTS on all three scales by both clinicians: it shows which features a separate predictor weighted, not what the predictor read. Against Record2Vec the two clinicians disagree on the ordering: the narrative reads more naturally and conveys the trajectory better, whereas our evidence can be redundant across windows and attach concepts at marginal cutoffs (Appendix~\ref{app:case_study}), so we treat the two as comparable. STEP-CTS is nonetheless the only presentation that both clinicians and both LLM judges place in the top two on every scale, although they disagree on whether STEP-CTS or Record2Vec ranks higher on interpretation and usefulness (Table~\ref{tab:llm_judge}).

\begin{table}[!htbp]
\centering
\caption{Clinician explanation evaluation on 50 P2012 test patients. All scales range from 1 to 5.}
\label{tab:explanation}

\small
\setlength{\tabcolsep}{1.0pt}
\renewcommand{\arraystretch}{0.9}

\begin{tabular*}{\linewidth}{@{\extracolsep{\fill}}l*{3}{c}@{}}
\toprule
Method
& \shortstack{Source\\traceability}
& \shortstack{Clinical interpretation\\quality}
& \shortstack{Clinical\\usefulness}
\\
\midrule

Record2Vec
& \metric{3.13}{0.72}
& \underline{\metric{3.38}{0.65}}
& \textbf{\underline{\metric{4.17}{0.68}}}
\\

Decode
& \underline{\metric{4.89}{0.42}}
& \metric{2.28}{1.44}
& \metric{2.18}{1.14}
\\

SHAP-to-Text
& \metric{4.72}{0.53}
& \metric{3.29}{0.81}
& \metric{2.93}{0.73}
\\

\textbf{Ours} (concept-grouped)
& \textbf{\underline{\metric{4.92}{0.27}}}
& \textbf{\underline{\metric{4.13}{0.84}}}
& \underline{\metric{3.86}{0.79}}
\\

\bottomrule
\end{tabular*}

\Note{
Two clinicians rated four presentations for every patient (mean $\pm$ sample standard deviation).
\textbf{\underline{Bold and underlined}}: best mean;
\underline{underlined}: second-best mean among the displayed rows.
Method names and outcomes were hidden;
each presentation showed its own model probability and validation best-F1 threshold.
Reviewer-specific results and agreement are in Table~\ref{tab:explanation_r1}.}
\end{table}

\begin{figure}[!htbp]
\centering
\includegraphics[width=0.90\linewidth]{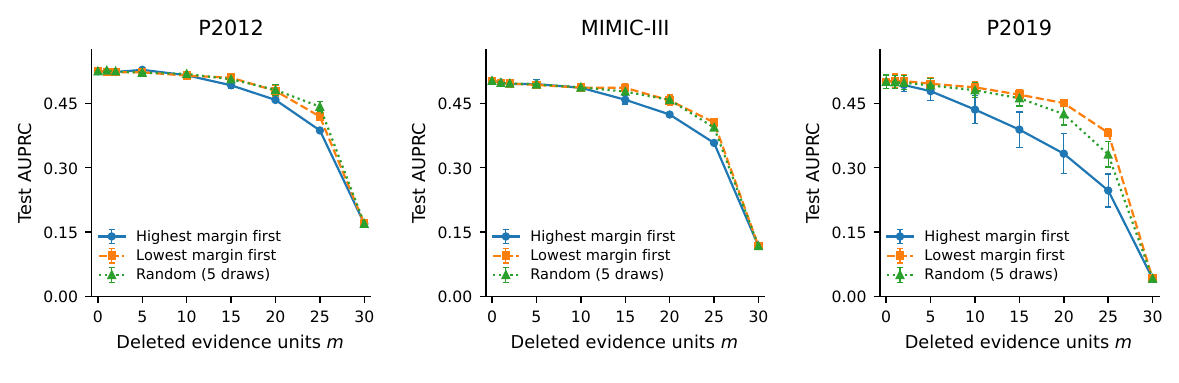}
\caption{Prediction reliance under evidence deletion. With $K=30$ fixed, evidence is removed by descending, ascending, or random ES margin. $m=30$ uses statics alone.}
\label{fig:selector_behavior}
\label{fig:deletion}
\end{figure}

\paragraph{Prediction reliance.}
We use $K=30$ throughout; the matched train--test budget sweep is in Appendix~\ref{app:budget_sweep}. To test whether selector scores track predictive value, we delete selected evidence at test time in descending, random, or ascending score order without re-selection, following the comprehensiveness test of \citet{deyoung2020eraser} (Figure~\ref{fig:deletion}; Appendix~\ref{app:deletion}). Descending deletion hurts most beyond 10--15 removed units and increasingly thereafter, so the ranking places more predictive evidence earlier; yet all three orders degrade with $m$, so the signal is spread across the selected set rather than concentrated in a few units. Deleting everything falls to the statics-only level near the base rate, confirming that the prediction is carried by the selected evidence (Appendix~\ref{app:deletion}).

\section{Conclusion}
We presented STEP-CTS, a text interface for irregular clinical time series that verbalizes multi-scale threshold rules as deterministic, source-traceable predicates, attaches clinical concepts only where an offline annotator can justify them, and learns jointly with the LM reader which evidence units to present. Across three ICU benchmarks it outperforms all evaluated text interfaces and matches or approaches dedicated numerical models, with no patient-level generation and inference in milliseconds rather than the seconds required by summarization interfaces. The evidence a clinician inspects is a deterministic regrouping of the same units, with no additional generation, and blinded clinicians rated it above generated summaries on clinical interpretation. The evidence is source-traceable rather than causal, and the clinician evaluation is limited in scale; we discuss these limitations in Appendix~\ref{app:limitations}.

\subsection*{AI use statement}

Generative models are components of this work and are described in the paper: a frozen Qwen3.5-27B annotator attaches clinical concepts to threshold predicates offline (Section~\ref{sec:method}, Appendix~\ref{app:concept_prompt}), Qwen3-Embedding-8B provides predicate embeddings, the reproduced baselines use the generators specified by their original papers, and GPT-5.5 and Claude Sonnet 5 serve as automated raters corroborating the clinician study (Appendix~\ref{app:llm_judge}). Beyond these, we used generative AI tools for drafting and restructuring parts of the manuscript, suggesting candidate related work, and drafting the initial cutoff inventory described in Appendix~B.1, which was verified by the authors and fixed before experiments. We did not use generative AI to generate synthetic data. We take responsibility for the final content of this work, including text, claims, or artifacts produced with the aid of generative AI.

\clearpage
\bibliography{iclr2027_conference}
\bibliographystyle{iclr2027_conference}
\clearpage

\appendix

\counterwithin*{table}{section}
\counterwithin*{figure}{section}

\renewcommand{\thetable}{\thesection\arabic{table}}
\renewcommand{\thefigure}{\thesection\arabic{figure}}

\renewcommand{\theHtable}{appendix.\thesection.\arabic{table}}
\renewcommand{\theHfigure}{appendix.\thesection.\arabic{figure}}

\section{Dataset protocols and cohorts}
\label{app:data}\label{app:benchmarks}
Table~\ref{tab:datasets}(a--c) specifies the tasks and observation windows, shared cohorts, and exclusions. P2012 uses an outcome-stratified split (seed 145469037); P2019 uses a patient-level split stratified by ever-sepsis (seed 16742963). MIMIC-III follows the benchmark train/test split and validation subset of \citet{harutyunyan2019multitask}. All methods use the same post-exclusion cohorts.
\begin{table}[!htbp]
\centering
\caption{Dataset protocols and cohorts: (a) task definitions and splits.}
\label{tab:datasets}
\footnotesize
\setlength{\tabcolsep}{1.0pt}\renewcommand{\arraystretch}{1.16}
\begin{tabularx}{\linewidth}{@{}>{\hsize=0.3300\hsize\linewidth=\hsize\raggedright\arraybackslash}X>{\hsize=0.5700\hsize\linewidth=\hsize\raggedright\arraybackslash}X>{\hsize=2.1000\hsize\linewidth=\hsize\raggedright\arraybackslash}X@{}}
\toprule
Dataset & Field & Specification\\
\midrule
P2012 & Preprocessing & PhysioNet/CinC 2012 sets A+B+C via the medical datasets extraction of \citet{horn2020seft}; minimum observation=5\\
 & Target & In-hospital death (Outcomes-*.txt)\\
 & Observation / horizon & First 48 h of the ICU stay $\rightarrow$ death during the hospital stay\\
 & Split & Stratified by outcome: 80/20 train+val / test, then 80/20 train / val (seed 145469037)\\
MIMIC-III & Preprocessing & MIMIC-III benchmark in-hospital-mortality episodes \citep{harutyunyan2019multitask} via the medical datasets reader; minimum observation=5\\
 & Target & In-hospital mortality\\
 & Observation / horizon & First 48 h of the ICU stay $\rightarrow$ death during the hospital stay\\
 & Split & Benchmark train/test listfiles; val = the benchmark's validation subset of train\\
P2019 & Preprocessing & PhysioNet/CinC 2019 (all 40,333 records) via medical datasets; Raindrop cohort filter (Table~\ref{tab:exclusions}(c))\\
 & Target & Challenge SepsisLabel at the prediction point, i.e. sepsis onset within the next 6 h\\
 & Observation / horizon & Record up to 60 h $\rightarrow$ sepsis within 6 h\\
 & Split & Patient-level, stratified by ever-sepsis: 80/20 train+val / test, then 80/20 train / val (seed 16742963)\\
\bottomrule\end{tabularx}
\end{table}
\begin{table}[!htbp]
\ContinuedFloat
\centering
\caption{Dataset protocols and cohorts (continued): (b) post-exclusion cohorts.}
\label{tab:cohorts}
\footnotesize
\setlength{\tabcolsep}{1.0pt}\renewcommand{\arraystretch}{1.16}
\begin{tabularx}{\linewidth}{@{}>{\hsize=1.0645\hsize\linewidth=\hsize\raggedright\arraybackslash}X>{\hsize=1.0645\hsize\linewidth=\hsize\raggedright\arraybackslash}X>{\hsize=0.9677\hsize\linewidth=\hsize\raggedright\arraybackslash}X>{\hsize=0.9677\hsize\linewidth=\hsize\raggedright\arraybackslash}X>{\hsize=0.9677\hsize\linewidth=\hsize\raggedright\arraybackslash}X>{\hsize=0.9677\hsize\linewidth=\hsize\raggedright\arraybackslash}X@{}}
\toprule
Dataset & Split & Patients & Positives & Negatives & Positive rate\\
\midrule
P2012 & train & 7,670 & 1,093 & 6,577 & 14.25\%\\
 & validation & 1,916 & 273 & 1,643 & 14.25\%\\
 & test & 2,400 & 341 & 2,059 & 14.21\%\\
MIMIC-III & train & 14,632 & 1,978 & 12,654 & 13.52\%\\
 & validation & 3,208 & 435 & 2,773 & 13.56\%\\
 & test & 3,217 & 374 & 2,843 & 11.63\%\\
P2019 & train & 24,817 & 1,014 & 23,803 & 4.09\%\\
 & validation & 6,217 & 254 & 5,963 & 4.09\%\\
 & test & 7,766 & 325 & 7,441 & 4.18\%\\
\bottomrule\end{tabularx}
\end{table}
\begin{table}[!htbp]
\ContinuedFloat
\centering
\caption{Dataset protocols and cohorts (continued): (c) shared exclusions.}
\label{tab:exclusions}
\footnotesize
\setlength{\tabcolsep}{1.0pt}\renewcommand{\arraystretch}{1.16}
\begin{tabularx}{\linewidth}{@{}>{\hsize=1.3800\hsize\linewidth=\hsize\raggedright\arraybackslash}X>{\hsize=0.4800\hsize\linewidth=\hsize\raggedright\arraybackslash}X>{\hsize=1.1400\hsize\linewidth=\hsize\raggedright\arraybackslash}X@{}}
\toprule
Criterion & Affected datasets & Excluded patients\\
\midrule
Severe observation sparsity: $\leq$4 raw time points, so no valid feature under minimum observation=5 & P2012, MIMIC-III & P2012 2 (train 1, val 1); MIMIC-III 50 (train 30, val 9, test 11)\\
Task-cohort misalignment: first measurement at hour $\geq$269, outside the 60 h sepsis window & P2019 & 3\\
Raindrop \citep{zhang2022raindrop} short/long-record filter - cohort definition only, not their splits & P2019 & 1,530 (P2019 total with the row above: train 996, val 237, test 300)\\
Total &  & 1,585\\
\bottomrule\end{tabularx}
\Note{Exclusions are applied identically to all pipelines and total 1,585 patients.}
\end{table}

\FloatBarrier

\section{Implementation and evidence specification}
\label{app:impl}\label{app:numeric_full}\label{app:selector_metadata}
Table~\ref{tab:implementation}(a--c) specifies the evaluated text interfaces. Tables~\ref{tab:training_settings0}(a,b) and~\ref{tab:numerical_lr} give training settings and numerical-baseline learning rates. Table~\ref{tab:selector_metadata} defines the selector inputs, Table~\ref{tab:evidence_schema} records the evidence schema, and Figure~\ref{fig:evidence_example} illustrates the chronological $P+C$ reader input. The default pipeline uses a frozen rule inventory and cached annotations; ES and the LM reader are trained jointly.

\paragraph{TimeCP instantiation.}
TimeCP is the contextualize-and-predict component of TimeCAP \citep{lee2025timecp}. The faithful variant uses a contextualizer and an LLM predictor; the adapted variant retains the report and substitutes BioClinical ModernBERT. Table~\ref{tab:implementation_interfaces_2}(b) specifies the shared generator used in this reproduction.

\begin{table}[!htbp]
\centering
\caption{Instantiated text interfaces: (a) WRDP and Record2Vec.}
\label{tab:implementation}
\footnotesize
\setlength{\tabcolsep}{1.0pt}\renewcommand{\arraystretch}{1.16}
\begin{tabularx}{\linewidth}{@{}>{\hsize=0.3300\hsize\linewidth=\hsize\raggedright\arraybackslash}X>{\hsize=0.6900\hsize\linewidth=\hsize\raggedright\arraybackslash}X>{\hsize=1.9800\hsize\linewidth=\hsize\raggedright\arraybackslash}X@{}}
\toprule
Method & Field & Specification\\
\midrule
WRDP & Input representation & Hourly value-change narrative of the raw record\\
 & Generator & rule-based narrative\\
 & Embedding model & Frozen Llama-3-8B-Instruct, mean-pooled hidden state\\
 & Faithful predictor & XGBoost on the pooled embedding\\
 & Adapted predictor & BioClinical ModernBERT fine-tune on the narrative, max 3072 tokens\\
 & Static-prefix policy & [Patient] \ldots{} line prepended to the text\\
Record2Vec & Input representation & Released-code serialization: hour-averaged X has value v in hour h, X has no values\\
 & Generator & Qwen3.5-27B summary with the released ICD prompt\\
 & Embedding model & Frozen Qwen3-Embedding-8B\\
 & Faithful predictor & PatchTSMixer (batch 512, early stopping)\\
 & Adapted predictor & BioClinical ModernBERT on the summary, max 3072 tokens\\
 & Static-prefix policy & [Patient] \ldots{} line prepended to the text\\
\bottomrule\end{tabularx}
\end{table}
\begin{table}[!htbp]
\ContinuedFloat
\centering
\caption{Instantiated text interfaces (continued): (b) Decode and TimeCP.}
\label{tab:implementation_interfaces_2}
\footnotesize
\setlength{\tabcolsep}{1.0pt}\renewcommand{\arraystretch}{1.16}
\begin{tabularx}{\linewidth}{@{}>{\hsize=0.3300\hsize\linewidth=\hsize\raggedright\arraybackslash}X>{\hsize=0.6900\hsize\linewidth=\hsize\raggedright\arraybackslash}X>{\hsize=1.9800\hsize\linewidth=\hsize\raggedright\arraybackslash}X@{}}
\toprule
Method & Field & Specification\\
\midrule
Decode & Input representation & 6-h per-event verbalization (technical style)\\
 & Generator & ---\\
 & Embedding model & ---\\
 & Faithful predictor & Llama-3-8B-Instruct + LoRA (r 64, $\alpha$ 32) trained on context and response, $\leq$5 epochs, score = first-token P(Yes)\\
 & Adapted predictor & BioClinical ModernBERT on the verbalization, max 3072 tokens\\
 & Static-prefix policy & [Patient] \ldots{} line prepended to the text\\
TimeCP & Input representation & 6-h per-event verbalization\\
 & Generator & Qwen3.5-27B contextualizer: $\leq$5-sentence report (TimeCAP prompt, domain words set to ICU)\\
 & Embedding model & ---\\
 & Faithful predictor & Qwen3.5-27B predictor prompt answers yes / no (temperature 0.7); no training\\
 & Adapted predictor & BioClinical ModernBERT on the report, max 3072 tokens\\
 & Static-prefix policy & [Patient] \ldots{} line prepended to the text\\
\bottomrule\end{tabularx}
\end{table}
\begin{table}[!htbp]
\ContinuedFloat
\centering
\caption{Instantiated text interfaces (continued): (c) STEP-CTS.}
\label{tab:implementation_interfaces_4}
\footnotesize
\setlength{\tabcolsep}{1.0pt}\renewcommand{\arraystretch}{1.16}
\begin{tabularx}{\linewidth}{@{}>{\hsize=0.3300\hsize\linewidth=\hsize\raggedright\arraybackslash}X>{\hsize=0.6900\hsize\linewidth=\hsize\raggedright\arraybackslash}X>{\hsize=1.9800\hsize\linewidth=\hsize\raggedright\arraybackslash}X@{}}
\toprule
Method & Field & Specification\\
\midrule
Ours & Input representation & Source-linked units $(z,P,C)$ from the fixed window-specific rule inventories\\
 & Generator & Qwen3.5-27B offline concept annotator per rule (no task, outcome or leaf statistic given; abstention allowed)\\
 & Embedding model & Qwen3-Embedding-8B (4,096-d) of the predicate text --- selector input only\\
 & Faithful predictor & BioClinical ModernBERT reader over the K=30 ES-selected units (max\_length 3072)\\
 & Adapted predictor & --- (Ours has one predictor)\\
 & Static-prefix policy & [Patient] \ldots{} line prepended to the reader input\\
\bottomrule\end{tabularx}
\Note{All adapted predictors use batch size 8, 30 epochs, and learning rate, checkpoint selection by validation AUROC. maxlength 3072 exceeds the longest training text of every interface, so no input is truncated.}
\end{table}
\begin{table}[!htbp]
\centering
\caption{Training settings: (a) concept construction, trees, and candidate pools.}
\label{tab:training_settings0}
\footnotesize
\setlength{\tabcolsep}{1.0pt}\renewcommand{\arraystretch}{1.16}
\begin{tabularx}{\linewidth}{@{}>{\hsize=0.6000\hsize\linewidth=\hsize\raggedright\arraybackslash}X>{\hsize=1.6800\hsize\linewidth=\hsize\raggedright\arraybackslash}X>{\hsize=0.7200\hsize\linewidth=\hsize\raggedright\arraybackslash}X@{}}
\toprule
Component / parameter & Rule and selected values & Checkpoint / seeds\\
\midrule
TEV concepts / Annotator and cutoff check & Fixed: Qwen3.5-27B, clause-list prompt without task / outcome / statistics, abstain when no supported cutoff; a concept is kept only if its cited clauses reach the finding's cut-off ; one generation per unique rule; concepts dropped 256 / 1,048 / 573 (P2012 / MIMIC-III / P2019) & ---; seeds: single generation\\
Trees / Windows, statistics & Fixed: W $\in$ \{1, 2, 4, 8, 16, 32, 48\} h; 9 statistics per variable (last, mean, min, max, std, count, $\Delta$, gap, missing); same for all & ---; seeds: single fit\\
Trees / XGBoost per window & Fixed: up to 30 trees, max\_depth 5, subsample 0.8, colsample 0.8, min\_child\_weight 5, scale\_pos\_weight, early stopping 10 rounds on validation; same for all & best iteration on validation ($\le$30 trees); seeds: single fit\\
Trees / Candidate pool & Fixed: top-5 leaves per (window, time step) $\rightarrow$ merged units (M=5); M=30 for the selection rules; \textasciitilde{}150 units / patient (M=5), \textasciitilde{}1,200 (M=30) & ---; seeds: ---\\
\bottomrule\end{tabularx}
\end{table}
\begin{table}[!htbp]
\ContinuedFloat
\centering
\caption{Training settings (continued): (b) selector, reader, and numerical baselines.}
\label{tab:training_settings4}
\footnotesize
\setlength{\tabcolsep}{1.0pt}\renewcommand{\arraystretch}{1.16}
\begin{tabularx}{\linewidth}{@{}>{\hsize=0.6000\hsize\linewidth=\hsize\raggedright\arraybackslash}X>{\hsize=1.6800\hsize\linewidth=\hsize\raggedright\arraybackslash}X>{\hsize=0.7200\hsize\linewidth=\hsize\raggedright\arraybackslash}X@{}}
\toprule
Component / parameter & Rule and selected values & Checkpoint / seeds\\
\midrule
ES / Selector architecture & Predicate-embedding projection $4096\!\rightarrow\!256\!\rightarrow\!64$, token projection, learnable CLS token, 2-layer Transformer encoder, and linear selection head; 8 scalar metadata features & Fixed across datasets; seeds: as LM\\

ES / Selector objective & Self-critical policy-gradient loss with $\lambda_{\mathrm{sel}}=0.1$ and entropy regularization with $\lambda_{\mathrm{ent}}=0.01$; evidence budget enforced by top-$K$ with $K=30$ & Fixed across datasets; seeds: as LM\\

ES / Selector learning rate & $1\times10^{-4}$ & Fixed across datasets; seeds: as LM\\
LM (reader) / Backbone & Fixed: BioClinical ModernBERT-base, max\_length 3072, batch 8, weight decay 1e-4; same for all & ---; seeds: ---\\
LM (reader) / Learning rate & \{3e-5 \textasciitilde{} 3e-6\} by mean best validation AUROC; P2012 3e-6, MIMIC-III 1e-5, P2019 3e-6 & epoch with the best validation AUROC; seeds: P2012 0--2; MIMIC-III 0--2; P2019 0--2\\
LM (reader) / Epochs & Fixed per dataset; P2012 30, MIMIC-III 30, P2019 30 epoch schedule & same; seeds: same\\
Numerical baselines / Learning rate & \{1e-3 - 1e-6\} by validation AUROC; selected values in Table~\ref{tab:numerical_lr} & best validation AUROC; seeds: 3\\
\bottomrule\end{tabularx}
\end{table}
\begin{table}[!htbp]
\centering
\caption{Selected numerical-baseline learning rates.}
\label{tab:numerical_lr}
\footnotesize
\setlength{\tabcolsep}{1.0pt}\renewcommand{\arraystretch}{1.16}
\begin{tabularx}{\linewidth}{@{}>{\hsize=1.6000\hsize\linewidth=\hsize\raggedright\arraybackslash}X>{\hsize=0.8000\hsize\linewidth=\hsize\raggedright\arraybackslash}X>{\hsize=0.8000\hsize\linewidth=\hsize\raggedright\arraybackslash}X>{\hsize=0.8000\hsize\linewidth=\hsize\raggedright\arraybackslash}X@{}}
\toprule
Numerical model & P2012 & MIMIC-III & P2019\\
\midrule
LOCF LSTM & $10^{-4}$ & $10^{-4}$ & $10^{-5}$\\
LOCF Transformer & $10^{-4}$ & $10^{-4}$ & $10^{-5}$\\
GRU-D & $10^{-4}$ & $10^{-4}$ & $10^{-4}$\\
STraTS & $10^{-5}$ & $10^{-4}$ & $10^{-4}$\\
mTAND Transformer & $10^{-3}$ & $10^{-4}$ & $10^{-4}$\\
SeFT & $10^{-3}$ & $10^{-3}$ & $10^{-3}$\\
KEDGN & $10^{-3}$ & $10^{-3}$ & $10^{-3}$\\
DuETT & $10^{-3}$ & $10^{-4}$ & $10^{-4}$\\
GARLIC & $10^{-3}$ & $10^{-4}$ & $10^{-5}$\\
\bottomrule\end{tabularx}
\Note{The validation-AUROC search sweep range is $\{10^{-3} - 10^{-6}\}$.}
\end{table}

\begin{table}[!htbp]
\centering
\caption{Selector metadata, in input order.}
\label{tab:selector_metadata}
\footnotesize
\setlength{\tabcolsep}{1.0pt}\renewcommand{\arraystretch}{1.16}
\begin{tabularx}{\linewidth}{@{}>{\hsize=0.5000\hsize\linewidth=\hsize\raggedright\arraybackslash}X>{\hsize=1.5000\hsize\linewidth=\hsize\raggedright\arraybackslash}X@{}}
\toprule
Name & Exact formula\\
\midrule
recency\_score & max(0, 1 - recency / 48), recency = hours from the unit's window end to the prediction point\\
xgb\_score & score of the fired leaf in the XGBoost model of window W\\
abs\_dir & |p\_leaf - p0|, p\_leaf = training positive rate at the leaf, p0 = cohort positive rate\\
dir & p\_leaf - p0\\
log1p\_n & log(1 + n\_samples), n\_samples = training samples reaching the leaf\\
w\_norm & W / 48, W = window length in hours\\
log1p\_ri\_count & log(1 + subword count of the predicate text)\\
ci & 1 if the leaf keeps $\geq$1 concept, else 0\\
\bottomrule\end{tabularx}
\end{table}

\begin{table}[!htbp]
\centering
\caption{Evidence schema for the default $e=(z,P,C)$ interface; $A$ is an optional ablation field.}
\label{tab:evidence_schema}
\footnotesize
\setlength{\tabcolsep}{1.0pt}\renewcommand{\arraystretch}{1.16}
\begin{tabularx}{\linewidth}{@{}>{\hsize=0.2800\hsize\linewidth=\hsize\raggedright\arraybackslash}X>{\hsize=1.7200\hsize\linewidth=\hsize\raggedright\arraybackslash}X@{}}
\toprule
Field & Construction and example\\
\midrule
Source tuple $z$ & $(t_{\mathrm{start}},t_{\mathrm{end}},W,r)$: source interval, window size, and rule identifier $r$ (window-specific tree/leaf identity); consecutive activations of the same rule are merged\par\smallskip\textit{Example:} [15.4-47.4 h | W=32 h]\\
P & Path predicates rendered as clauses on named statistics (GCS min less than 8.0); a clause met only by a missingness fill is removed and its variable listed once as not measured: \ldots{}\par\smallskip\textit{Example:} GCS last less than 8.00, static Age less than 69, Glucose mean at least 144.50, and MAP min less than 59.00\\
C & Concept names the generator attached to the leaf, after the cutoff check\par\smallskip\textit{Example:} concept: impaired consciousness; hyperglycemia.\\
A & target association: +x pp (p\_leaf vs p0 base); support s\%, leave-one-patient-out smoothed ($\alpha$ = 10) on train; val/test never see a label\par\smallskip\textit{Example:} target association: +22.8pp; support 2\%.\\
\bottomrule\end{tabularx}
\Note{The Ours reader uses $P+C$ and a static prefix; $A$ appears only in $P+A$ and $P+C+A$.}
\end{table}

\begin{figure}[!htbp]
\centering
\begin{minipage}{\linewidth}\small\ttfamily\raggedright
[Patient] Age: 56, Gender: Male, Height: 172.7, ICUType: SICU\par\smallskip
[15.4-47.4 h] GCS last less than 8.00, static Age less than 69, Glucose mean at least 144.50, and MAP min less than 59.00. concept: impaired consciousness; hyperglycemia.
\end{minipage}
\caption{Reader-input example ($P+C$) for P2012 test patient 1000. A static prefix is followed by a source interval, deterministic threshold predicates, and available clinical concepts. The rule identifier and concept-support indices are stored as provenance rather than printed in this input.}
\label{fig:evidence_example}
\end{figure}
\FloatBarrier

\subsection{Concept annotation and the clause-cutoff check}
\label{app:concept_prompt}\label{app:gloss_prompt}
Concepts are produced in two steps. \emph{Generation} is the only LLM step: Qwen3.5-27B receives the numbered predicate clauses of one rule and the dataset's variable reference (units and scales), but no patient record, task description, outcome, rule statistic, or clinical threshold; it names the clinical states the clauses entail from its own knowledge, cites the minimal clause set for each, may assert normality only on scales whose maximum is the healthy value (GCS and its components), and is instructed to abstain when unsure. One call per unique rule at temperature 0 (prompt verbatim in Appendix~\ref{app:prompts}). \emph{The clause-cutoff check} is deterministic and calls no LLM: for each concept in Table~\ref{tab:cutoffs}, the concept is kept only if one cited clause on the listed quantity, using a level statistic (last, mean, min, or max), reaches the cut-off; it is dropped if the cited clauses on that quantity cannot reach the cut-off or none of them concerns that quantity. Concepts outside the table pass unchanged. The check removed 256, 1,048, and 573 annotations on P2012, MIMIC-III, and P2019. The table was drafted with the annotator model and fixed in code before any experiment; guideline-defined entries were verified by the authors against the cited source, and the remaining entries are the conventional adult limits. The full-labeling control removes abstention and the task-aware control adds the task description; both retain the same check (Table~\ref{tab:concept_inventory_controls}). The entailment audit evaluates the retained annotations independently (Appendix~\ref{app:concept_entailment}).
\begin{table}[!htbp]
\centering
\caption{Cut-offs applied by the clause-cutoff check. A concept is kept only if one cited clause on the listed quantity reaches the cut-off.}
\label{tab:cutoffs}
\footnotesize
\setlength{\tabcolsep}{1.0pt}\renewcommand{\arraystretch}{1.16}
\begin{tabularx}{\linewidth}{@{}>{\hsize=0.95\hsize\linewidth=\hsize\raggedright\arraybackslash}X>{\hsize=0.80\hsize\linewidth=\hsize\raggedright\arraybackslash}X>{\hsize=1.25\hsize\linewidth=\hsize\raggedright\arraybackslash}X@{}}
\toprule
Concept & Cut-off & Source\\
\midrule
tachycardia / bradycardia & HR $>100$ / $<60$ bpm & standard adult reference range\\
tachypnea / bradypnea & RR $>20$ / $<12$ /min & standard adult reference range; RR $>20$ is also a SIRS criterion \citep{bone1992sepsis}\\
hypoxemia & SpO$_2$ $<90\%$ & clinical convention\\
hypotension & SBP $<90$ mmHg or MAP $<65$ mmHg & SBP: \citet{bone1992sepsis}; MAP: \citet{singer2016sepsis3,evans2021ssc}\\
systolic / diastolic hypotension & SBP $<90$ / DBP $<60$ mmHg & SBP: \citet{bone1992sepsis}; DBP: clinical convention\\
hypertension & SBP $\geq 140$ mmHg & \citet{williams2018esc}; ACC/AHA uses $\geq 130$ \citep{whelton2018accaha}\\
hyperglycemia / hypoglycemia & glucose $>140$ / $<70$ mg/dL & \citet{ada2018hospital}\\
acidemia / alkalemia & pH $<7.35$ / $>7.45$ & standard adult reference range\\
hypocapnia / hypercapnia & PaCO$_2$ $<35$ / $>45$ mmHg & standard adult reference range\\
fever & temperature $\geq 38$\,$^\circ$C & SIRS criterion \citep{bone1992sepsis}; the ICU fever guideline uses $\geq 38.3$\,$^\circ$C \citep{ogrady2008fever}\\
thrombocytopenia & platelets $<150\times10^{9}$/L & standard adult reference range\\
\bottomrule
\end{tabularx}
\Note{Rows marked ``standard adult reference range'' or ``clinical convention'' have no single guideline definition; the value is the conventional adult limit. Concepts not in this table (e.g., impaired consciousness, elevated BUN, hyperthermia, hypothermia, leukocytosis, elevated lactate, hyperbilirubinemia) have no fixed cut-off and pass the check unchanged; they are covered only by the annotator's abstention rule and the entailment audit (Appendix~\ref{app:concept_entailment}). The check tests the threshold only: it does not distinguish a bound on every observation (``min higher than $a$'') from a bound on one observation (``max higher than $a$''); that distinction is left to the annotator and audited separately.}
\end{table}

\subsection{Outcome-association controls}
$A$ is an optional field used only in the $P+A$ and $P+C+A$ ablations. It reports training-derived rule support and outcome association, with leave-one-patient-out smoothing ($\alpha=10$) for training examples; validation and test labels are never used to construct it. ES metadata still includes training-derived rule statistics (Table~\ref{tab:selector_metadata}), independently of whether $A$ is shown to the reader.

\section{Numerical controls over the same window statistics}
\label{app:xgb}
We compare STEP-CTS with numerical predictors built from the same multi-scale summary features. \textsc{XGB-WindowStats} fits a single ensemble to the concatenated features, whereas STEP-CTS selects evidence from separate window-specific inventories. Each fitted tree activates one path per patient; the STEP-CTS reader receives up to $K$ selected units. We therefore compare the tree cap $n$ with the evidence budget $K$: the budget is matched on the number of rules a reader consumes per patient, not on inventory size or computation. STEP-CTS's seven window-specific inventories contain up to 210 trees in total, its candidate pool holds several hundred units per patient, and \textsc{ES} scores all of them before retaining $K$. Only the cap is tuned for the control; its remaining hyperparameters (Setup below) are fixed across caps, so ``tuned'' denotes a validation-selected cap, and a fuller search could only strengthen the unbudgeted control.

\paragraph{Setup.}
\textsc{XGB-WindowStats} receives nine statistics per variable over the seven window sizes, plus the same static covariates as STEP-CTS: Age/Gender/Height/ICUType for P2012, Height for MIMIC-III, and Age/Gender/HospAdmTime for P2019. For every cap $n\in\{1,5,10,15,20,25,30,35,40,60,100,200\}$, we refit seeds 0, 1, and 2 with learning rate 0.03, maximum depth 5, minimum child weight 5, row and column subsampling 0.8, class weighting via \texttt{scale\_pos\_weight}, and validation early stopping with patience 10. The cap is an upper bound on the fitted tree count. Table~\ref{tab:xgb} summarizes the controls; Table~\ref{tab:xgb_sweep} reports the full test and validation results. For the pooled window-ensemble controls, we reuse the same seven window-specific XGBoost models used by STEP-CTS and aggregate their per-window prediction probabilities by mean or max pooling; these rows retain the reported single-fit results.

\begin{table}[!htbp]
\centering
\caption{Numerical controls over the same window statistics. STEP-CTS and XGB-WS report test mean $\pm$ standard deviation over three seeds; $^\dagger$ denotes a single fit. XGB-WS abbreviates \textsc{XGB-WindowStats}.}
\label{tab:xgb}

\fontsize{7}{8.5}\selectfont
\setlength{\tabcolsep}{1pt}
\renewcommand{\arraystretch}{1.15}

\begin{tabular*}{\linewidth}{@{\extracolsep{\fill}}l*{6}{c}@{}}
\toprule
Predictor
& \multicolumn{2}{c}{P2012}
& \multicolumn{2}{c}{MIMIC-III}
& \multicolumn{2}{c}{P2019}\\
\cmidrule(lr){2-3}
\cmidrule(lr){4-5}
\cmidrule(lr){6-7}
& AUROC & AUPRC
& AUROC & AUPRC
& AUROC & AUPRC\\
\midrule

\multicolumn{7}{@{}l}{\textit{Matched caps: at most 30 trees or $K=30$ selected units}}\\

XGB-WS, $n=30$
& \metric{0.8482}{0.0022}
& \metric{0.5002}{0.0046}
& \metric{0.8391}{0.0005}
& \metric{0.4290}{0.0106}
& \metric{0.9051}{0.0011}
& \metric{0.4208}{0.0006}\\

STEP-CTS, $P$
& \metric{0.8510}{0.0001}
& \metric{0.4997}{0.0043}
& \metric{0.8503}{0.0015}
& \metric{0.4650}{0.0158}
& \underline{\metric{0.9147}{0.0048}}
& \metric{0.4783}{0.0040}\\

STEP-CTS, $P+C$
& \underline{\metric{0.8561}{0.0017}}
& \underline{\metric{0.5274}{0.0049}}
& \textbf{\underline{\metric{0.8563}{0.0024}}}
& \textbf{\underline{\metric{0.5025}{0.0056}}}
& \metric{0.9115}{0.0016}
& \textbf{\underline{\metric{0.5004}{0.0157}}}\\

\midrule
\multicolumn{7}{@{}l}{\textit{Validation-tuned numerical control}}\\

XGB-WS (tuned)
& \textbf{\underline{\metric{0.8716}{0.0005}}}
& \textbf{\underline{\metric{0.5371}{0.0018}}}
& \underline{\metric{0.8552}{0.0006}}
& \underline{\metric{0.4897}{0.0125}}
& \textbf{\underline{\metric{0.9279}{0.0002}}}
& \underline{\metric{0.4802}{0.0034}}\\

\midrule
\multicolumn{7}{@{}l}{\textit{Pooled window ensembles}}\\

Window ensembles, mean$^\dagger$
& 0.8203
& 0.4328
& 0.8158
& 0.3827
& 0.8144
& 0.1968\\

Window ensembles, max$^\dagger$
& 0.8211
& 0.4682
& 0.7615
& 0.3634
& 0.7773
& 0.1587\\

\bottomrule
\end{tabular*}

\Note{\textbf{Bold} and underline mark the highest mean, and underline alone marks the second-highest mean, in each column. Tuned caps are $n=200$, 100, and 100 on P2012, MIMIC-III, and P2019, respectively, selected by mean validation AUROC. Pooling evaluates approximately 2,000--2,700 paths per patient across the seven window ensembles.}
\end{table}

\paragraph{Matched prediction-input caps.}
At $n=30$ versus $K=30$, STEP-CTS improves mean AUPRC over \textsc{XGB-WindowStats} by 2.72, 7.35, and 7.96 percentage points on P2012, MIMIC-III, and P2019. Pooling the per-window ensembles of STEP-CTS by mean or max (about 2,000--2,700 activated paths per patient, single fit) reaches only 0.43/0.38/0.20 and 0.47/0.36/0.16 AUPRC on P2012/MIMIC-III/P2019.

\paragraph{Validation-selected tree budgets.}
For each dataset, the tuned control uses the cap with the highest mean validation AUROC over seeds 0--2, yielding $n=200$, 100, and 100 for P2012, MIMIC-III, and P2019, respectively. This selects one cap per dataset, not a separate cap or best-performing seed for each run. The tuned control has higher mean test AUROC and AUPRC than STEP-CTS on P2012 and higher mean AUROC on P2019; STEP-CTS has higher mean AUROC and AUPRC on MIMIC-III and higher mean AUPRC on P2019. Across the full sweep, the first tested cap exceeding the default STEP-CTS AUPRC is 100 on P2012 and 200 on P2019; none of the tested caps through 200 exceeds it on MIMIC-III.

\begin{table*}[t]
\centering
\caption{
\textsc{XGB-WindowStats} tree-cap sweep.
Test AUROC/AUPRC and validation AUROC are mean $\pm$ standard deviation over
seeds 0, 1, and 2.
Bold marks the validation-selected cap and its corresponding test results
for each dataset.
STEP-CTS uses $K=30$ and the same three seed indices.
}
\label{tab:xgb_sweep}

\fontsize{6.5}{7.5}\selectfont
\setlength{\tabcolsep}{0.8pt}
\renewcommand{\arraystretch}{1.10}

\begin{adjustbox}{max width=\textwidth,center}
\begin{tabular}{@{}l*{9}{c}@{}}
\toprule
\multirow{2}{*}{Trees $n$}
& \multicolumn{3}{c}{P2012}
& \multicolumn{3}{c}{MIMIC-III}
& \multicolumn{3}{c}{P2019}
\\
\cmidrule(lr){2-4}
\cmidrule(lr){5-7}
\cmidrule(lr){8-10}
& AUROC & AUPRC & Val. AUROC
& AUROC & AUPRC & Val. AUROC
& AUROC & AUPRC & Val. AUROC
\\
\midrule

1
& \metric{0.7698}{0.0114}
& \metric{0.3584}{0.0025}
& \metric{0.7588}{0.0044}
& \metric{0.7859}{0.0037}
& \metric{0.3337}{0.0064}
& \metric{0.7914}{0.0076}
& \metric{0.7995}{0.0003}
& \metric{0.2342}{0.0057}
& \metric{0.7797}{0.0009}
\\

5
& \metric{0.8129}{0.0046}
& \metric{0.4161}{0.0062}
& \metric{0.8158}{0.0036}
& \metric{0.8228}{0.0020}
& \metric{0.3856}{0.0055}
& \metric{0.8264}{0.0034}
& \metric{0.8826}{0.0091}
& \metric{0.3571}{0.0040}
& \metric{0.8602}{0.0022}
\\

10
& \metric{0.8264}{0.0039}
& \metric{0.4467}{0.0047}
& \metric{0.8323}{0.0007}
& \metric{0.8301}{0.0001}
& \metric{0.4016}{0.0004}
& \metric{0.8330}{0.0020}
& \metric{0.8880}{0.0022}
& \metric{0.3742}{0.0050}
& \metric{0.8735}{0.0009}
\\

15
& \metric{0.8378}{0.0006}
& \metric{0.4743}{0.0068}
& \metric{0.8401}{0.0019}
& \metric{0.8349}{0.0014}
& \metric{0.4107}{0.0010}
& \metric{0.8360}{0.0021}
& \metric{0.8953}{0.0015}
& \metric{0.3944}{0.0088}
& \metric{0.8769}{0.0035}
\\

20
& \metric{0.8407}{0.0010}
& \metric{0.4864}{0.0031}
& \metric{0.8422}{0.0019}
& \metric{0.8354}{0.0013}
& \metric{0.4198}{0.0021}
& \metric{0.8388}{0.0013}
& \metric{0.9006}{0.0005}
& \metric{0.4042}{0.0024}
& \metric{0.8821}{0.0010}
\\

25
& \metric{0.8430}{0.0001}
& \metric{0.4955}{0.0040}
& \metric{0.8454}{0.0030}
& \metric{0.8387}{0.0025}
& \metric{0.4308}{0.0079}
& \metric{0.8405}{0.0016}
& \metric{0.9023}{0.0009}
& \metric{0.4087}{0.0023}
& \metric{0.8844}{0.0013}
\\

30
& \metric{0.8482}{0.0022}
& \metric{0.5002}{0.0046}
& \metric{0.8472}{0.0026}
& \metric{0.8391}{0.0005}
& \metric{0.4290}{0.0106}
& \metric{0.8435}{0.0008}
& \metric{0.9051}{0.0011}
& \metric{0.4208}{0.0006}
& \metric{0.8854}{0.0024}
\\

35
& \metric{0.8491}{0.0008}
& \metric{0.4979}{0.0067}
& \metric{0.8500}{0.0024}
& \metric{0.8427}{0.0019}
& \metric{0.4307}{0.0058}
& \metric{0.8454}{0.0015}
& \metric{0.9077}{0.0013}
& \metric{0.4251}{0.0025}
& \metric{0.8868}{0.0019}
\\

40
& \metric{0.8526}{0.0022}
& \metric{0.5024}{0.0077}
& \metric{0.8519}{0.0028}
& \metric{0.8445}{0.0015}
& \metric{0.4346}{0.0070}
& \metric{0.8468}{0.0013}
& \metric{0.9095}{0.0022}
& \metric{0.4305}{0.0017}
& \metric{0.8891}{0.0021}
\\

60
& \metric{0.8565}{0.0002}
& \metric{0.5226}{0.0066}
& \metric{0.8550}{0.0026}
& \metric{0.8495}{0.0008}
& \metric{0.4595}{0.0086}
& \metric{0.8522}{0.0013}
& \metric{0.9150}{0.0005}
& \metric{0.4627}{0.0020}
& \metric{0.8970}{0.0010}
\\

100
& \metric{0.8694}{0.0004}
& \metric{0.5361}{0.0051}
& \metric{0.8616}{0.0011}
& \textbf{\metric{0.8552}{0.0006}}
& \textbf{\metric{0.4897}{0.0125}}
& \textbf{\metric{0.8594}{0.0014}}
& \textbf{\metric{0.9279}{0.0002}}
& \textbf{\metric{0.4802}{0.0034}}
& \textbf{\metric{0.9179}{0.0010}}
\\

200
& \textbf{\metric{0.8716}{0.0005}}
& \textbf{\metric{0.5371}{0.0018}}
& \textbf{\metric{0.8663}{0.0016}}
& \metric{0.8610}{0.0007}
& \metric{0.5010}{0.0054}
& \metric{0.8561}{0.0022}
& \metric{0.9270}{0.0009}
& \metric{0.5202}{0.0008}
& \metric{0.9175}{0.0021}
\\

\midrule
\textbf{STEP-CTS ($K=30$)}
& \metric{0.8561}{0.0017}
& \metric{0.5274}{0.0049}
& ---
& \metric{0.8563}{0.0024}
& \metric{0.5025}{0.0056}
& ---
& \metric{0.9115}{0.0016}
& \metric{0.5004}{0.0157}
& ---
\\

\bottomrule
\end{tabular}
\end{adjustbox}

\Note{
The tree cap is selected separately for each dataset by mean validation AUROC:
$n=200$ for P2012 and $n=100$ for MIMIC-III and P2019.
Bold marks the validation-selected cap and its corresponding test results,
rather than the best test performance.
Each cap is refit under the same recipe with validation early stopping;
$n$ is an upper bound on the number of activated tree paths.
}
\end{table*}

\begin{figure}[!htbp]
\centering
\includegraphics[width=0.323\linewidth]{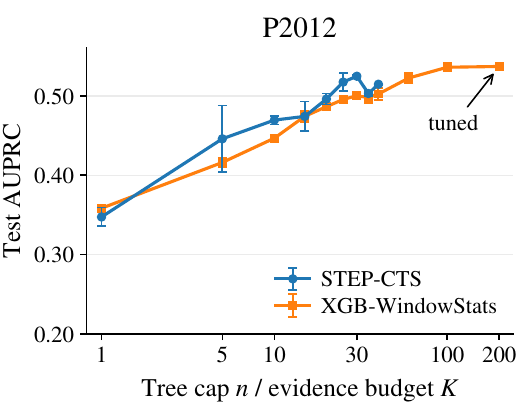}\hfill
\includegraphics[width=0.323\linewidth]{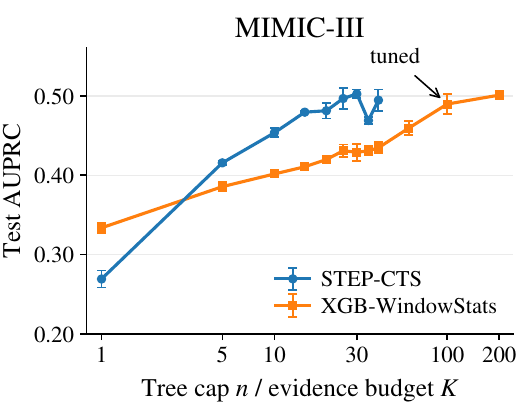}\hfill
\includegraphics[width=0.323\linewidth]{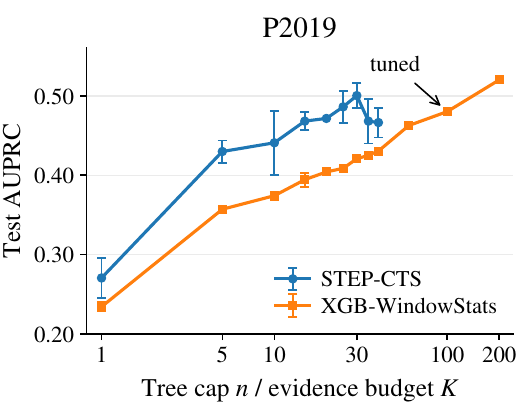}
\caption{Test AUPRC versus the maximum tree count $n$ or selected-evidence budget $K$ (logarithmic horizontal axis). Both series connect the reported means; error bars show across-seed standard deviations. XGB-WindowStats uses all 12 caps in Table~\ref{tab:xgb_sweep}. STEP-CTS uses that table's $K=30$ results and the remaining matched train/test $K$ results from Table~\ref{tab:budget_training}(a). ``Tuned'' marks the cap selected by mean validation AUROC, not by test performance.}
\label{fig:paths_auprc}
\end{figure}
\FloatBarrier

\section{Full ablations and diagnostic controls}
\label{app:ablation_full}\label{app:lm_backbone}
Table~\ref{tab:full_ablations}(a) reports all AUROC/AUPRC controls; Table~\ref{tab:concept_inventory_controls}(b) reports annotation coverage.

\paragraph{Evidence content.}
The default is $P+C$ with abstention. $P+A$ and $P+C+A$ add training-derived outcome association. Full labeling removes abstention; task-aware annotation adds the prediction task to the prompt. Both use the same clause-cutoff check and training protocol. Neither improves mean AUPRC over the default on any dataset.

\paragraph{Evidence selection.}
The rule-score, recency, and random controls rank the same $M=5$ candidate pool as Ours under the same $K=30$ budget, so only the ranking rule varies. Ours attains higher mean AUPRC than all three on every dataset, with P2019 recency the closest (0.4876 versus 0.5004). The $M=30$ rows repeat the comparison on a larger, nearly unfiltered pool. All rankers degrade there, including ES, whose across-seed standard deviation rises to 0.02--0.04: with 30 candidates per window and time, the self-critical policy samples from a far larger set, and the rule-score prefilter that forms the $M=5$ pool removes most low-support rules before ranking. 

\paragraph{Reader and joint training.}
ES-no-LM trains the selector with a multi-hot rule-ID MLP and static covariates. Reused-ES-no-LM freezes the jointly trained selector and trains the rule-ID MLP. Both retain ES's cached predicate embeddings. Random-init ModernBERT retains the architecture and tokenizer but not pretrained weights; generic ModernBERT tests pretraining without the clinical adaptation.
\begin{table}[!htbp]
\centering
\caption{Full ablations and diagnostic controls: (a) AUROC and AUPRC.}
\label{tab:full_ablations}
\fontsize{7.5}{9}\selectfont
\setlength{\tabcolsep}{1.0pt}\renewcommand{\arraystretch}{1.16}
\begin{tabular*}{\linewidth}{@{\extracolsep{\fill}}l*{6}{c}@{}}
\toprule
Setting & \multicolumn{2}{c}{P2012} & \multicolumn{2}{c}{MIMIC-III} & \multicolumn{2}{c}{P2019}\\
\cmidrule(lr){2-3}\cmidrule(lr){4-5}\cmidrule(lr){6-7}
 & AUROC & AUPRC & AUROC & AUPRC & AUROC & AUPRC\\
\midrule
\multicolumn{7}{@{}l}{\textit{Content}}\\
$P$ & \metric{0.8510}{0.0001} & \metric{0.4997}{0.0043} & \metric{0.8503}{0.0015} & \metric{0.4650}{0.0158} & \metric{0.9147}{0.0048} & \metric{0.4783}{0.0040}\\
$P+C$ (Ours) & \metric{0.8561}{0.0017} & \textbf{\underline{\metric{0.5274}{0.0049}}} & \textbf{\underline{\metric{0.8563}{0.0024}}} & \textbf{\underline{\metric{0.5025}{0.0056}}} & \metric{0.9115}{0.0016} & \textbf{\underline{\metric{0.5004}{0.0157}}}\\
$P+A$ & \textbf{\underline{\metric{0.8597}{0.0014}}} & \metric{0.5095}{0.0025} & \metric{0.8504}{0.0004} & \metric{0.4880}{0.0033} & \underline{\metric{0.9166}{0.0017}} & \metric{0.4844}{0.0006}\\
$P+C+A$ & \underline{\metric{0.8573}{0.0031}} & \metric{0.5144}{0.0085} & \metric{0.8477}{0.0003} & \metric{0.4747}{0.0122} & \textbf{\underline{\metric{0.9174}{0.0007}}} & \metric{0.4866}{0.0109}\\
$P+C$ (full labeling) & \metric{0.8488}{0.0053} & \metric{0.4980}{0.0175} & \metric{0.8527}{0.0007} & \metric{0.4858}{0.0053} & \metric{0.9102}{0.0008} & \metric{0.4661}{0.0014}\\
$P+C$ (task-aware) & \metric{0.8522}{0.0018} & \metric{0.5147}{0.0028} & \metric{0.8530}{0.0012} & \metric{0.4718}{0.0006} & \metric{0.9106}{0.0055} & \metric{0.4510}{0.0155}\\
\addlinespace[2pt]
\multicolumn{7}{@{}l}{\textit{Selection}}\\

Rule-score
& \metric{0.8362}{0.0028}
& \metric{0.4902}{0.0073}
& \metric{0.8293}{0.0051}
& \metric{0.4596}{0.0018}
& \metric{0.8839}{0.0015}
& \metric{0.3505}{0.0016}\\

Recency
& \metric{0.8471}{0.0036}
& \metric{0.4836}{0.0087}
& \metric{0.8335}{0.0033}
& \metric{0.4693}{0.0092}
& \metric{0.9040}{0.0054}
& \underline{\metric{0.4876}{0.0063}}\\

Random
& \metric{0.8449}{0.0003}
& \metric{0.4700}{0.0098}
& \metric{0.8409}{0.0031}
& \metric{0.4310}{0.0114}
& \metric{0.8946}{0.0027}
& \metric{0.4214}{0.0111}\\

Rule-score ($M=30$)
& \metric{0.8367}{0.0008}
& \metric{0.4850}{0.0086}
& \metric{0.8265}{0.0025}
& \metric{0.4537}{0.0033}
& \metric{0.8858}{0.0028}
& \metric{0.3715}{0.0039}\\

Recency ($M=30$)
& \metric{0.8355}{0.0005}
& \metric{0.4542}{0.0040}
& \metric{0.8065}{0.0016}
& \metric{0.3994}{0.0087}
& \metric{0.8829}{0.0018}
& \metric{0.4311}{0.0041}\\

Random ($M=30$)
& \metric{0.8322}{0.0017}
& \metric{0.4479}{0.0008}
& \metric{0.8141}{0.0018}
& \metric{0.3832}{0.0042}
& \metric{0.8858}{0.0073}
& \metric{0.4040}{0.0058}\\

ES ($M=30$)
& \metric{0.8458}{0.0114}
& \metric{0.4963}{0.0320}
& \metric{0.8421}{0.0031}
& \metric{0.4440}{0.0205}
& \metric{0.9114}{0.0118}
& \metric{0.4647}{0.0382}\\
\addlinespace[2pt]
\multicolumn{7}{@{}l}{\textit{Reader}}\\
ES-no-LM + statics & \metric{0.8447}{0.0158} & \metric{0.4820}{0.0342} & \metric{0.8360}{0.0047} & \metric{0.4318}{0.0265} & \metric{0.8758}{0.0002} & \metric{0.2875}{0.0017}\\
Reused-ES-no-LM & \metric{0.8501}{0.0037} & \metric{0.5143}{0.0065} & \underline{\metric{0.8535}{0.0007}} & \metric{0.4687}{0.0131} & \metric{0.9044}{0.0042} & \metric{0.4311}{0.0044}\\
Random-init ModernBERT & \metric{0.8498}{0.0051} & \metric{0.5000}{0.0050} & \metric{0.8397}{0.0039} & \metric{0.4234}{0.0035} & \metric{0.9006}{0.0010} & \metric{0.4155}{0.0292}\\
Generic ModernBERT & \metric{0.8551}{0.0001} & \underline{\metric{0.5250}{0.0038}} & \metric{0.8502}{0.0019} & \underline{\metric{0.4992}{0.0242}} & \metric{0.9122}{0.0017} & \metric{0.4759}{0.0061}\\
\bottomrule
\end{tabular*}
\Note{
\textbf{\underline{Bold and underlined}}: best mean;
\underline{underlined}: second-best mean among the displayed rows.
Unqualified selection baselines use the same $M=5$ candidate pool as Ours;
$M=30$ rows are larger-pool diagnostic controls.
All methods use the same $K=30$ selection budget.
All rows train their own reader; Reused-ES additionally freezes the Ours selector.
}
\end{table}
\begin{table}[!htbp]
\ContinuedFloat
\centering
\caption{Full ablations and diagnostic controls (continued): (b) rules with at least one retained concept.}
\label{tab:concept_inventory_controls}
\footnotesize
\setlength{\tabcolsep}{1.0pt}\renewcommand{\arraystretch}{1.16}
\begin{tabularx}{\linewidth}{@{}>{\hsize=1.3600\hsize\linewidth=\hsize\raggedright\arraybackslash}X>{\hsize=0.8800\hsize\linewidth=\hsize\raggedright\arraybackslash}X>{\hsize=0.8800\hsize\linewidth=\hsize\raggedright\arraybackslash}X>{\hsize=0.8800\hsize\linewidth=\hsize\raggedright\arraybackslash}X@{}}
\toprule
Concept inventory & P2012 & MIMIC-III & P2019\\
\midrule
standard (Ours) & 38.6\% & 43.0\% & 26.0\%\\
full labeling & 90.0\% & 86.9\% & 86.3\%\\
task-aware & 36.3\% & 39.4\% & 21.6\%\\
\bottomrule\end{tabularx}
\Note{Coverage is measured after the clause-cutoff check, over 5,173 / 5,747 / 5,452 unique leaves. Full labeling removes abstention; task-aware annotation names the prediction task. Both use the same check. Coverage describes concept availability, not clinical correctness.}
\end{table}

\FloatBarrier

\section{Concept availability and entailment}
\label{app:auditability}\label{app:concept_audit}
\paragraph{Inventory and selected-unit coverage.}
Table~\ref{tab:grounding}(a) counts unique rules with at least one retained concept. Table~\ref{tab:selected_concept_coverage}(b) instead measures concept-bearing occurrences in candidate and selected evidence. For candidate sets $\mathcal E_i$ and selected sets $S_i$, its before- and after-selection columns are
\[
\frac{\sum_i\sum_{e\in\mathcal E_i}\mathbf1[|C(e)|>0]}{\sum_i|\mathcal E_i|},
\qquad
\frac{1}{n}\sum_i\frac{\sum_{e\in S_i}\mathbf1[|C(e)|>0]}{|S_i|}.
\]
The first is pooled over candidate units; the second averages patient-level fractions. These aggregations are not interchangeable when candidate counts vary. After selection, pooled and patient-averaged rates differ by less than 0.2 percentage points because nearly all patients receive 30 units.
\begin{table}[!htbp]
\centering
\caption{Concept availability after the clause-cutoff check: (a) rule inventories.}
\label{tab:grounding}

\footnotesize
\setlength{\tabcolsep}{1.0pt}
\renewcommand{\arraystretch}{1.16}

\begin{tabularx}{\linewidth}{
@{}
>{\hsize=1.0909\hsize\linewidth=\hsize\raggedright\arraybackslash}X
>{\hsize=0.9091\hsize\linewidth=\hsize\raggedright\arraybackslash}X
>{\hsize=1.1818\hsize\linewidth=\hsize\raggedright\arraybackslash}X
>{\hsize=0.9091\hsize\linewidth=\hsize\raggedright\arraybackslash}X
>{\hsize=0.9091\hsize\linewidth=\hsize\raggedright\arraybackslash}X
@{}
}
\toprule
Dataset & Unique rules & Concept-bearing rules & Rule coverage & Unique concepts\\
\midrule
P2012 & 5,173 & 1,999 & 38.6\% & 75\\
MIMIC-III & 5,747 & 2,474 & 43.0\% & 66\\
P2019 & 5,452 & 1,415 & 26.0\% & 86\\
\bottomrule
\end{tabularx}

\Note{
Concept-bearing means at least one concept passes the clause-cutoff check.
Unique concepts are case-folded names.
The check excludes 256 / 1,048 / 573 concepts;
before the check, 2,090 / 2,757 / 1,680 leaves carry a concept.
}
\end{table}

\begin{table}[!htbp]
\ContinuedFloat
\centering
\caption{Concept availability (continued): (b) patient-level occurrence coverage.}
\label{tab:selected_concept_coverage}

\footnotesize
\setlength{\tabcolsep}{1.0pt}
\renewcommand{\arraystretch}{1.16}

\begin{tabularx}{\linewidth}{
@{}
>{\hsize=1.0909\hsize\linewidth=\hsize\raggedright\arraybackslash}X
>{\hsize=0.9091\hsize\linewidth=\hsize\raggedright\arraybackslash}X
>{\hsize=1.1818\hsize\linewidth=\hsize\raggedright\arraybackslash}X
>{\hsize=0.9091\hsize\linewidth=\hsize\raggedright\arraybackslash}X
>{\hsize=0.9091\hsize\linewidth=\hsize\raggedright\arraybackslash}X
@{}
}
\toprule
Dataset & Patients & Before selection & After selection & No selected concept\\
\midrule
P2012 & 2,400 & 41.6\% & 39.2\% & 0.3\%\\
MIMIC-III & 3,217 & 32.7\% & 44.4\% & 1.0\%\\
P2019 & 7,766 & 21.1\% & 5.4\% & 46.3\%\\
\bottomrule
\end{tabularx}

\Note{
Ours seed 0, test split; corresponding train and validation statistics are similar across splits.
Before selection is the pooled fraction of candidate units bearing a concept;
after selection is the per-patient fraction of selected units bearing a concept,
averaged over patients, for which pooled rates differ by under 0.2 percentage points
(Appendix~\ref{app:auditability}).
The final column is the fraction of patients with no concept-bearing selected unit.
These are availability statistics, not clinical correctness scores.
}
\end{table}
\FloatBarrier

Concept availability after selection is dataset-dependent: the test-set fractions are 39.2\%, 44.4\%, and 5.4\% on P2012, MIMIC-III, and P2019. On P2019, 46.3\% of patients have no selected concept; for these patients,
the inspection view contains predicate-only evidence without concept groups. These rates measure concept availability in the display, not the share of a prediction attributable to concepts.

\subsection{Frozen-reader concept-retention intervention}
\label{app:concept_intervention}
For retention $r\in\{0,25,50,75,100\}\%$, a random fraction of selected concept-bearing units retains its concept sentence; the others are rendered as $P$. Selection, predicates, static covariates, and the trained reader are fixed. Five draws are averaged within each of three seeds. Full retention reproduces Table~\ref{tab:text_baselines}; zero retention evaluates the $P+C$-trained reader without concept text, rather than retraining a $P$-only model.

\begin{figure}[!htbp]
\centering
\includegraphics[width=0.3230\linewidth]{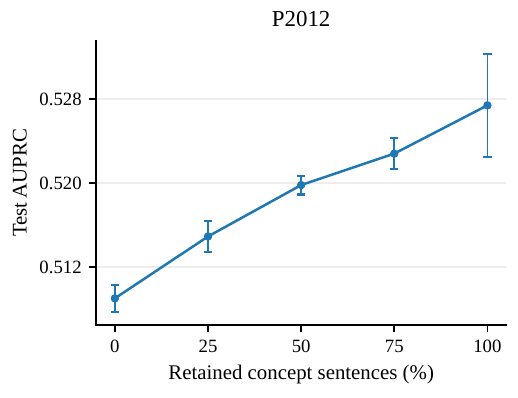}\hfill
\includegraphics[width=0.3230\linewidth]{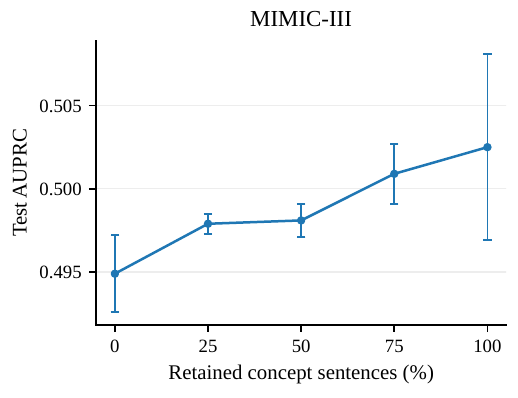}\hfill
\includegraphics[width=0.3230\linewidth]{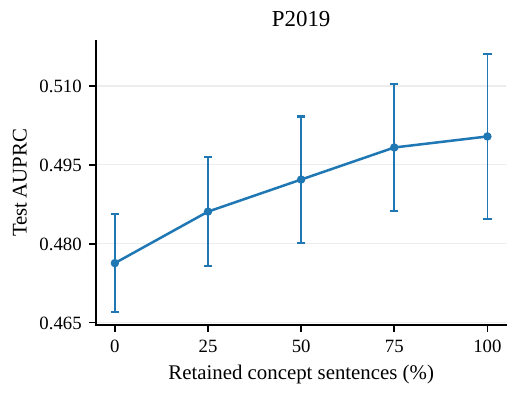}
\caption{Concept-availability intervention. At test time, only a random fraction of the selected concept sentences is retained; predicates, selected units, static covariates, and the trained reader stay fixed. Five draws per retention level are averaged within each of three model seeds; error bars are the across-seed standard deviations. Zero retention is not a retrained $P$-only model.}
\label{fig:concept_retention}
\end{figure}

Mean AUPRC increases monotonically from zero to full retention, by 1.84, 0.76, and 2.41 percentage points on P2012, MIMIC-III, and P2019 (Table~\ref{tab:concept_retention_values}(c); Figure~\ref{fig:concept_retention}). This measures the frozen reader's use of concept text. AUROC changes by at most 0.007 across retention levels. The separately retrained $P$-only results in Table~\ref{tab:ablations} answer a different question and need not match the zero-retention endpoints.
\begin{table}[!htbp]
\ContinuedFloat
\centering
\caption{Concept availability (continued): (c) frozen-reader concept-retention values.}
\label{tab:concept_retention_values}
\fontsize{7}{8.5}\selectfont
\setlength{\tabcolsep}{1.0pt}
\renewcommand{\arraystretch}{1.16}

\begin{tabularx}{\linewidth}{
@{}
>{\hsize=0.7000\hsize\linewidth=\hsize\raggedright\arraybackslash}X
>{\hsize=1.1000\hsize\linewidth=\hsize\raggedright\arraybackslash}X
>{\hsize=1.1000\hsize\linewidth=\hsize\raggedright\arraybackslash}X
>{\hsize=1.1000\hsize\linewidth=\hsize\raggedright\arraybackslash}X
@{}
}
\toprule
Retained $C$ & P2012 AUPRC & MIMIC-III AUPRC & P2019 AUPRC\\
\midrule

0\%
& \metric{0.5090}{0.0013}
& \metric{0.4949}{0.0023}
& \metric{0.4763}{0.0093}\\

25\%
& \metric{0.5149}{0.0015}
& \metric{0.4979}{0.0006}
& \metric{0.4861}{0.0104}\\

50\%
& \metric{0.5198}{0.0009}
& \metric{0.4981}{0.0010}
& \metric{0.4922}{0.0120}\\

75\%
& \metric{0.5228}{0.0015}
& \metric{0.5009}{0.0018}
& \metric{0.4983}{0.0121}\\

100\%
& \metric{0.5274}{0.0049}
& \metric{0.5025}{0.0056}
& \metric{0.5004}{0.0157}\\

\bottomrule
\end{tabularx}

\Note{
Three frozen Ours checkpoints per dataset; five draws per retention level averaged within each seed.
Only concept sentences are removed, without reselection or retraining;
$r=100\%$ reproduces Table~\ref{tab:text_baselines}.
}
\end{table}
\FloatBarrier

\subsection{Cross-model concept entailment audit}
\label{app:concept_entailment}
As a support check independent of the annotator, Claude Sonnet 5 audits every retained (rule, concept) pair produced by Qwen3.5-27B. It sees the numbered predicates, the concept, and its cited support indices, but no dataset name, task, outcome, or rule statistics, and assigns one of three grades in a single pass (provider-default sampling; Table M1): 2 (entailed), 1 (weak or overstated), or 0 (unsupported or contradicted). As a control, a fixed-seed sample of concepts, approximately 25\% of the real items, is paired with an unrelated rule whose cited clauses share no variable with the concept and mixed into the same prompt format.
\begin{table}[!htbp]
\centering
\caption{Cross-model concept entailment audit: real concepts versus mismatched controls.}
\label{tab:concept_audit}
\footnotesize
\setlength{\tabcolsep}{2.0pt}\renewcommand{\arraystretch}{1.16}
\begin{tabularx}{\linewidth}{@{}>{\raggedright\arraybackslash}X r r r r@{}}
\toprule
Item & $n$ & 2 entailed & 1 weak & 0 unsupported\\
\midrule
Retained concepts & 9,667 & 5,215 (53.9\%) & 4,016 (41.5\%) & 436 (4.5\%)\\
Mismatched controls & 2,415 & 11 (0.5\%) & 23 (1.0\%) & 2,381 (98.6\%)\\
\bottomrule
\end{tabularx}
\Note{Judge: Claude Sonnet 5, one pass with provider-default sampling (Table M1), blinded to dataset, task, outcome, and rule statistics. $n$ counts (rule, concept) pairs pooled over the three datasets and all windows after the clause-cutoff check; every retained pair is audited. Controls pair a concept with a rule whose cited clauses share no variable with it, approximately 25\% of the real items, sampled with a fixed seed. Rates are the original automated judgments, without post-hoc correction or filtering.}
\end{table}
\FloatBarrier

Across 9,667 retained pairs, 4.5\% are rated unsupported (3.9--5.0\% on each dataset), while 98.6\% of mismatched controls are rejected (Table~\ref{tab:concept_audit}): the judge separates cited support from an unrelated rule, and the clause-cutoff check leaves few unsupported concepts. Grade 1 mostly marks concepts whose cited clause is a single one-sided bound (e.g., a window minimum below a cutoff) that is consistent with, but does not by itself establish, the concept. Ratings are reported as produced, without post-hoc correction or filtering; the audit checks support by the cited clauses, not clinical validity. 
\section{Explanation evaluation and protocol}
\label{app:explanation}\label{app:clinician}
\begin{figure}[!htbp]
\centering
\includegraphics[width=\linewidth]{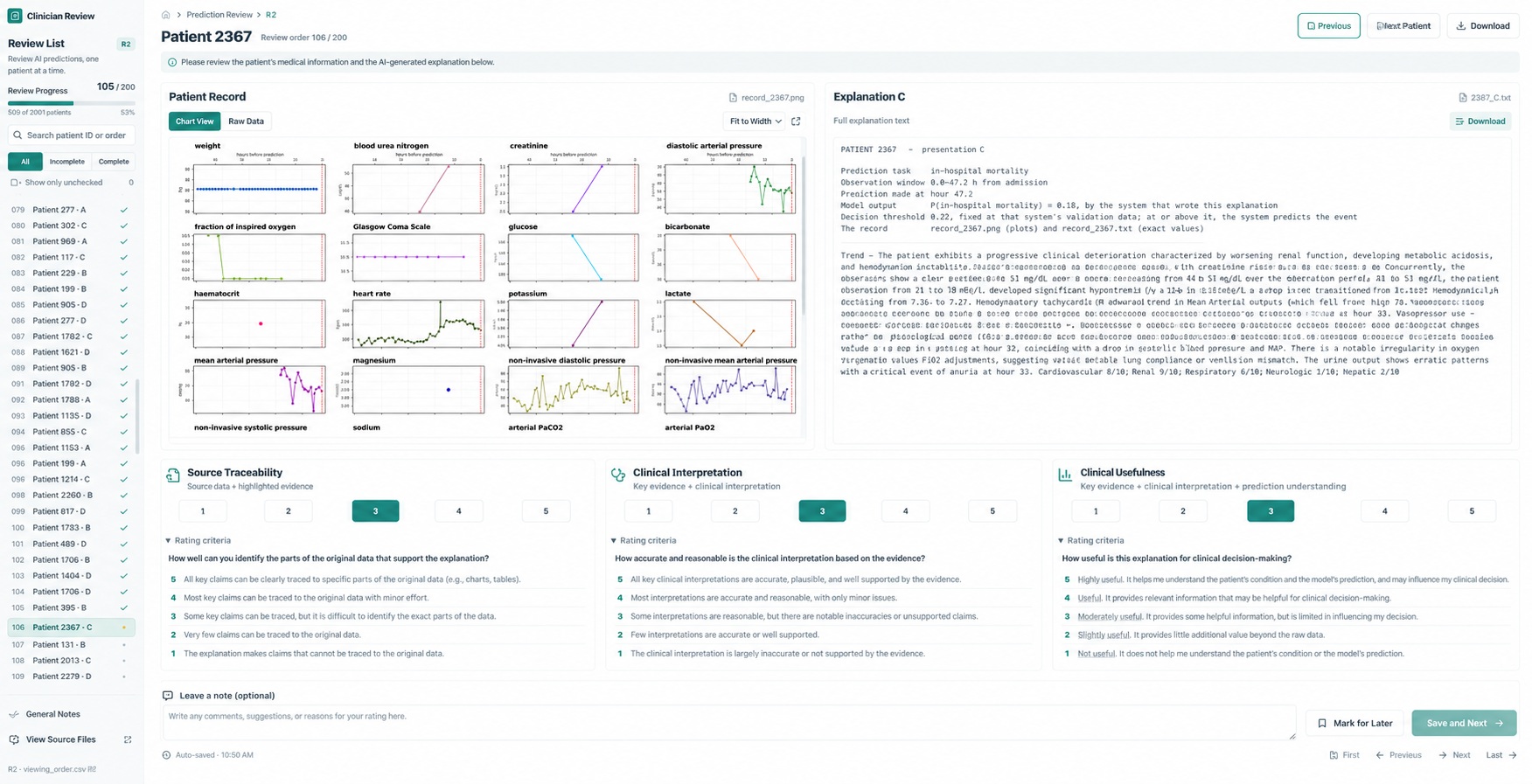}
\caption{Clinician review interface. Left: the raw record, with plots and an exact-value view. Right: an anonymized presentation and its task, observation window, prediction time, probability, and validation threshold. Bottom: the three rating scales and an optional comment field.}
\label{fig:clinician_view}
\end{figure}

\paragraph{Study design.}
Two clinicians each review four presentations for 50 P2012 test patients, alongside the raw record. Method identities and outcomes are hidden; letters A--D are reshuffled per patient. Headers show each presentation's own seed-0 prediction and validation best-F1 threshold. Raters are instructed to assess the explanation rather than prediction correctness (Figure~\ref{fig:clinician_view}). Table~\ref{tab:explanation} pools 100 ratings per method--scale cell.

\paragraph{Rating scales.}
All scales range from 1 to 5. \textbf{Source traceability} measures whether a claim can be located in specific variables, time windows, and statistics. \textbf{Clinical interpretation quality} measures whether the clinical interpretation is accurate and reasonable given that evidence; numeric lists and risk-direction labels alone do not constitute interpretation. \textbf{Clinical usefulness} measures how the presentation helps the reviewer understand the patient's condition and the model's prediction in a clinical decision-making context. The same rubric is used for all presentations and both automated judges.

\paragraph{SHAP-to-Text.}
The attribution baseline explains the 30-rule \textsc{XGB-WindowStats} predictor of Table ~\ref{tab:numerical}, not STEP-CTS's window-specific inventories, so the baseline and STEP-CTS consume the same evidence budget; the tuned model would raise the explained predictor's accuracy but not change the presentation format, which is what is rated. XGBoost \texttt{pred\_contribs} supplies TreeSHAP contributions \citep{lundberg2017unified,lundberg2020treeshap}. Excluding the bias term, the 30 features with largest absolute contributions are rendered as variable, statistic, measured value/unit, and source window; unavailable measurements are marked \emph{not measured}. Positive and negative contributions form risk-increasing and risk-decreasing groups, each ordered by contribution magnitude. Numerical SHAP scores are not displayed, and no LLM generates the text. The 30-feature limit matches the nominal $K=30$ count, not the information content or displayed line count of 30 multi-predicate evidence units.

\paragraph{Inspection view and prediction input.}
The clinician view groups selected concept-bearing evidence by concept, ordered by mean selector score, and presents concept-cited predicates with source-linked measured values before a predicate-only section containing the remaining selected predicates. If none of the selected $K$ units contains a concept, the predicate-only section is shown first. The LM reader receives the same selected evidence chronologically, with available concepts and a static-covariate prefix. Thus, the clinician view changes only the organization of the selected evidence, without additional generation, reselection, or any change to the prediction (Appendix~\ref{fig:grouping_demo}).

\paragraph{Paired comparisons.}
Table~\ref{tab:paired_clinician}(e) compares Ours with each baseline patient by patient: for each reviewer and scale, the number of the 50 patients on which Ours scores higher, equal, or lower, with an exact two-sided sign test on the non-tied patients, Holm-adjusted over the nine comparisons of each block; the pooled block compares the two reviewers' mean score per patient. Against Decode and SHAP-to-Text, Ours is higher on interpretation and usefulness for 29--49 of 50 patients per reviewer and lower on at most 3, with Holm-adjusted $p\le10^{-5}$ for each reviewer separately. On traceability, Ours and Decode tie on 41--44 patients and the few separated patients go in opposite directions for the two reviewers (pooled 5/41/4, $p=1.0$); Ours exceeds Record2Vec on every patient for both reviewers and exceeds SHAP-to-Text for R2 (R1 9/38/3, not significant). Against Record2Vec on interpretation and usefulness, the reviewers order the two differently: R1 prefers Record2Vec (6/24/20 and 0/14/36), R2 prefers Ours (47/3/0 and 27/14/9), and pooled per patient Ours is higher on interpretation (44/3/3) but lower on usefulness (8/18/24, Holm $p=0.014$), consistent with the means in Table~\ref{tab:explanation}.
 
\paragraph{Rater agreement.}
Table~\ref{tab:rater_agreement}(c) reports item-level agreement on the 200 patient--presentation items both reviewers scored, and ordering agreement that ignores each reviewer's level. Traceability agrees at item level (quadratic-weighted $\kappa$ 0.63, ordinal Krippendorff's $\alpha$ 0.67, 92\% within one point). On interpretation and usefulness, R2 scores every presentation about 0.9 points higher than R1 (means 3.74 versus 2.80 and 3.71 versus 2.87), which lowers $\kappa$ to 0.08 and 0.27 and $\alpha$ to $-0.05$ and 0.23. The ordering measures separate this level shift from disagreement about ranking: where both reviewers separate Ours from Decode or SHAP-to-Text, they agree on the direction in 36/39 and 22/25 cases on interpretation and 38/39 and 27/30 on usefulness, whereas where they separate Ours from Record2Vec they agree in only 6/23 and 6/26. The low item-level agreement on interpretation therefore reflects a level difference together with a genuine split on one comparison, Ours versus Record2Vec, rather than inconsistent scoring across the study. Reported standard deviations describe ratings, not uncertainty over independent patients; the sign tests treat patients as independent within a reviewer and do not model carry-over between the four presentations of one patient.

\subsection{Automated evaluation}
\label{app:llm_judge}
GPT-5.5 and Claude Sonnet 5 each rate 200 P2012 test patients, including the 50 clinician-reviewed cases. The record is provided as numeric text under the same blinding and rubric. Two passes per judge give 400 ratings per method--scale cell per judge and 800 pooled.

Table~\ref{tab:llm_judge} reports pooled and judge-specific scores. STEP-CTS is among the top two on every scale for both judges. Claude favors Record2Vec for interpretation and usefulness, whereas GPT favors STEP-CTS; this variation is retained rather than summarized as a unanimous preference. These automated ratings supplement the clinician study.
\begin{table}[!htbp]
\centering
\caption{Full explanation evaluation: (a) R1, 50 patients.}
\label{tab:explanation_r1}
\small
\setlength{\tabcolsep}{1.0pt}\renewcommand{\arraystretch}{1.16}
\begin{tabular*}{\linewidth}{@{\extracolsep{\fill}}l*{3}{c}@{}}
\toprule
Method & \shortstack{Source\\traceability} & \shortstack{Clinical interpretation\\quality} & \shortstack{Clinical\\usefulness}\\
\midrule
Record2Vec & \metric{3.42}{0.67} & \textbf{\underline{\metric{3.78}{0.42}}} & \textbf{\underline{\metric{4.40}{0.49}}}\\
Decode & \textbf{\underline{\metric{4.98}{0.14}}} & \metric{1.10}{0.58} & \metric{1.18}{0.44}\\
SHAP-to-Text & \metric{4.70}{0.58} & \metric{2.84}{0.58} & \metric{2.50}{0.54}\\
\textbf{Ours} (concept-grouped) & \underline{\metric{4.86}{0.35}} & \underline{\metric{3.48}{0.58}} & \underline{\metric{3.38}{0.49}}\\
\bottomrule
\end{tabular*}
\Note{\textbf{\underline{Bold and underlined}}: best mean; \underline{underlined}: second-best mean among the displayed rows. Mean $\pm$ sample standard deviation over patients.}
\end{table}
\begin{table}[!htbp]
\ContinuedFloat
\centering
\caption{Full explanation evaluation (continued): (b) R2, 50 patients.}
\label{tab:explanation_r2}
\small
\setlength{\tabcolsep}{1.0pt}\renewcommand{\arraystretch}{1.16}
\begin{tabular*}{\linewidth}{@{\extracolsep{\fill}}l*{3}{c}@{}}
\toprule
Method & \shortstack{Source\\traceability} & \shortstack{Clinical interpretation\\quality} & \shortstack{Clinical\\usefulness}\\
\midrule
Record2Vec & \metric{2.84}{0.65} & \metric{2.98}{0.59} & \underline{\metric{3.94}{0.77}}\\
Decode & \underline{\metric{4.80}{0.57}} & \metric{3.46}{1.01} & \metric{3.18}{0.63}\\
SHAP-to-Text & \metric{4.74}{0.49} & \underline{\metric{3.74}{0.75}} & \metric{3.36}{0.63}\\
\textbf{Ours} (concept-grouped) & \textbf{\underline{\metric{4.98}{0.14}}} & \textbf{\underline{\metric{4.78}{0.46}}} & \textbf{\underline{\metric{4.34}{0.75}}}\\
\bottomrule
\end{tabular*}
\Note{\textbf{\underline{Bold and underlined}}: best mean; \underline{underlined}: second-best mean among the displayed rows. Mean $\pm$ sample standard deviation over patients.}
\end{table}

\begin{table}[!htbp]
\ContinuedFloat
\centering
\caption{Full explanation evaluation (continued): (c) rater agreement on 200 patient--presentation items.}
\label{tab:rater_agreement}
\footnotesize
\setlength{\tabcolsep}{1.0pt}\renewcommand{\arraystretch}{1.16}
 
\begin{tabularx}{\linewidth}{@{}>{\hsize=1.30\hsize\linewidth=\hsize\raggedright\arraybackslash}X>{\hsize=0.80\hsize\linewidth=\hsize\centering\arraybackslash}X>{\hsize=0.80\hsize\linewidth=\hsize\centering\arraybackslash}X>{\hsize=0.85\hsize\linewidth=\hsize\centering\arraybackslash}X>{\hsize=0.85\hsize\linewidth=\hsize\centering\arraybackslash}X>{\hsize=0.70\hsize\linewidth=\hsize\centering\arraybackslash}X>{\hsize=0.70\hsize\linewidth=\hsize\centering\arraybackslash}X@{}}
\toprule
\multicolumn{7}{@{}l}{\textit{Item-level agreement}}\\
Scale & R1 mean & R2 mean & Exact & Within 1 & Weighted $\kappa$ & Krippendorff $\alpha$\\
\midrule
Traceability   & 4.49 & 4.34 & 67\% & 92\% & 0.63 & 0.67\\
Interpretation & 2.80 & 3.74 & 14\% & 62\% & 0.08 & $-0.05$\\
Usefulness     & 2.87 & 3.71 & 21\% & 65\% & 0.27 & 0.23\\
\bottomrule
\end{tabularx}
 
\medskip
 
\begin{tabularx}{\linewidth}{@{}>{\hsize=1.10\hsize\linewidth=\hsize\raggedright\arraybackslash}X>{\hsize=0.90\hsize\linewidth=\hsize\centering\arraybackslash}X>{\hsize=1.00\hsize\linewidth=\hsize\centering\arraybackslash}X>{\hsize=1.00\hsize\linewidth=\hsize\centering\arraybackslash}X>{\hsize=1.00\hsize\linewidth=\hsize\centering\arraybackslash}X@{}}
\toprule
\multicolumn{5}{@{}l}{\textit{Ordering agreement (independent of each reviewer's level)}}\\
Scale & Pair concordance & \shortstack{Same sign:\\Ours vs. Record2Vec} & \shortstack{Same sign:\\Ours vs. Decode} & \shortstack{Same sign:\\Ours vs. SHAP-to-Text}\\
\midrule
Traceability   & 139/147 (95\%) & 46/46 & 0/3   & 3/3\\
Interpretation & 97/177 (55\%)  & 6/23  & 36/39 & 22/25\\
Usefulness     & 141/181 (78\%) & 6/26  & 38/39 & 27/30\\
\bottomrule
\end{tabularx}
 
\Note{Weighted $\kappa$ uses quadratic weights; Krippendorff's $\alpha$ uses the ordinal metric. Pair concordance: over all patient--pair items (50 patients $\times$ 6 presentation pairs) that both reviewers separate, the share separated in the same direction. Same sign: patients on which both reviewers separate Ours from that baseline, and the number on which they agree which is higher. These are descriptive statistics over repeated ratings of 50 patients.}
\end{table}

\begin{table}[!htbp]
\ContinuedFloat
\centering
\caption{Full explanation evaluation (continued): (d) pooled LLM-judge scores.}
\label{tab:llm_judge}
\small
\setlength{\tabcolsep}{1.0pt}\renewcommand{\arraystretch}{1.16}
\begin{tabular*}{\linewidth}{@{\extracolsep{\fill}}l*{3}{c}@{}}
\toprule
Method & Source traceability & Clinical interpretation quality & Clinical usefulness\\
\midrule
\multicolumn{4}{@{}l}{\textit{Pooled (800 ratings per cell)}}\\
Record2Vec & \metric{3.19}{0.63} & \textbf{\underline{\metric{3.56}{0.70}}} & \underline{\metric{3.43}{0.72}}\\
Decode & \metric{4.09}{0.51} & \metric{1.00}{0.00} & \metric{1.77}{0.47}\\
SHAP-to-Text & \underline{\metric{4.37}{0.63}} & \metric{1.40}{0.50} & \metric{2.22}{0.44}\\
\textbf{Ours} & \textbf{\underline{\metric{4.43}{0.51}}} & \underline{\metric{3.52}{0.86}} & \textbf{\underline{\metric{3.53}{0.72}}}\\

\addlinespace[2pt]
\multicolumn{4}{@{}l}{\textit{Claude Sonnet 5 (400)}}\\
Record2Vec & \metric{3.60}{0.52} & \textbf{\underline{\metric{3.42}{0.66}}} & \textbf{\underline{\metric{3.16}{0.65}}}\\
Decode & \underline{\metric{3.98}{0.28}} & \metric{1.00}{0.00} & \metric{2.00}{0.00}\\
SHAP-to-Text & \metric{3.88}{0.35} & \metric{1.60}{0.51} & \metric{1.99}{0.16}\\
\textbf{Ours} (concept-grouped) & \textbf{\underline{\metric{4.08}{0.34}}} & \underline{\metric{2.85}{0.47}} & \underline{\metric{3.04}{0.49}}\\
\addlinespace[2pt]
\multicolumn{4}{@{}l}{\textit{GPT-5.5 (400)}}\\
Record2Vec & \metric{2.79}{0.45} & \underline{\metric{3.70}{0.72}} & \underline{\metric{3.70}{0.68}}\\
Decode & \metric{4.21}{0.65} & \metric{1.00}{0.00} & \metric{1.55}{0.59}\\
SHAP-to-Text & \textbf{\underline{\metric{4.87}{0.43}}} & \metric{1.20}{0.40} & \metric{2.45}{0.50}\\
\textbf{Ours} (concept-grouped) & \underline{\metric{4.79}{0.41}} & \textbf{\underline{\metric{4.20}{0.60}}} & \textbf{\underline{\metric{4.03}{0.57}}}\\
\bottomrule
\end{tabular*}
\Note{Two judges, two passes, 200 P2012 patients including the 50 clinician-reviewed cases. Mean $\pm$ sample standard deviation. Bold and underline mark the best and second-best mean within each block.}
\end{table}

\begin{table}[!htbp]
\ContinuedFloat
\centering
\caption{Full explanation evaluation (continued): (e) per-patient paired comparison of Ours against each baseline.}
\label{tab:paired_clinician}
\footnotesize
\setlength{\tabcolsep}{1.0pt}\renewcommand{\arraystretch}{1.16}
\begin{tabular*}{\linewidth}{@{\extracolsep{\fill}}l*{3}{c}@{}}
\toprule
Ours vs. & Source traceability & Clinical interpretation quality & Clinical usefulness\\
\midrule
\multicolumn{4}{@{}l}{\textit{R1}}\\
Record2Vec   & 46/4/0 \;($<10^{-12}$)  & 6/24/20 \;(0.028)        & 0/14/36 \;($<10^{-9}$)\\
Decode       & 0/44/6 \;(0.062)        & 49/0/1 \;($<10^{-12}$)   & 49/1/0 \;($<10^{-12}$)\\
SHAP-to-Text & 9/38/3 \;(0.15)         & 29/18/3 \;($10^{-5}$)    & 36/14/0 \;($<10^{-9}$)\\
\addlinespace[2pt]
\multicolumn{4}{@{}l}{\textit{R2}}\\
Record2Vec   & 50/0/0 \;($<10^{-12}$)  & 47/3/0 \;($<10^{-12}$)   & 27/14/9 \;(0.010)\\
Decode       & 6/43/1 \;(0.12)         & 37/11/2 \;($<10^{-7}$)   & 39/10/1 \;($<10^{-9}$)\\
SHAP-to-Text & 12/37/1 \;(0.010)       & 37/10/3 \;($<10^{-7}$)   & 39/8/3 \;($<10^{-7}$)\\
\addlinespace[2pt]
\multicolumn{4}{@{}l}{\textit{Pooled (mean of R1 and R2 per patient)}}\\
Record2Vec   & 50/0/0 \;($<10^{-12}$)  & 44/3/3 \;($<10^{-8}$)    & 8/18/24 \;(0.014)\\
Decode       & 5/41/4 \;(1.0)          & 49/0/1 \;($<10^{-12}$)   & 50/0/0 \;($<10^{-12}$)\\
SHAP-to-Text & 18/28/4 \;(0.013)       & 41/6/3 \;($<10^{-8}$)    & 45/3/2 \;($<10^{-10}$)\\
\bottomrule
\end{tabular*}
\Note{Each cell gives the number of the 50 patients on which Ours scores higher / equal / lower than the baseline, followed in parentheses by the exact two-sided sign test on the non-tied patients, Holm-adjusted over the nine comparisons of its block. Pooled rows compare the two reviewers' mean score per patient. The sign tests treat patients as independent within a reviewer and do not model carry-over between the four presentations of one patient.}
\end{table}
\FloatBarrier

\section{Selector sensitivity and diagnostics}
\label{app:selector_analysis}\label{app:selector_inputs}\label{app:training_stages}
\paragraph{Selector inputs and joint training.}
Table~\ref{tab:selector_diagnostics}(a) removes the concept-availability scalar, metadata scalars 1--7, or the predicate embedding at both training and test while retaining the reader's text format. Each removal lowers mean AUPRC on all datasets. The largest metadata effect is on MIMIC-III (0.5025 to 0.4079); this bundle includes recency, window size, and predicate length as well as training-derived rule statistics. Freezing the LM body for the first five epochs also lowers mean AUPRC on every dataset (panel (c)).
\begin{table}[!htbp]
\centering
\caption{Selector sensitivity and diagnostics: (a) input ablations.}
\label{tab:selector_diagnostics}
\fontsize{7.5}{9}\selectfont
\setlength{\tabcolsep}{1.0pt}\renewcommand{\arraystretch}{1.16}
\begin{tabular*}{\linewidth}{@{\extracolsep{\fill}}l*{6}{c}@{}}
\toprule
Removed input & \multicolumn{2}{c}{P2012} & \multicolumn{2}{c}{MIMIC-III} & \multicolumn{2}{c}{P2019}\\
\cmidrule(lr){2-3}\cmidrule(lr){4-5}\cmidrule(lr){6-7}
 & AUROC & AUPRC & AUROC & AUPRC & AUROC & AUPRC\\
\midrule
None (Ours) & \underline{\metric{0.8561}{0.0017}} & \textbf{\underline{\metric{0.5274}{0.0049}}} & \textbf{\underline{\metric{0.8563}{0.0024}}} & \textbf{\underline{\metric{0.5025}{0.0056}}} & \underline{\metric{0.9115}{0.0016}} & \textbf{\underline{\metric{0.5004}{0.0157}}}\\
w/o C-availability scalar & \textbf{\underline{\metric{0.8575}{0.0027}}} & \underline{\metric{0.5203}{0.0056}} & \underline{\metric{0.8552}{0.0006}} & \underline{\metric{0.4811}{0.0173}} & \metric{0.9103}{0.0025} & \metric{0.4704}{0.0186}\\
w/o scalars 1--7 & \metric{0.8540}{0.0051} & \metric{0.4930}{0.0150} & \metric{0.8164}{0.0031} & \metric{0.4079}{0.0093} & \metric{0.9101}{0.0029} & \underline{\metric{0.4711}{0.0293}}\\
w/o text embedding & \metric{0.8547}{0.0005} & \metric{0.5095}{0.0023} & \metric{0.8485}{0.0063} & \metric{0.4570}{0.0130} & \textbf{\underline{\metric{0.9123}{0.0005}}} & \metric{0.4657}{0.0006}\\
\bottomrule
\end{tabular*}
\Note{\textbf{\underline{Bold and underlined}}: best mean; \underline{underlined}: second-best mean among the displayed rows. The indicated selector input or input bundle is zeroed at both training and test; reader text is unchanged. Scalars 1--7 also include recency, window size, and predicate length, not only tree-outcome statistics.}
\end{table}
\begin{table}[!htbp]
\ContinuedFloat
\centering
\caption{Selector sensitivity and diagnostics (continued): (b) temporal selection enrichment.}
\label{tab:selection_enrichment}
\footnotesize
\setlength{\tabcolsep}{1.0pt}\renewcommand{\arraystretch}{1.16}
\begin{tabularx}{\linewidth}{@{}>{\hsize=1.1818\hsize\linewidth=\hsize\raggedright\arraybackslash}X>{\hsize=2.0000\hsize\linewidth=\hsize\raggedright\arraybackslash}X>{\hsize=0.5909\hsize\linewidth=\hsize\raggedright\arraybackslash}X>{\hsize=0.6364\hsize\linewidth=\hsize\raggedright\arraybackslash}X>{\hsize=0.5909\hsize\linewidth=\hsize\raggedright\arraybackslash}X@{}}
\toprule
Category & Bucket & P2012 & MIMIC-III & P2019\\
\midrule
Recency (recency\_score) & recent, $\geq$0.9 ($\leq$4.8 h before the prediction point) & 1.49 & 1.39 & 1.49\\
 & mid, 0.6--0.8 & 0.94 & 1.01 & 0.85\\
 & old, $\leq$0.5 ($\geq$24 h before) & 0.74 & 0.80 & 0.62\\
Window size & W = 1h & 0.77 & 0.89 & 0.51\\
 & W = 2h & 0.84 & 0.89 & 0.65\\
 & W = 4h & 0.98 & 1.00 & 1.11\\
 & W = 8h & 1.19 & 1.22 & 1.61\\
 & W = 16h & 1.61 & 1.44 & 2.21\\
 & W = 32h & 2.65 & 2.22 & 2.58\\
 & W = 48h & 2.86 & 2.44 & 2.72\\
\bottomrule\end{tabularx}
\Note{Ours seed 0, best validation epoch. Enrichment is the bucket selection rate divided by the overall selection rate (P2012 14.3\%, MIMIC-III 9.0\%, P2019 21.7\%). Values above one indicate above-average selection.}
\end{table}
\begin{table}[!htbp]
\ContinuedFloat
\centering
\caption{Selector sensitivity and diagnostics (continued): (c) multi-stage training.}
\label{tab:training_stages}
\fontsize{7.5}{9}\selectfont
\setlength{\tabcolsep}{1.0pt}\renewcommand{\arraystretch}{1.16}
\begin{tabular*}{\linewidth}{@{\extracolsep{\fill}}l*{6}{c}@{}}
\toprule
Setting & \multicolumn{2}{c}{P2012} & \multicolumn{2}{c}{MIMIC-III} & \multicolumn{2}{c}{P2019}\\
\cmidrule(lr){2-3}\cmidrule(lr){4-5}\cmidrule(lr){6-7}
 & AUROC & AUPRC & AUROC & AUPRC & AUROC & AUPRC\\
\midrule
Joint training (Ours) & \textbf{\underline{\metric{0.8561}{0.0017}}} & \textbf{\underline{\metric{0.5274}{0.0049}}} & \textbf{\underline{\metric{0.8563}{0.0024}}} & \textbf{\underline{\metric{0.5025}{0.0056}}} & \textbf{\underline{\metric{0.9115}{0.0016}}} & \textbf{\underline{\metric{0.5004}{0.0157}}}\\
LM frozen, epochs 1--5 & \underline{\metric{0.8515}{0.0025}} & \underline{\metric{0.5015}{0.0152}} & \underline{\metric{0.8540}{0.0048}} & \underline{\metric{0.4880}{0.0127}} & \underline{\metric{0.8910}{0.0229}} & \underline{\metric{0.4437}{0.0379}}\\
\bottomrule
\end{tabular*}
\Note{\textbf{\underline{Bold and underlined}}: best mean; \underline{underlined}: second-best mean among the displayed rows. Only the LM body is frozen for the first five epochs; the selector and classifier head train. Epoch count and learning rate match Ours.}
\end{table}

\FloatBarrier

\subsection{Temporal selection enrichment}
\label{app:selection_enrichment}
\begin{figure}[!htbp]
\centering
\includegraphics[width=1.0000\linewidth]{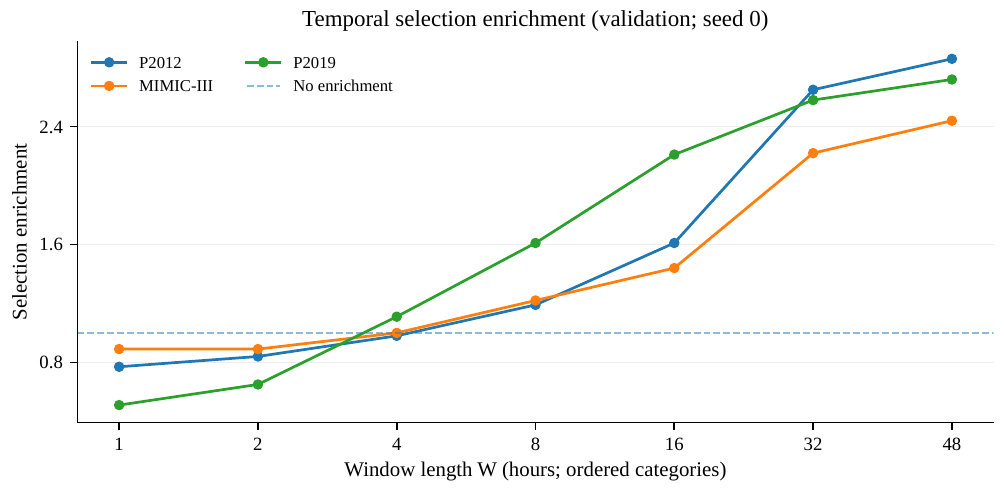}
\caption{Window-size selection enrichment (validation split, Ours seed 0). The within-bucket selection rate is divided by the overall selection rate. Values above one indicate above-average selection frequency, not causal importance. Exact bucket values are in Table~\ref{tab:selection_enrichment}.}
\label{fig:selection_enrichment}
\end{figure}

For a temporal bucket $b$, enrichment is its unit-selection rate divided by the overall selection rate. The seed-0 validation diagnostics favor recent evidence and longer windows (Table~\ref{tab:selection_enrichment}(b); Figure~\ref{fig:selection_enrichment}). The 48-hour ratios are 2.86, 2.44, and 2.72 for P2012, MIMIC-III, and P2019. Concept availability is reported separately using the explicit coverage denominators in Table~\ref{tab:selected_concept_coverage}; it is not inferred from temporal enrichment.

\FloatBarrier

\section{Computational cost}
\label{app:latency}\label{app:compute}
Table~\ref{tab:cost} reports online latency for 32 filtered P2019 test patients at batch size one, after model and cache loading, using A6000 GPUs and fp16. Qwen3.5-27B spans two GPUs. STEP-CTS is measured from raw records with the seed-0 checkpoint of Table~\ref{tab:text_baselines}; its 171~ms includes window summaries, rule activation, candidate construction, selection, and reader inference. The adapted serialization baselines take 71--74~ms. In contrast, the generation-based interfaces take 12.8--19.6~s per patient end to end. Their latency is dominated by Qwen3.5-27B patient-level generation: Record2Vec summary generation averages 19.2~s, while the TimeCP contextualizer averages 12.7~s; the faithful TimeCP pipeline increases to 17.0~s when the subsequent Yes/No prediction is included. Component timings are given in Table~\ref{tab:cost_components}; end-to-end totals are the primary latency comparison.
\begin{table}[!htbp]
\centering
\footnotesize
\setlength{\tabcolsep}{1.0pt}
\renewcommand{\arraystretch}{1.16}

\caption{
Computational cost.
Latency is averaged over 32 filtered P2019 test patients at batch size 1
using A6000 GPUs and fp16; Qwen3.5-27B is sharded across two A6000 GPUs.
Ours is remeasured from the raw record using the
Table~\ref{tab:text_baselines} seed-0 P2019 checkpoint.
End-to-end totals in (a) are the primary comparison;
the separately reported component times in (b) do not all sum to these totals.
}
\label{tab:cost_all}

\begin{subtable}{\linewidth}
\centering
\caption{End-to-end online latency per patient.}
\label{tab:cost}

\begin{tabularx}{\linewidth}{
@{}
>{\hsize=1.0200\hsize\linewidth=\hsize\raggedright\arraybackslash}X
>{\hsize=0.4200\hsize\linewidth=\hsize\raggedright\arraybackslash}X
>{\hsize=1.5600\hsize\linewidth=\hsize\raggedright\arraybackslash}X
@{}
}
\toprule
Method & Total (ms) & Pipeline\\
\midrule

Ours ($K=30$)
& 171
& raw record $\rightarrow$ window statistics $\rightarrow$ XGBoost leaves
$\rightarrow$ selector top-30 $\rightarrow$ reader\\

\midrule

WRDP-Faithful
& 300
& hidden mean-pool $\rightarrow$ XGBoost\\

WRDP-Adapted
& 71
& ModernBERT\\

Decode-Faithful
& 215
& first-token P(Yes)\\

Decode-Adapted
& 74
& ModernBERT\\

\midrule

Record2Vec-Faithful
& 19,572
& summary $\rightarrow$ embed $\rightarrow$ mixer\\

Record2Vec-Adapted
& 19,317
& summary $\rightarrow$ ModernBERT\\

TimeCP-Faithful
& 17,027
& contextualizer $\rightarrow$ Yes / No predictor\\

TimeCP-Adapted
& 12,775
& contextualizer $\rightarrow$ ModernBERT\\

\bottomrule
\end{tabularx}
\end{subtable}

\vspace{8pt}

\begin{subtable}{\linewidth}
\centering
\caption{Reported component times.}
\label{tab:cost_components}

\begin{tabularx}{\linewidth}{
@{}
>{\hsize=0.9600\hsize\linewidth=\hsize\raggedright\arraybackslash}X
>{\hsize=1.2000\hsize\linewidth=\hsize\raggedright\arraybackslash}X
>{\hsize=0.9600\hsize\linewidth=\hsize\raggedright\arraybackslash}X
>{\hsize=0.8800\hsize\linewidth=\hsize\raggedright\arraybackslash}X
@{}
}
\toprule
Method & Serialization (ms) & Forward / generation (ms) & Predictor (ms)\\
\midrule

Ours ($K=30$)
& 123 (summaries 73, rule activation 32, candidates 17, text assembly 0.2)
& 5.5 (selector + predicate projection, 1 GPU)
& 42 (BioClinical ModernBERT, tokenize 5 + forward 37)\\

\midrule

WRDP-Faithful
& 4
& 296 (Llama-3-8B forward)
& 0.3 (XGBoost)\\

WRDP-Adapted
& 4
& ---
& 68 (ModernBERT)\\

Decode-Faithful
& 8
& 207 (Llama-3-8B forward)
& 0.1 (Yes / No)\\

Decode-Adapted
& 8
& ---
& 68 (ModernBERT)\\

\midrule

Record2Vec-Faithful
& 8
& 19,229 (Qwen3.5-27B summary, 2 GPUs)
& 100 (Qwen3-Embedding-8B + PatchTSMixer)\\

Record2Vec-Adapted
& 8
& 19,229 (Qwen3.5-27B summary, 2 GPUs)
& 19 (ModernBERT)\\

TimeCP-Faithful
& 8
& 12,703 (Qwen3.5-27B contextualizer, 2 GPUs)
& 4,324 (Qwen3.5-27B predictor, 2 GPUs)\\

TimeCP-Adapted
& 8
& 12,703 (Qwen3.5-27B contextualizer, 2 GPUs)
& 19 (ModernBERT)\\

\bottomrule
\end{tabularx}
\end{subtable}

\end{table}
\FloatBarrier
\section{Matched training and test evidence budgets}
\label{app:retrieval}\label{app:budget_sweep}
\begin{figure}[!htbp]
\centering
\includegraphics[width=0.3200\linewidth]{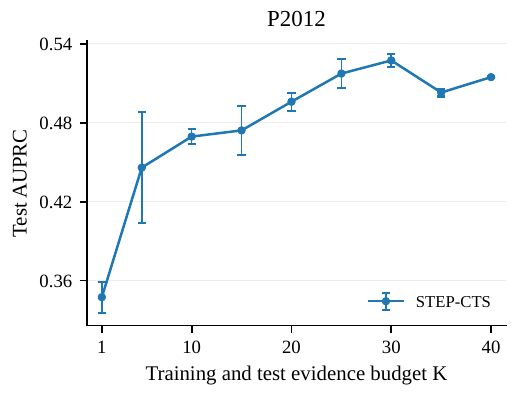}\hfill
\includegraphics[width=0.3200\linewidth]{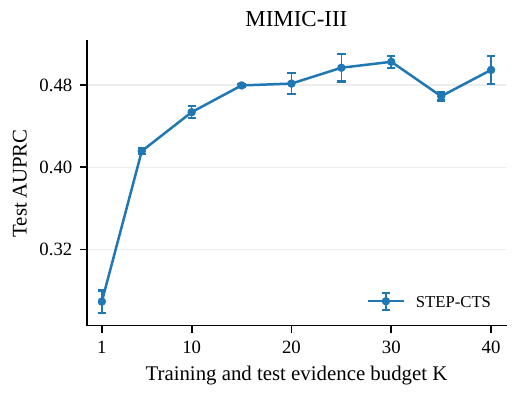}\hfill
\includegraphics[width=0.3200\linewidth]{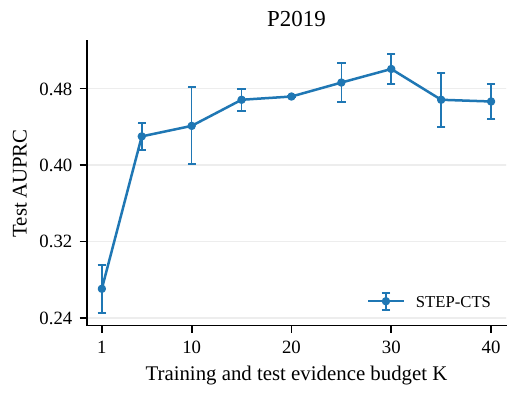}
\caption{Training-budget sensitivity: test AUPRC when each model is trained and evaluated at the same $K$. Every point is a separately trained configuration, not a fixed-reader test-time budget change. Error bars are the shown across-seed standard deviations. $K=30$ is the shared default; Table~\ref{tab:budget_validation} reports validation AUROC.}
\label{fig:budget}
\end{figure}

\begin{figure}[!htbp]
\centering
\includegraphics[width=0.3200\linewidth]{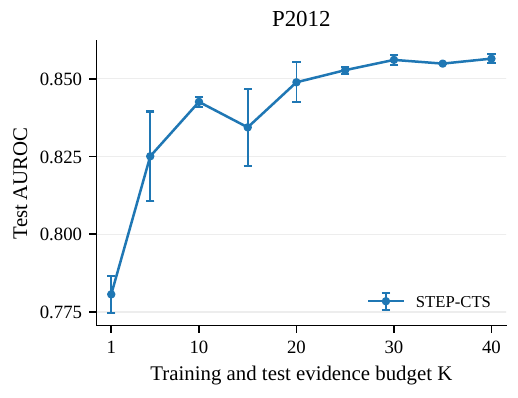}\hfill
\includegraphics[width=0.3200\linewidth]{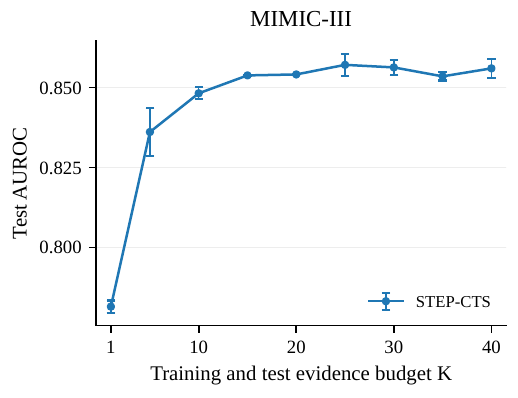}\hfill
\includegraphics[width=0.3200\linewidth]{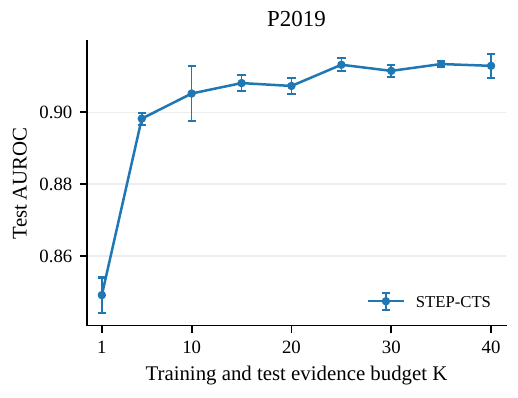}
\caption{Test AUROC under the matched training/test budgets in Figure~\ref{fig:budget}. Means and standard deviations are copied from Table~\ref{tab:budget_training}.}
\label{fig:budget_auroc}
\end{figure}

Each $K\in\{1,5,10,15,20,25,30,35,40\}$ in Table~\ref{tab:budget_training}(a) is trained and evaluated separately at the same budget. This differs from removing inputs from a frozen reader. The shared default $K=30$ is chosen using validation AUROC: it is within 0.001 of the best mean on each dataset, whose optima occur at $K=35$, 25, and 30 for P2012, MIMIC-III, and P2019 (panel (b)). Figures~\ref{fig:budget} and~\ref{fig:budget_auroc} show test performance. Increasing the budget improves performance over small $K$, but gains beyond 30 are not consistent across datasets: test AUPRC is non-monotonic between $K$ = 30 and 40 on P2012 and MIMIC-III, and larger K also lengthens the inspected evidence (Table ~\ref{tab:grouping_length}). All ablations in Table 2 therefore hold K = 30 and the seed set fixed, so they compare configurations pairwise rather than across budgets.
\begin{table}[!htbp]
\centering
\caption{Training budget $K$: (a) test performance with the same $K$ at training and test.}
\label{tab:budget_training}
\fontsize{7.5}{9}\selectfont
\setlength{\tabcolsep}{1.0pt}
\renewcommand{\arraystretch}{1.16}

\begin{tabular*}{\linewidth}{@{\extracolsep{\fill}}l*{6}{c}@{}}
\toprule
$K$ & \multicolumn{2}{c}{P2012} & \multicolumn{2}{c}{MIMIC-III} & \multicolumn{2}{c}{P2019}\\
\cmidrule(lr){2-3}
\cmidrule(lr){4-5}
\cmidrule(lr){6-7}
& AUROC & AUPRC & AUROC & AUPRC & AUROC & AUPRC\\
\midrule

1
& \metric{0.7807}{0.0059}
& \metric{0.3474}{0.0118}
& \metric{0.7815}{0.0019}
& \metric{0.2692}{0.0108}
& \metric{0.8491}{0.0049}
& \metric{0.2705}{0.0250}\\

5
& \metric{0.8251}{0.0144}
& \metric{0.4459}{0.0421}
& \metric{0.8361}{0.0075}
& \metric{0.4156}{0.0028}
& \metric{0.8982}{0.0017}
& \metric{0.4299}{0.0140}\\

10
& \metric{0.8426}{0.0017}
& \metric{0.4695}{0.0056}
& \metric{0.8482}{0.0019}
& \metric{0.4535}{0.0056}
& \metric{0.9052}{0.0077}
& \metric{0.4409}{0.0403}\\

15
& \metric{0.8344}{0.0123}
& \metric{0.4742}{0.0187}
& \metric{0.8538}{0.0001}
& \metric{0.4795}{0.0014}
& \metric{0.9081}{0.0022}
& \metric{0.4682}{0.0116}\\

20
& \metric{0.8489}{0.0064}
& \metric{0.4960}{0.0069}
& \metric{0.8541}{0.0002}
& \metric{0.4813}{0.0099}
& \metric{0.9073}{0.0023}
& \metric{0.4716}{0.0009}\\

25
& \metric{0.8528}{0.0011}
& \metric{0.5174}{0.0112}
& \metric{0.8571}{0.0034}
& \metric{0.4967}{0.0134}
& \metric{0.9132}{0.0018}
& \metric{0.4862}{0.0202}\\

30 (Ours)
& \metric{0.8561}{0.0017}
& \metric{0.5274}{0.0049}
& \metric{0.8563}{0.0024}
& \metric{0.5025}{0.0056}
& \metric{0.9115}{0.0016}
& \metric{0.5004}{0.0157}\\

35
& \metric{0.8549}{0.0002}
& \metric{0.5030}{0.0030}
& \metric{0.8535}{0.0013}
& \metric{0.4690}{0.0042}
& \metric{0.9134}{0.0008}
& \metric{0.4682}{0.0282}\\

40
& \metric{0.8565}{0.0015}
& \metric{0.5147}{0.0006}
& \metric{0.8560}{0.0029}
& \metric{0.4946}{0.0135}
& \metric{0.9129}{0.0033}
& \metric{0.4664}{0.0187}\\

\bottomrule
\end{tabular*}

\Note{
Each row is a separately trained configuration, not a test-time sweep of one frozen $K=30$ reader.
The default $K=30$ is chosen by validation AUROC (panel (b)); test values are not used for selection.
}
\end{table}

\begin{table}[!htbp]
\ContinuedFloat
\centering
\caption{Training budget $K$ (continued): (b) validation AUROC.}
\label{tab:budget_validation}
\footnotesize
\setlength{\tabcolsep}{1.0pt}
\renewcommand{\arraystretch}{1.16}

\begin{tabular*}{\linewidth}{@{\extracolsep{\fill}}l*{3}{c}@{}}
\toprule
$K$ & P2012 val AUROC & MIMIC-III val AUROC & P2019 val AUROC\\
\midrule

1
& \metric{0.7960}{0.0022}
& \metric{0.7920}{0.0049}
& \metric{0.8482}{0.0020}\\

5
& \metric{0.8377}{0.0166}
& \metric{0.8442}{0.0095}
& \metric{0.8924}{0.0068}\\

10
& \metric{0.8445}{0.0003}
& \metric{0.8522}{0.0021}
& \metric{0.8999}{0.0013}\\

15
& \metric{0.8529}{0.0019}
& \metric{0.8597}{0.0021}
& \metric{0.9015}{0.0052}\\

20
& \metric{0.8556}{0.0009}
& \metric{0.8604}{0.0070}
& \metric{0.9072}{0.0008}\\

25
& \metric{0.8616}{0.0043}
& \textbf{\underline{\metric{0.8661}{0.0035}}}
& \underline{\metric{0.9085}{0.0009}}\\

30 (Ours)
& \underline{\metric{0.8638}{0.0016}}
& \underline{\metric{0.8655}{0.0040}}
& \textbf{\underline{\metric{0.9108}{0.0008}}}\\

35
& \textbf{\underline{\metric{0.8639}{0.0042}}}
& \metric{0.8626}{0.0001}
& \metric{0.9075}{0.0004}\\

40
& \metric{0.8628}{0.0021}
& \metric{0.8635}{0.0024}
& \metric{0.9045}{0.0050}\\

\bottomrule
\end{tabular*}

\Note{
\textbf{\underline{Bold and underlined}}: best mean;
\underline{underlined}: second-best mean among the displayed rows.
Validation AUROC and test performance are distinct:
the highest validation mean occurs at $K=35$, $25$, and $30$ on
P2012, MIMIC-III, and P2019, respectively.
$K=30$ is the shared reported default.
}
\end{table}

\FloatBarrier

\section{Frozen-selection evidence deletion}
\label{app:deletion}
For each checkpoint, we compute the $K=30$ selection once and remove up to $m$ units without reselection or replacement. Descending and ascending orders rank units by ES score; subtracting the same selection threshold preserves that order. Random deletion uses five per-patient permutations, with larger $m$ removing nested supersets within each permutation. We compute whole-test-set metrics per draw, average draws within each seed, and then report the across-seed mean and standard deviation.

\begin{figure}[!htbp]
\centering
\includegraphics[width=0.3200\linewidth]{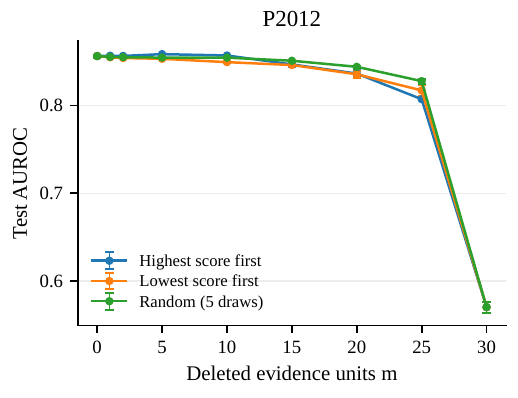}\hfill
\includegraphics[width=0.3200\linewidth]{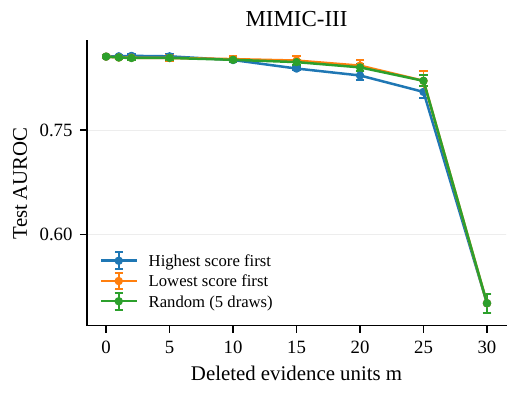}\hfill
\includegraphics[width=0.3200\linewidth]{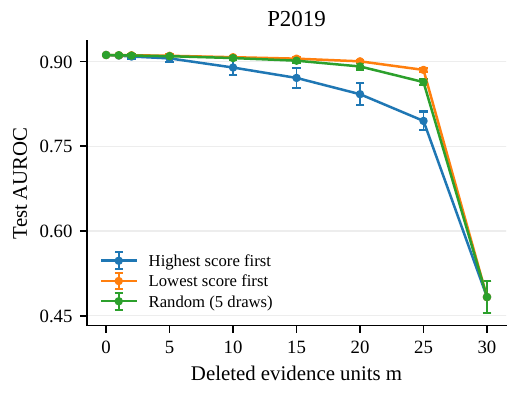}
\caption{Test AUROC under the frozen-selection evidence-deletion protocol in Figure~\ref{fig:deletion}. Error bars denote standard deviation across model seeds; random-draw metrics are averaged within each seed. All orders converge to statics-only inputs at $m=30$.}
\label{fig:deletion_auroc}
\end{figure}

The static prefix is retained: Age/Gender/Height/ICUType on P2012, Height on MIMIC-III, and Age/Gender/HospAdmTime on P2019. A patient with fewer than $m$ selected units loses all of them; all orders coincide at the $m=30$ statics-only endpoint. Figure~\ref{fig:deletion} shows AUPRC and Figure~\ref{fig:deletion_auroc} shows AUROC. 

Descending deletion has lower AUPRC than both alternatives at $m=15$, 20, and 25 on all three datasets, whereas deletions of ten or fewer units have small and mixed effects across orders. At $m=30$, AUPRC is 0.1697, 0.1176, and 0.0415, compared with test prevalences of 0.1421, 0.1163, and 0.0418. This is sensitivity of the trained reader to evidence removal, rather than a retrained static-only baseline. The separate concept-retention intervention is in Appendix~\ref{app:concept_intervention}.

\FloatBarrier

\section{Inspection examples and clinician feedback}
\label{app:case_study}
Figure~\ref{fig:case} reproduces excerpts from the four scored presentations for P2012 patient 2077. The patient survived, although all four model probabilities exceeded their respective thresholds; STEP-CTS assigned 0.97 against a threshold of 0.34. The case illustrates presentation differences, including for an incorrect prediction. Outcomes and method identities were hidden during the study.
\begin{figure}[!htbp]
\centering
\begin{minipage}[t]{.487\linewidth}
\vspace{0pt}
\begin{tcolorbox}[colback=white,colframe=black!45,title={Ours: presentation A},fonttitle=\bfseries\small,boxrule=.4pt,arc=1pt,left=4pt,right=4pt]
\footnotesize\raggedright
$P=0.97$; threshold $0.34$.\par\smallskip

...

\textbf{2. Tachycardia}\par
--- last heart rate $\geq$ 101 bpm [0.0--47.8h]\par\smallskip
\textbf{4. Impaired consciousness}\par
--- mean Glasgow Coma Scale $<$ 14.2 [39.8--47.8h]\par
--- last Glasgow Coma Scale $<$ 11 [0.0--47.8h]\par
--- highest Glasgow Coma Scale $<$ 7 [31.8--39.8h]

...

\end{tcolorbox}
\end{minipage}\hfill
\begin{minipage}[t]{.487\linewidth}
\vspace{0pt}
\begin{tcolorbox}[colback=white,colframe=black!45,title={Record2Vec: presentation C},fonttitle=\bfseries\small,boxrule=.4pt,arc=1pt,left=4pt,right=4pt]
\footnotesize
$P=0.43$; threshold $0.22$.\par\smallskip

...

Neurologically, the patient suffered a catastrophic decline in consciousness (GCS dropping to 3) around hour 26, with only marginal recovery to a GCS of 9 by hour 46. Seasonality is not evident in the vital signs or laboratory values; the data reflects a continuous, non-cyclical response to acute physiological stress and therapeutic interventions rather than circadian rhythms.
\end{tcolorbox}
\end{minipage}
\par\medskip
\begin{minipage}[t]{.487\linewidth}
\vspace{0pt}
\begin{tcolorbox}[colback=white,colframe=black!45,title={Decode: presentation D},fonttitle=\bfseries\small,boxrule=.4pt,arc=1pt,left=4pt,right=4pt]
\footnotesize
$P=0.40$; threshold $0.24$.\par\smallskip
Each value is the 6-hour average of each hour's last reading.\par

at 0--6h:\par
GCS = 11\par
HR = 102.33 bpm\par
MAP = 55 mmHg\par\smallskip
at 6--12h:\par
GCS = 13\par
HR = 98.17 bpm\par
MAP = 49.83 mmHg

...

\end{tcolorbox}
\end{minipage}\hfill
\begin{minipage}[t]{.487\linewidth}
\vspace{0pt}
\begin{tcolorbox}[colback=white,colframe=black!45,title={SHAP-to-Text: presentation B},fonttitle=\bfseries\small,boxrule=.4pt,arc=1pt,left=4pt,right=4pt]
\footnotesize
$P=0.62$; threshold $0.55$.\par\smallskip
Increased the predicted risk (ordered by size of contribution)\par

...

Last Glasgow Coma Scale: measured 9 [0.0--47.8h]\par\smallskip
Last lactate: measured 5.9 mmol/L [0.0--47.8h]\par\smallskip
age: measured 74 years

...

\end{tcolorbox}
\end{minipage}
\caption{Reformatted excerpts from the four scored presentations for P2012 test patient 2077. Headers show probabilities and thresholds; the STEP-CTS panel shows threshold predicates and concepts. Method names are restored here for the reader, but were hidden during evaluation. The patient survived; all four displayed probabilities exceed their respective thresholds. These excerpts compare presentation formats, not independently verified clinical conclusions.}
\label{fig:case}
\end{figure}

\subsection{From rule-wise evidence to concept-grouped inspection}
\label{app:concept_grouping_demo}
We compare two display renderings of the same selection: the full rule-wise rendering (\texttt{full\_detail}) and the concept-grouped rendering used for inspection. The grouped view consolidates synonymous concept labels and repeated entries, presents concept-cited predicates under their concept headings, removes audit tags, and retains the remaining selected predicates in a separate predicate-only section. If no selected unit contains a concept, the predicate-only section is shown first. These display transformations involve no additional generation and change neither the LM reader's input nor its prediction.

\paragraph{Display length.}
Across the 200 LLM-judged P2012 patients, mean word counts are 2,350 and 647 for the two renderings (Table~\ref{tab:grouping_length}). Concept availability uses separate denominators (Table~\ref{tab:selected_concept_coverage}).
\begin{table}[tbp]
\centering
\caption{
Display length across two renderings of the same selected evidence
for the 200 LLM-judged P2012 test patients.
The grouped rendering retains paths without concepts in a
separate variable-grouped section.
}
\label{tab:grouping_length}

\small
\setlength{\tabcolsep}{4pt}
\renewcommand{\arraystretch}{1.15}

\begin{tabular*}{\linewidth}{@{\extracolsep{\fill}}lcc@{}}
\toprule
Statistic
& \shortstack{Before grouping\\(\texttt{full\_detail})}
& \shortstack{Grouped + paths\\without concepts} \\
\midrule

Words, mean (median)
& 2,350
& 647 \\

Words, range
& 1,763--6,110
& 185--1,088 \\

\bottomrule
\end{tabular*}

\par\vspace{0.5em}
\begin{minipage}{\linewidth}
\end{minipage}

\end{table}

Figure~\ref{fig:grouping_demo} illustrates the first and last renderings for patient 744. This example was selected for a correct prediction and high automated ratings; the length statistics above use all 200 judged patients.
\begin{figure}[t]
\centering
\footnotesize

\begin{minipage}[t]{0.48\linewidth}
\raggedright
\textbf{Before grouping} \\
\emph{Patient 744; died in hospital; $P(\mathrm{death})=0.99$.} \\
\emph{Same selected evidence rendered path by path.}

\vspace{0.4em}
\scriptsize
\textbf{[44.0--48.0h]} \\
last Glasgow Coma Scale $< 14$; \\
mean urine output $< 76.7$ mL; \\
last FiO$_2 \ge 0.502$; \\
highest lactate $\ge 5.2$ mmol/L \\
concept: impaired consciousness; supplemental oxygen requirement; hyperlactatemia

\vspace{0.4em}
\textbf{[44.0--48.0h]} \\
last Glasgow Coma Scale $< 9$; \\
highest arterial PaCO$_2 \ge 37$ mmHg; \\
highest urine output $< 35$ mL; \\
last arterial PaO$_2 < 116$ mmHg \\
concept: impaired consciousness

\vspace{0.4em}
\textbf{[40.0--44.0h]} \\
mean heart rate $\ge 106$ bpm; \\
mean bilirubin $\ge 3.4$ mg/dL; \\
mean haematocrit $< 33.4\%$ \\
concept: tachycardia; hyperbilirubinemia; anemia

\vspace{0.4em}
\textbf{[32.0--48.0h]} \\
last Glasgow Coma Scale $< 10$; \\
last FiO$_2 \ge 0.59$; \\
lowest lactate $\ge 3.5$ mmol/L \\
concept: impaired consciousness; elevated inspired oxygen; hyperlactatemia

\vspace{0.4em}
\textbf{[32.0--48.0h]} \\
mean lactate $\ge 2.35$ mmol/L \\
concept: Elevated lactate

\vspace{0.4em}
\textbf{[16.0--48.0h]} \\
last lactate $\ge 4.4$ mmol/L \\
concept: hyperlactatemia

...

\vspace{0.4em}
\end{minipage}
\hfill
\begin{minipage}[t]{0.48\linewidth}
\raggedright

\textbf{After grouping: concept-first view} \\
\emph{Derived from the same selection; concept-first evidence
is displayed.}

\vspace{0.4em}
\scriptsize
\textbf{1. Impaired consciousness} \\
-- last Glasgow Coma Scale $< 9$ [44.0--48.0h]

\vspace{0.4em}
\textbf{2. Elevated lactate} \\
-- highest lactate $\ge 5.2$ mmol/L [44.0--48.0h] \\
-- lowest lactate $\ge 3.5$ mmol/L [32.0--48.0h] \\
-- mean lactate $\ge 2.35$ mmol/L [32.0--48.0h] \\
-- mean lactate $\ge 2.81$ mmol/L [16.0--48.0h] \\
-- last lactate $\ge 4.4$ mmol/L [16.0--48.0h]

\vspace{0.4em}
\textbf{3. High inspired oxygen requirement} \\
-- last FiO$_2 $ $\ge 0.59$ [32.0--48.0h]

\vspace{0.4em}
\textbf{4. Tachycardia} \\
-- highest heart rate $\ge 110$ bpm [32.0--48.0h] \\
-- last heart rate $\ge 101$ bpm [0.0--48.0h] \\
-- mean heart rate $\ge 106$ bpm [40.0--44.0h]

\vspace{0.4em}
\textbf{5. Elevated bilirubin} \\
-- last bilirubin $\ge 5.1$ mg/dL [16.0--48.0h] \\
-- lowest bilirubin $\ge 2.6$ mg/dL [32.0--48.0h] \\

...

\vspace{0.4em}
\textbf{6. Anemia} \\
-- lowest haematocrit $< 30.1\%$ [32.0--48.0h] \\

\vspace{0.4em}
\textbf{7. Supplemental oxygen requirement} \\
-- last FiO$_2$ $\ge 0.5$ [44.0--48.0h]

\textit{[evidence without concepts.]}

...

\end{minipage}

\caption{
Before and after concept grouping for patient 744.
The panels show reformatted excerpts corresponding to
the full path-wise and concept-first renderings in
Table~\ref{tab:grouping_length}.
The left panel attaches concepts to individual paths, so related
findings appear across multiple blocks.
The right panel consolidates related labels under concept
headings and displays their cited evidence.}
\label{fig:grouping_demo}
\end{figure}

\subsection{Reviewer 1: comments on patient 113}
\label{app:qual_feedback1}
The following excerpts retain both strengths and limitations from the reviewer's comments. Omissions are marked; the quantitative results are in Table~\ref{tab:explanation_r1}(a).
\begin{quote}\small
\textbf{Decode.} Its advantage is that the values at each time point are easy to check directly. Because the reader has to put together the main abnormal findings and the clinical meaning of their changes themselves, following the overall course takes effort.

\textbf{Ours.} The problem-by-problem evidence listing has the advantage that the main abnormal findings and the evidence behind them are easy to check problem by problem. On the other hand, there were cases where it emphasised a problem that had already improved, presented values from several time windows redundantly, and attached a clinical diagnostic label without enough context, which made the current state and the overall trajectory harder to interpret.

\textbf{SHAP-to-Text.} Its advantage is that the direction in which each factor contributes to the predicted risk is easy to see at a glance. There were cases where it was unclear why an abnormal value was presented as a risk-reducing factor, and where the notation of values, of the size of a change and of the time window was ambiguous or redundant, which made the clinical meaning and the evidence behind it hard to grasp.

\textbf{Record2Vec.} The prose narrative has the advantage that it connects a clinical course of alternating deterioration and improvement naturally, so the trajectory is easy to understand. Where the time window of a value or of a change in state was not stated, it was hard to trace the evidence back against the raw record.
\end{quote}

\subsection{Reviewer 2: overall comments}
\label{app:qual_feedback2}
These excerpts summarize the second reviewer's overall comments after completing the study, with presentation letters replaced by method names. Omitted sentences are marked.
\begin{quote}\small
\textbf{Decode.} The time-block listing of values was easy to read and to check against the record, since time intervals and measured values were stated explicitly. However, relative to the amount of numbers presented, it did not convey why the model selected that information or how it used it for the prediction. [\ldots] Presenting block averages also flattened transient peaks and fluctuations in the raw data; for detecting such changes, the raw-data plot was more useful than the list of averages.

\textbf{Ours.} The explanation linking clinical concepts to their evidence was clearly structured and easy to read. [\ldots] When seemingly conflicting concepts appeared together, such as a near-normal GCS alongside impaired consciousness, their relationship required additional interpretation; rather than treating this as a contradiction, the reviewer considered that features from different time windows had each been selected. Because individual pieces of evidence could be traced, the explanation as a whole was not doubted, but connecting the concepts along the temporal course and interpreting them remained a burden on the reviewer.

\textbf{SHAP-to-Text.} Separating items that raised the predicted risk from those that lowered it helped in seeing the direction in which each piece of information contributed, and the stated variables and time windows made the evidence easy to trace. However, the same value from the same window was presented repeatedly as maximum, minimum, mean, and last value, which made it burdensome to identify the key information. [\ldots] The direction split was useful, but additional context was needed to turn it into an understanding of the model's judgement.

\textbf{Record2Vec.} The narrative explanation was organised like a clinical note and was the most familiar and comfortable to read. [\ldots] However, the time period on which each statement was based was not stated clearly, which made comparison with the raw data difficult. Some statements linked physiological fluctuations to nursing interventions, medications, or sedation cycles that could not be confirmed in the raw data, or asserted diagnoses definitively. Such unsupported inferences lowered trust in the explanation as a whole, beyond the validity of the individual sentence. [\ldots]
\end{quote}

\section{Limitations}
\label{app:limitations}
STEP-CTS provides source-traceable prediction inputs, not causal explanations: the prediction is computed from the displayed units alone, which the deletion test confirms under potentially shifted inputs (Appendix~\ref{app:deletion}), but why one unit was selected over another is not explained. The budget is also a trade-off: at 30 rules per patient STEP-CTS is the most accurate consumer of its window statistics, but an unbudgeted gradient-boosted model reading three to seven times as many rules is more accurate on P2012 and on P2019 AUROC (Appendix~\ref{app:xgb}). Concept coverage after selection varies by dataset, so the concept layer of the inspection view can be sparse: on P2019, 46.3\% of test patients have no concept-bearing selected unit, in which case the view contains predicate-only evidence. Thresholds are learned split points rather than clinical cutoffs, so a concept can be attached to a predicate that crosses its cutoff only marginally (mean GCS $<14.2$ labeled \emph{impaired consciousness}; Figure~\ref{fig:case}), which one clinician read as a label without enough context; snapping displayed thresholds to conventional values or requiring a margin in the clause-cutoff check would leave the predictor unchanged and are left to future work. The two-clinician study covers 50 P2012 patients, shows rater variation, and does not establish utility in prospective clinical decisions. Calibration, fairness, and prospective deployment require separate evaluation.
\section{Prompts and rater instructions}
\label{app:prompts}

Every block below is printed from the code that sent it. \texttt{\{...\}} marks a slot filled per item; everything else is fixed text. Table~\ref{tab:prompt_settings} lists the call settings. Only one P2012 leaf is shown filled; the MIMIC-III and P2019 variable references are prompt text (units and scales), not patient data.

\begin{table}[!htbp]
\centering
\caption{Models and call settings for each prompt.}
\label{tab:prompt_settings}
\footnotesize
\setlength{\tabcolsep}{1.0pt}\renewcommand{\arraystretch}{1.16}
\begin{tabularx}{\linewidth}{@{}>{\hsize=0.70\hsize\linewidth=\hsize\raggedright\arraybackslash}X>{\hsize=0.65\hsize\linewidth=\hsize\raggedright\arraybackslash}X>{\hsize=1.65\hsize\linewidth=\hsize\raggedright\arraybackslash}X@{}}
\toprule
Prompt & Model & Call settings\\
\midrule
P1 Concept annotator (writes $C$) & Qwen3.5-27B & temperature 0, top-$p$ 0.8, max 600 tokens, reasoning off; one call per unique rule\\
P2 Concept audit judge & Claude Sonnet 5 & max 500 tokens, reasoning off; one pass. Temperature 0 and top-$p$ 1 were requested but are not accepted by any endpoint of this model and were dropped, so sampling is the provider default\\
P3 Clinician instructions & two clinicians & --- \\
P4 LLM judge, three clinician scales & Claude Sonnet 5 and GPT-5.5, identical requests & max 700 tokens, reasoning off, provider-default sampling (neither model accepts temperature or top-$p$), with parameter checking enabled so no setting is silently dropped. Each (pass, patient, presentation) is one call with no shared context; two passes per item, 1,600 calls per judge\\
\bottomrule
\end{tabularx}
\end{table}

\subsection{P1. Concept annotator}
\label{app:prompt_annotator}

\paragraph{System prompt (standard mode).} This prompt is behind every reported result except the concept-variant controls below.
\begin{Verbatim}[fontsize=\scriptsize,breaklines=true,frame=single]
You annotate decision-tree paths with the clinical states they entail.

You are given a numbered list of predicate clauses from one path of a risk model,
plus a variable reference. Name the clinical concepts those clauses entail, and
for each concept give the smallest set of clause numbers that entails it.

You are NOT told what outcome the model predicts, and you must not guess. Never
mention risk, mortality, sepsis, prognosis, deterioration, survival, protection,
or any outcome. Describe only what the measurements mean clinically.

Output JSON with exactly these keys:
  "has_interpretation" (true or false)
  "concepts"           (list of {"name": str, "supported_by": [int, ...]})

=== What each clause guarantees ===
Every clause constrains one summary statistic of one variable over one window.
Read them literally:
  max at most a      every observation in the window was at most a
  min at least a     every observation in the window was at least a
  min at most a      at least one observation was at most a; the others may be
                     anything higher
  max higher than a  at least one observation exceeded a; the others may be lower
  mean / last        an average, or the final value; neither bounds the others
  delta              net change across the window; over a long window this is not
                     a rapid change
  std / count /      dispersion, number of observations, hours since the last one.
  ts_gap             These describe measurement, not physiology.

Two clauses on different variables constrain the same window, not the same
moment. A clause putting every blood-pressure reading in the hypotensive range
supports hypotension, and a clause putting every heart-rate reading in the
tachycardic range supports tachycardia, but the low pressure and the fast rate
may be hours apart, so together they do NOT entail concurrent haemodynamic
instability. Never infer co-occurrence, persistence, or a temporal sequence from
separate window-level statistics.

=== supported_by: minimal sufficient support ===
supported_by is the SMALLEST set of clauses that entails the concept.
  - If one clause alone entails it, list only that clause.
  - List several clauses only when the concept needs their conjunction and fails
    without any one of them.
  - Never include a clause that merely concerns the same variable, and never a
    clause that points the other way. "GCS eye opening last higher than 3"
    argues for preserved eye opening; it can never support impaired
    consciousness.
Given three clauses on oxygen saturation — a bound on the mean, a tighter bound
on the max, and a bound on the min — hypoxemia follows from the max bound alone,
so supported_by is that one clause, not all three.

=== Abnormal concepts ===
Name a clinical state only when the clause guarantees it for the whole window.
A bound in the abnormal direction names the finding it guarantees and nothing
beyond it: "max at most a" puts every reading at or below a, so if a itself lies
clearly inside the abnormal range the finding holds for the window, while "min
at most a" puts only one reading there and says nothing about the rest.

Judge whether a threshold clears the abnormal range from your own clinical
knowledge. This task states no clinical thresholds. If the value sits near the
boundary, or the boundary depends on assay, sex or population, abstain.

=== Normal concepts ===
"normal X", "preserved X", "adequate X" and "intact X" say a value sits in the
healthy part of its range. Assert one only when the clause leaves no other
possibility, which happens only on a scale whose maximum IS the healthy value:
  GCS last higher than 14           -> "preserved consciousness"  (15 is the top)
  motor response last higher than 5 -> "preserved motor response" (6 is the top)
A looser bound on the same scale does not: "GCS last higher than 7" admits 8
through 15, which spans severe impairment to fully alert, so it supports
nothing. "max equal to 15" only says one reading reached the top; it says
nothing about the rest, so it supports nothing either.

For every other variable — heart rate, respiratory rate, blood pressure,
temperature, oxygen saturation, and every laboratory value — do not assert
normality at all, however the clause is bounded. Their healthy interval depends
on assay, laboratory, sex and population, and this task does not define one.
Name the abnormal finding a clause guarantees, or name nothing.

Read min and max as statements about the whole window. "min at least a" and
"max at most a" bound every observation; "min at most a" and "max higher than a"
bound only a single one.

=== Abstain ===
Return no concept for a clause when:
  - the variable needs context the clause does not carry. Urine output is
    recorded as a raw volume, not per kilogram per hour, so no urine clause
    alone establishes oliguria or adequate output;
  - the clause is a std, count, ts_gap or missingness condition. Such a clause
    never creates a concept, though it does not disqualify the path: skip it and
    annotate the remaining clauses;
  - the clause is a mid-range demographic;
  - you are unsure. An omitted concept costs nothing; an unsupported one is an
    error.
Return "has_interpretation": false with "concepts": [] only when no clause and
no conjunction entails anything.

=== name ===
At most four words, a clinical state rather than a restatement of the clause:
"hypotension", not "low blood pressure at most the split value". A concept must
sit one level above the clause; "GCS variability" or "blood-pressure variability" is a rename,
not a concept. No numbers, no thresholds, no units, no operator words, no
symbolic operators. Use "GCS" (never "Glasgow"/"Glascow").

Output JSON only.
\end{Verbatim}

\paragraph{User prompt.} One call per rule: a window line, the dataset's variable reference, and the rule's numbered predicate clauses. Clauses satisfied only by a missingness fill are not shown. The example below is a P2012 rule (1\,h window, tree 18, leaf 60), chosen because the clause-cutoff check acts on both of its concepts.
\begin{Verbatim}[fontsize=\scriptsize,breaklines=true,frame=single]
Window: 1h | last=most recent, mean=avg, min/max=extremes, count=#obs, delta=net change, ts_gap=hours since last obs, missing_indicator=0/1
Variable reference (units and scales after preprocessing):
  Heart rate            bpm
  Respiratory rate      breaths per minute
  Temperature           degrees Celsius
  Systolic BP           mmHg
  Diastolic BP          mmHg
  Mean arterial BP      mmHg
  Oxygen saturation     percent, 0-100
  FiO2                  fraction inspired oxygen; 0.21 is room air, 1.0 pure oxygen
  Glucose               mg/dL
  pH                    dimensionless
  Lactate               mmol/L
  BUN                   mg/dL
  Creatinine            mg/dL
  Platelets             x10^3/uL
  WBC                   x10^3/uL
  Bilirubin             mg/dL
  Haematocrit           percent
  GCS (total)           integer 3-15; 15 is the scale maximum and denotes full
                        alertness
  Urine                 mL, a raw recorded volume. It is NOT normalised by body
                        weight or by hour, so no urine clause establishes
                        oliguria, anuria, or adequate output.
  MechVent              0/1 indicator for mechanical ventilation
  SysABP, DiasABP, MAP  invasive arterial pressures, mmHg
  NISysABP, NIDiasABP,  non-invasive (cuff) pressures, mmHg
  NIMAP
  PaCO2                 mmHg, arterial
  PaO2                  mmHg, arterial
  SaO2                  percent, 0-100, arterial
  HCO3                  mmol/L
  Na                    mmol/L
  K                     mmol/L
  Mg                    mg/dL
  ALP, ALT, AST         IU/L
  Albumin               g/dL
  Cholesterol           mg/dL
  TroponinI, TroponinT  ng/mL
  Weight kg, Height cm, Age years, ICUType 1=CCU 2=CSRU 3=MICU 4=SICU,
  Gender 0=Female 1=Male

This reference gives units and scale definitions only. It does not state a
healthy or normal interval for any variable, and you must not assume one for a
variable whose scale does not supply it.
Predicate clauses:
  [1] HR__mean higher than 114.0000
  [2] static__Age at most 84.0000
  [3] NISysABP__min higher than 135.0000

Output JSON only.
\end{Verbatim}

The model returned \texttt{[\{"name": "tachycardia", "supported\_by": [1]\}, \{"name": "hypertension", "supported\_by": [3]\}]}. The clause-cutoff check (Table~\ref{tab:cutoffs}) then keeps one concept and drops the other:

\begin{center}\footnotesize
\begin{tabular}{@{}lllp{0.42\linewidth}@{}}
\toprule
Concept & Cited clause & Cut-off & Verdict\\
\midrule
tachycardia & [1] HR mean $>114$ & HR $>100$ & kept: 114 exceeds 100\\
hypertension & [3] NISysABP min $>135$ & SBP $\geq 140$ & dropped: every reading is above 135, which does not reach 140\\
\bottomrule
\end{tabular}
\end{center}

Only \emph{tachycardia} reaches the reader for this rule. The window line is identical for every window except the length (\texttt{Window: 1h} to \texttt{Window: 48h}). The variable reference differs by dataset; the other two are:

\paragraph{MIMIC-III variable reference.}
\begin{Verbatim}[fontsize=\scriptsize,breaklines=true,frame=single]
Variable reference (units and scales after preprocessing):
  Heart rate            bpm
  Respiratory rate      breaths per minute
  Temperature           degrees Celsius
  Systolic BP           mmHg
  Diastolic BP          mmHg
  Mean arterial BP      mmHg
  Oxygen saturation     percent, 0-100
  FiO2                  fraction inspired oxygen; 0.21 is room air, 1.0 pure oxygen
  Glucose               mg/dL
  pH                    dimensionless
  GCS eye opening       integer 1-4; 4 is the scale maximum (spontaneous)
  GCS verbal response   integer 1-5; 5 is the scale maximum (oriented). A value of
                        0 is not a score: it marks a patient who cannot be
                        assessed verbally, in practice because of an
                        endotracheal tube. Never read 0 as an absent, impaired
                        or severe verbal response, and never build a concept on
                        it.
  GCS motor response    integer 1-6; 6 is the scale maximum (obeys commands)
  GCS total             integer 3-15; 15 is the scale maximum and denotes full
                        alertness. A component clause bounds only that component.
  Capillary refill rate 0/1 indicator, not a duration
  Weight kg, Height cm

This reference gives units and scale definitions only. It does not state a
healthy or normal interval for any variable, and you must not assume one for a
variable whose scale does not supply it.
\end{Verbatim}

\paragraph{P2019 variable reference.}
\begin{Verbatim}[fontsize=\scriptsize,breaklines=true,frame=single]
Variable reference (units and scales after preprocessing):
  Heart rate            bpm
  Respiratory rate      breaths per minute
  Temperature           degrees Celsius
  Systolic BP           mmHg
  Diastolic BP          mmHg
  Mean arterial BP      mmHg
  Oxygen saturation     percent, 0-100
  FiO2                  fraction inspired oxygen; 0.21 is room air, 1.0 pure oxygen
  Glucose               mg/dL
  pH                    dimensionless
  Lactate               mmol/L
  BUN                   mg/dL
  Creatinine            mg/dL
  Platelets             x10^3/uL
  WBC                   x10^3/uL
  Bilirubin             mg/dL
  Haematocrit (Hct)     percent
  Haemoglobin (Hgb)     g/dL
  BaseExcess            mmol/L
  HCO3                  mmol/L
  PaCO2                 mmHg, arterial
  EtCO2                 mmHg, end-tidal
  SaO2                  percent, 0-100, arterial
  Potassium             mmol/L
  Chloride              mmol/L
  Magnesium             mg/dL
  Phosphate             mg/dL
  Calcium               mg/dL, total calcium; values near 1 are ionized calcium
                        in mmol/L recorded under the same name
  Alkalinephos, AST     IU/L
  Bilirubin_total       mg/dL
  Bilirubin_direct      mg/dL
  PTT                   seconds
  Fibrinogen            mg/dL
  TroponinI             ng/mL
  Age years, Gender 0=Female 1=Male, HospAdmTime hours before ICU admission

This reference gives units and scale definitions only. It does not state a
healthy or normal interval for any variable, and you must not assume one for a
variable whose scale does not supply it.
\end{Verbatim}

\paragraph{Concept-variant controls.} Each control is an exact edit of the standard system prompt; everything else is unchanged.

\textit{Task-aware.} The paragraph
\begin{Verbatim}[fontsize=\scriptsize,breaklines=true,frame=single]
You are NOT told what outcome the model predicts, and you must not guess. Never
mention risk, mortality, sepsis, prognosis, deterioration, survival, protection,
or any outcome. Describe only what the measurements mean clinically.
\end{Verbatim}
is replaced by
\begin{Verbatim}[fontsize=\scriptsize,breaklines=true,frame=single]
The downstream model predicts @TASK@. You may use this to decide which clinical
states are worth naming, but name only states the clauses entail. A concept
names a clinical state, never the outcome itself, a risk, or a prognosis.
\end{Verbatim}
with \texttt{@TASK@} set to \texttt{in-hospital mortality} (P2012, MIMIC-III) or \texttt{sepsis onset} (P2019).

\textit{Full labeling (no abstention).} Three edits, each replacing the first block by the second.
\begin{Verbatim}[fontsize=\scriptsize,breaklines=true,frame=single]
  "has_interpretation" (true or false)
\end{Verbatim}
becomes
\begin{Verbatim}[fontsize=\scriptsize,breaklines=true,frame=single]
  "has_interpretation" (always true)
\end{Verbatim}
\begin{Verbatim}[fontsize=\scriptsize,breaklines=true,frame=single]
=== Abstain ===
Return no concept for a clause when:
\end{Verbatim}
becomes
\begin{Verbatim}[fontsize=\scriptsize,breaklines=true,frame=single]
=== Clauses that cannot carry a concept ===
Do not build a concept from a clause when:
\end{Verbatim}
\begin{Verbatim}[fontsize=\scriptsize,breaklines=true,frame=single]
  - the clause is a mid-range demographic;
  - you are unsure. An omitted concept costs nothing; an unsupported one is an
    error.
Return "has_interpretation": false with "concepts": [] only when no clause and
no conjunction entails anything.
\end{Verbatim}
becomes
\begin{Verbatim}[fontsize=\scriptsize,breaklines=true,frame=single]
  - the clause is a mid-range demographic.

=== At least one concept ===
Every path must be annotated. Return "has_interpretation": true and at least
one concept; an empty list is not allowed. Build it from the clauses that say
the most about the patient's physiology. If no clause puts a variable clearly
inside an abnormal range, name the clinical state the strongest clause does
guarantee. Every concept still needs supported_by with the smallest sufficient
clause set, and still names only what those clauses guarantee.
\end{Verbatim}

\subsection{P2. Concept audit judge}
\label{app:prompt_audit}

\paragraph{System prompt.}
\begin{Verbatim}[fontsize=\scriptsize,breaklines=true,frame=single]
You audit one claim made about one path of a tabular risk model.

You are given a numbered list of predicate clauses and one clinical concept
that some annotator attached to a stated subset of those clauses. Decide
whether the cited clauses entail that concept.

=== What each clause guarantees ===
Every clause constrains one summary statistic of one variable over one window.
Read them literally:
  max at most a      every observation in the window was at most a
  min at least a     every observation in the window was at least a
  min at most a      at least one observation was at most a; the others may be
                     anything higher
  max higher than a  at least one observation exceeded a; the others may be lower
  mean / last        an average, or the final value; neither bounds the others
  delta              net change across the window; over a long window this is not
                     a rapid change
  std / count /      dispersion, number of observations, hours since the last one.
  ts_gap             These describe measurement, not physiology.

Two clauses on different variables constrain the same window, not the same
moment, so they never entail co-occurrence, persistence or a temporal sequence.

A clinical state may be named only when the cited clauses guarantee it for the
whole window. Normality may be asserted only on a scale whose maximum is the
healthy value (GCS, motor response); for every other variable — heart rate,
blood pressure, laboratory values and the rest — a healthy interval is not
defined here, so no clause supports a claim of normality.

You are NOT told what outcome the model predicts, and must not guess. Judge
only whether the cited clauses support the clinical claim. Judge the claim as
written: do not credit it for a weaker claim it could have made, and do not
penalise it for a stronger claim it could have made but did not.

Score:
  2  entailed — the cited clauses guarantee this state for the window
  1  weak — plausible but not guaranteed, or overstated relative to what the
     cited clauses bound
  0  unsupported — the cited clauses do not bear on this state, or point the
     other way

Respond with ONLY a single JSON object, no other text, no markdown fences:
{"verdict": <0, 1 or 2>, "reason": "<one sentence>"}
\end{Verbatim}

\paragraph{User prompt.} One call per (rule, concept) item; the mismatched-control items have exactly the same form. Filled with the P1 example rule and its first concept:
\begin{Verbatim}[fontsize=\scriptsize,breaklines=true,frame=single]
Predicate clauses:
  [1] HR__mean higher than 114.0000
  [2] static__Age at most 84.0000
  [3] NISysABP__min higher than 135.0000

Claim: clauses [1] entail the clinical concept "tachycardia".

Output JSON only.
\end{Verbatim}

\subsection{P3. Clinician instructions}
\label{app:clinician_readme}

The instructions below were given to both clinicians in full (as a Markdown file with an accompanying translation) and are reproduced here unchanged in content. The same text, in its Markdown form, is inserted verbatim into the LLM-judge system prompt (P4).

\begin{tcolorbox}[colback=white,colframe=black!45,boxrule=.4pt,arc=1pt,left=4pt,right=4pt,breakable]
\footnotesize
\textbf{Clinician Review --- Evaluation Instructions}

\textbf{Task.} You will evaluate four explanations for each of 50 ICU patients, for a total of 200 explanations. All four systems use the same patient measurements. Please independently rate each explanation on the following three 1--5 scales.
\begin{enumerate}[leftmargin=1.5em,itemsep=0pt]
\item \textbf{Source traceability}: Can you identify the original patient data on which the presented evidence is based? (Original data $\leftarrow$ selected/important evidence)
\item \textbf{Clinical interpretation quality}: Does the explanation appropriately translate the presented evidence into clinically meaningful interpretation? (Selected/important evidence $\rightarrow$ clinical interpretation)
\item \textbf{Clinical usefulness}: Assuming that the stated patient facts and clinical interpretations were valid, does the presentation help you understand the prediction? (Selected/important evidence + clinical interpretation $\rightarrow$ prediction understanding)
\end{enumerate}
The patient's actual outcome and the system identities are hidden. Labels A--D are reassigned for each patient. The predicted probability and decision threshold are provided only to identify the prediction being explained. Please do not evaluate whether the prediction itself is correct or clinically appropriate, and do not compare predicted probabilities across systems.

\textbf{Files for each patient.} \texttt{record\_\{id\}.png}: plots of the patient measurements; the dashed line indicates the prediction time. \texttt{record\_\{id\}.txt}: original numerical data for checking exact values. \texttt{\{id\}\_A.txt} to \texttt{\{id\}\_D.txt}: the four prediction explanations. All times are hours since admission. Use only the information provided up to the prediction time. Do not assume diagnoses, treatments, or clinical events that are not recorded in the supplied data.

\textbf{Evaluation procedure.} Separate practice cases are provided to calibrate the evaluation criteria and are excluded from the final analysis. (1) Review the patient record before opening any explanation. (2) Follow the order assigned in \texttt{viewing\_order.csv}, reviewing one explanation at a time.

\textbf{Feature definitions.} Unless otherwise specified, statistics are calculated within the stated time window. \emph{last}: the last measurement within the time window. \emph{mean}: the mean used by the corresponding system within the time window. \emph{minimum / lowest}, \emph{maximum / highest}: the minimum or maximum value within the time window. \emph{count / number}: the number of measurements within the time window. \emph{Change (delta)}: the last measurement minus the first measurement in timestamp order; a negative value indicates a decrease; change is defined as 0 when fewer than two measurements are available within the window. \emph{Hours since last X}: the elapsed time from the last measurement of X to the \textbf{end of the stated time window}; this is not necessarily calculated relative to the final prediction time. For readability, ``Hours since last albumin = 0.83 h'' may instead be presented as ``Last albumin measurement: 0.83 h before the end of this window [23.9--31.9h]''.

\textbf{Observed values and model conditions.} A model cutoff is not the patient's actual measured value. For example, ``mean heart rate $\geq$ 106 bpm [31.2--35.2h]'' means that the observed mean heart rate was higher than 106 bpm; the model condition was $\geq$ 106 bpm; and the statistic was calculated over the 31.2--35.2 h time window.

\textbf{Rating scales.} The intended evaluation flow is: original data $\rightarrow$ selected/important evidence $\rightarrow$ clinical interpretation $\rightarrow$ prediction understanding.

\emph{1. Source traceability.} Can you determine which original patient data the evidence highlighted by the explanation was derived from? Evaluate whether the relevant variable, time or time window, and statistic or condition are sufficiently specified. Do not lower the score merely because you would need to manually recompute a mean, maximum, or other summary statistic. The key question is whether you can identify \textbf{which original data should be checked}.
\begin{itemize}[leftmargin=1.5em,itemsep=0pt]
\item[\textbf{5}] Nearly all presented evidence is clearly linked to specific variables and times/time windows/statistics, allowing the relevant original data to be identified directly.
\item[\textbf{3}] Some evidence is clearly traceable, but a substantial portion requires searching or inference because the time, statistic, or source is unclear.
\item[\textbf{1}] For most of the presented evidence, it is difficult to determine which measurements or time periods it came from.
\end{itemize}

\emph{2. Clinical interpretation quality.} Does the explanation interpret the presented evidence in a clinically meaningful and appropriate way? Simply listing numerical values or indicating that a feature increased or decreased the model's predicted risk does not by itself constitute clinical interpretation. Lower the score when the explanation makes overly strong claims about diagnoses, treatment effects, causal relationships, or clinical states without sufficient supporting evidence.
\begin{itemize}[leftmargin=1.5em,itemsep=0pt]
\item[\textbf{5}] The important evidence presented by the explanation is interpreted in a clinically meaningful way, is well supported by the data, and contains no meaningful overstatement.
\item[\textbf{3}] Some useful clinical interpretation is provided, but it is incomplete, generic, questionable, or includes some meaningful overinterpretation.
\item[\textbf{1}] Little or no clinical interpretation is provided, or most interpretations are poorly supported or substantially overstated.
\end{itemize}

\emph{3. Clinical usefulness.} Assuming that the stated patient facts and clinical interpretations were valid, does the presentation help you understand the prediction? Consider: Is it easy to distinguish important information from less important details? Do the clinical interpretations and/or model-attribution information help you understand the evidence? Can you understand the basis of the prediction without unnecessary effort? Even if you would not agree with the prediction, can you understand why the model made this prediction from the presented evidence and clinical interpretations?
\begin{itemize}[leftmargin=1.5em,itemsep=0pt]
\item[\textbf{5}] The important evidence is easy to identify, and the presentation is highly helpful for understanding the basis of the prediction.
\item[\textbf{3}] Part of the prediction basis is understandable, but important evidence is difficult to distinguish, insufficiently explained, or obscured by repetitive or ambiguous information.
\item[\textbf{1}] The presentation provides little help in identifying or understanding the evidence underlying the prediction.
\end{itemize}

\textbf{Keep the three scales separate.} For example: the source may be clear but have little clinical meaning (high traceability, low clinical interpretation quality); the clinical interpretation may appear appropriate but its source may be unclear (high clinical interpretation quality, low traceability); a presented clinical state may help explain the prediction, but the evidence supporting that clinical state may be weak (high usefulness, low clinical interpretation quality). Do not automatically lower all three scores because of a single problem.

\textbf{Overall feedback after completing all evaluations.} After completing all quantitative ratings, please provide brief free-text feedback on each of the four anonymized explanation formats: what was helpful; what was difficult to understand or verify; what information you would add or remove; recurring strengths or limitations you noticed. This feedback will be used only as qualitative feedback and will be analyzed separately from the numerical ratings.
\end{tcolorbox}

\subsection{P4. LLM judge on the three clinician scales}
\label{app:prompt_judge}

\paragraph{System prompt.} \texttt{\{README\}} is replaced by the full Markdown text of P3, unchanged. \texttt{record\_\{id\}} inside this prompt is literal text, not a slot.
\begin{Verbatim}[fontsize=\scriptsize,breaklines=true,frame=single]
You are acting as one blinded reviewer in a clinical-explanation study. Score
ONE explanation of ONE ICU patient prediction on the three scales below, exactly as a human
reviewer would under these instructions.

{README}

You do not have `record_{id}.png`. In its place you are given `record_{id}.txt`, the exact
numerical values the plot shows; treat it as the original patient data for source
traceability. Intermediate scores 2 and 4 are allowed between the stated anchors.

Respond with ONLY a single JSON object, no other text, no markdown fences:
{"traceability": <1-5 int>, "interpretation": <1-5 int>, "usefulness": <1-5 int>,
"reasoning": "<one sentence>"}
\end{Verbatim}

\paragraph{User prompt.} One call per pass $\times$ patient $\times$ presentation. \texttt{\{id\}}, \texttt{\{letter\}}, the record text, and the explanation file are filled from the same packet the clinicians received.
\begin{Verbatim}[fontsize=\scriptsize,breaklines=true,frame=single]
PATIENT {id}, presentation {letter}

=== record_{id}.txt (original patient data) ===
{contents of record_{id}.txt}

=== {id}_{letter}.txt (the explanation to score) ===
{contents of {id}_{letter}.txt}
\end{Verbatim}

\paragraph{Blinding and independence.} A call carries the system prompt above and one user message: one patient's record and one presentation, identified only by patient id and letter. The letter-to-system key and the outcome are never sent, no method name appears in any of the 800 explanation files, and no call sees another call's answer: not the other pass on the same item, and not the other three presentations of the same patient. The two judges receive byte-identical prompts.

\end{document}